# On the Selective and Invariant Representation of DCNN for High-Resolution Remote Sensing Image Recognition

Jie Chen, Chao Yuan, Min Deng, Chao Tao, Jian Peng, Haifeng Li*

School of Geosciences and Info Physics, Central South University

**Abstract:** Human vision possesses strong invariance in image recognition. The cognitive capability of deep convolutional neural network (DCNN) is close to the human visual level because of hierarchical coding directly from raw image. Owing to its superiority in feature representation, DCNN has exhibited remarkable performance in scene recognition of high-resolution remote sensing (HRRS) images and classification of hyper-spectral remote sensing images. In-depth investigation is still essential for understanding why DCNN can accurately identify diverse ground objects via its effective feature representation. Thus, we train the deep neural network called AlexNet on our large-scale remote sensing image recognition benchmark. At the neuron level in each convolution layer, we analyze the general properties of DCNN in HRRS image recognition by use of a framework of "visual stimulation–characteristic response combined with feature coding–classification decoding." Specifically, we use histogram statistics, representational dissimilarity matrix, and class activation mapping to observe the selective and invariance representations of DCNN in HRRS image recognition. We argue that selective and invariance representations play important roles in remote sensing images tasks, such as classification, detection, and segment. Also selective and invariance representations are significant to design new DCNN liked models for analyzing and understanding remote sensing images.

# 1. Introduction

Given that numer[1]ous high-resolution remote sensing (HRRS) images are low cost and readily available, an intelligent and automatic manner becomes increasingly necessary for interpreting such massive images. However, the variable properties of ground features exhibiting on the HRRS images always bring difficulty in image interpretation in the remote sensing community because the intra-class variation increases and the inter-class discrimination decreases (Zhao and Du, 2016). Many studies on analyzing and understanding remote sensing images are connected to exploiting most of the discriminative information of various ground features. The established models can achieve an effective representation through considering the representation from the perspectives of encoding and decoding.

---

[1] Corresponding author: lihaifeng@csu.edu.cn

Humans can recognize scenes and objects in a way that is invariant to scale, translation, and clutter although the spatial complexity and structural diversity universally exist in HRRS images, indicating a powerful system of recognition with high robustness in the brain (Chen et al., 2017). As visual signals flow from one brain layer to the next to gain different levels of abstraction, deep neural networks (DNNs) attempt to mimic the ability of the brain to learn and develop hierarchical feature representations in a multi-level way. Deep convolutional neural network (DCNN) is a biologically inspired multi-stage architecture composed of convolution and pooling, which are conceptually similar to simple and complex cells in the brain. The concurrent variation for the various HRRS images and the contained ground objects should be ascertained in the hierarchical feature learning model to analyze the feature representation generated by the DCNN.

DCNN can effectively encode spectral and spatial information on the basis of the raw image itself in a natural manner. Unlike shallow learning techniques (e.g., sparse coding), DNNs are the best performing models on object recognition benchmarks based on computer vision and can yield human performance levels in object categorization (He et al., 2015). Owing to the semantic properties abstracted from the upper layers of the DNN, deep learning often outperforms conventional handcrafted feature extraction methods (Scott et al., 2017). Cichy et al. (2016) trained on real-world image categorization tasks and found that the hierarchical relationship in processing stages between the brain and DCNN similarly emerges in space and time. Moreover, the pre-specified model architecture, training procedure, and learned task are the influencing factors of the similarity relations between DCNN and brains (Cichy et al., 2016).

In the remote sensing domain, methods based on deep learning have achieved success in many different remote sensing applications, such as automobile detection (Chen et al., 2014), ship detection (Tang et al., 2015), image classification (Ma et al., 2015), and oil spill (Fingas and Brown, 2014). Moreover, CNN architectures have been promoted by introducing a rotation-invariant layer (Cheng et al., 2016), considering multi-scale contextual features (Zhao et al., 2015) or using a multi-scale learning scheme (Zhao and Du, 2016). Deep learning as the new state-of-the-art solution has been used extensively from pixel-based hyper-spectral image classification to scene-based high-resolution image classification (Nogueira et al., 2016).

Nogueira et al. (2017) evaluated three deep learning strategies for remote sensing image classification: (i) using CNN as feature extractor, (ii) fine tuning a pre-trained network, and (iii) training them from scratch. They concluded that the second strategy is the best one for the aerial and remote sensing domains after conducting six popular DCNN architectures (OverFeatnetworks, AlexNet, CaffeNet, GoogLeNet, VGG16, and PatreoNet) on three datasets. Under this background, the following questions arise:

(1) What features can the DCNN learn in an optimal problem-specific performance by finding a proper selection of hyper-parameters?

(2) Why does the DCNN perform recognition tasks well with merely thousands or millions of parameters?

(3) How different are the learned internal representations in intra- or inter-class images?

The performance mechanism of deep learning in the remote sensing domain is still unclear. The current study attempts to investigate the selective and invariant visual representations in DCNN for HRRS image recognition. The contributions of this work are threefold:

(1) The feature representation of neurons in each convolution of DCNN is ascertained by use of statistical and visualization methods for the first time on the basis of the visual feature stimulus of large-scale images (approximately 24,000 images of 35 categories). The neurons perform sparsely active to respond to the visual stimulus of remote sensing images with an indication in intra-class consistency and inter-class dissimilarity, and this performance is evident at the high level. Therefore, the selection of feature representation plays an important role in remote sensing image recognition.

(2) By observing the activation of all neurons to the visual stimulus of remote sensing images, DCNN as a human vision technique achieves the invariant representation for various visual characteristics, such as posture, size, shape, and color, in the image recognition task. Consequently, the invariant representation capability of DCNN for diverse features of remote sensing image is verified.

(3) Combining visual feature encoding with class-specific location on the same images, the selective and invariant feature encodings of DCNN contribute to the decoding of intrinsic features of ground objects in the recognition task. Furthermore, the introduced representational dissimilarity matrix (RDM) provides an intuitive understanding of remote sensing image recognition based on DCNN with its geometric similarity response.

The remainder of the paper is organized as follows. In Section 2, related works on feature representation and DCNN understanding are reviewed. In Section 3, the dataset and experimental setup are presented. In Section 4, the methodologies used in our work are introduced. In Section 5, the details of the experimental results and discussion are presented. In Section 6, the conclusions of the study are elaborated and future works are provided.

## 2. Related works

Generating discriminating models is crucial for the recognition of remote sensing images; thus, advancing feature extraction algorithms is highly demanded to effectively encode spectral and spatial information (Nogueira et al., 2017). Many researchers have developed algorithms, such as scale-invariant feature transform (Han et al., 2015), histogram of oriented gradients (Shao et al., 2012), and saliency (Zhang et al., 2015), to extract invariant characteristics. However, the need for such degree of invariance brings difficulty in dealing with various problems in remote sensing images (Long et al., 2017). For these hand engineering specific features, much invested effort is suitable only for selective use cases (Luus et al., 2015).

The bag of visual words model and its variants have also been investigated and achieved promising results in satellite image classification (Cheriyadat, 2014). However, its initial step of extracting low-level visual features heavily depends on the domain knowledge of the designer (Liu et al., 2016). The sparse coding models as a single-layer feature learning encodes an input signal by selecting a few vectors from a large pool of possible bases, but the shallow architecture has shown effectiveness only in solving simple or well-constrained problems (Deng, 2014). These handcrafted feature extractions are limited in that they may inadequately capture variations in object appearance. The reason is that the features on a remote sensing image are often affected by various factors, such as sensor observation angle, solar elevation angle, and seasonal climate.

Encoding of spatial information in remote sensing images is still an open and challenging task (Benediktsson et al., 2013). In the computer vision domain, a long-time goal has been to build a system that achieves human-level recognition performance by emulating image recognition in cortex (Serre et al., 2005). Deep learning is inspired by the hierarchical cognition process of the human brain, a significant advancement in artificial intelligence, and has shown great potential for discriminative feature learning without human intervention (LeCun et al., 2015). Chen et al. (2017) presented a computational model of feed-forward ventral stream based on the properties of invariance theory and CNNs for exploring several aspects of invariance to scale, translation, and clutter in object recognition.

The biological contribution of neural networks to the DNN results has been extensively discussed. Kriegeskorte and Kievit (2013) linked neuronal activity to representational content and cognitive theory by considering the activity pattern across neurons as a representational geometry and thus revealed the concept of cognitive representation in the brain. Cichy et al. (2016) investigated the coding of visual information in the object recognition DCNN by determining the receptive field selectivity of the neurons in an eight-layer AlexNet architecture to understand visual object recognition in the cortex. In the said research, functional magnetic resonance imaging was compared with DNN based on the idea that, if two images are similarly represented in the brain, then they should also be similarly represented in the DNN.

From theoretical and empirical perspectives (Li et al., 2015), many recent studies have paid attention on understanding DNNs. For example, Paul and Venkatasubramanian (2014) studied a key algorithmic step called pre-training from the perspective of group theory, and this step explains why a deep learning network learns simple features first and representation complexity increases as the layers deepen. Yosinski et al. (2015) introduced two tools for visualizing and interpreting trained neural nets to understand the performance of intermediate layers. The intuition gained from these tools can help researchers in developing improved recognition methods. Mahendran and Vedaldi (2015) presented an invert representation framework to analyze the learned visual information and found that CNNs can retain photographically accurate information with different degrees of geometric and photometric invariance. Wei et al. (2015) visualized the intra-class knowledge inside the CNN to better understand how an object class is represented in the fully connected layers and demonstrated how CNN organizes different styles of templates for an object class. Furthermore, Zhou et al. (2016) proposed a general technique called class activation mapping (CAM) for CNNs with global average pooling (GAP) to show the remarkable localization capability of CNN. These scenarios help attain deep understanding of neural net but fail to determine the similarity and dissimilarity of the CNN learned internal representations in image recognition task. An impressive performance on texture recognition was obtained by pooling CNN features from convolutional layers by Fisher coding (Cimpoi et al., 2015). As features from convolutional layers are more generic than those from fully connected layers (Yosinski et al., 2014), the advantage of the features from all of the convolutional layers has attracted more attention than before (Long et al., 2017; Zhou et al., 2017).

Some observations comprehensively examined the black box of neural network models. Li et al. (2015) used a convergent learning to determine whether separately trained learning feature of DNNs can converge to span similar spaces. Alain and Bengio (2016) explored the dynamics inside a neural network and the role played by the individual intermediate layers by use of a linear classifier probe as a conceptual tool. Such neural network models can be designed in accordance with certain heuristics. In a high-performance deep convolutional network (DeepID2+) for face recognition, Sun et al. (2015) empirically discovered three properties of the deep neuronal activations critical for high performance: sparsity, selectiveness, and robustness. Accordingly, this work determined deep learning and its connection with existing computer vision studies, such as sparse representation.

In the remote sensing domain, Castelluccio et al. (2015) demonstrated that pre-trained CNN can be used to land use classification on the basis of CaffeNet and GoogLeNet with three design modalities. Penatti et al. (2015) claimed that pre-trained CNNs generalize well for aerial images and the potential for combining multiple CNNs and other descriptors. Hu et al. (2015) investigated the transfer activations of CNNs not only from fully connected layers but also from convolutional layers and showed that pre-trained CNN can be used for HRRS scene classification tasks. Hu et al. (2015) visualized the representations of each layer to intuitively understand the CNN activations and found that features of convolutional layers can be reconstructed to images similar to the original image. However, the use of CNNs for HRRS images and the features from convolutional layers have not been comprehensively investigated yet.

In the present study, many possible combinations of the neuron population within individual convolutional layers are considered to provide a multi-dimensional representation space. The set of all possible remote sensing images corresponds to a vast set of points in the space and forms the geometry that defines the nature of the representation. We characterize the geometry of a representation by comparing the activity patterns with the dissimilarity distance between their points in the representational space. Furthermore, we calculate the histogram of the activated neural numbers on each of the images and the histogram of the number of images on which each neuron is activated to verify the invariance on neural representation from a macro perspective. Corresponding to the above-mentioned statistical measurement, this study further tests whether different classes of images can be discriminated in the representation by classifier analysis. For this purpose, CAM is adopted to highlight the discriminative object parts detected by CNN.

## 3. Dataset and experimental setup

The dataset used in this study is a remote sensing image classification benchmark called RSI-CB-256 (Li, 2017), which contains approximately 24,000 images of 35 categories. As shown in Fig. 1, the size of each image is 256×256. The dataset possesses a) a large number of images (an average of 690) in each class, b) 35 categories covering most ground objects and keeping high inter-class diversity and intra-class variety, c) high spatial resolution for providing remote sensing with abundant visual features, and d) various intra-class characteristics that can be effectively utilized to improve the generalization and robustness of a training network model.

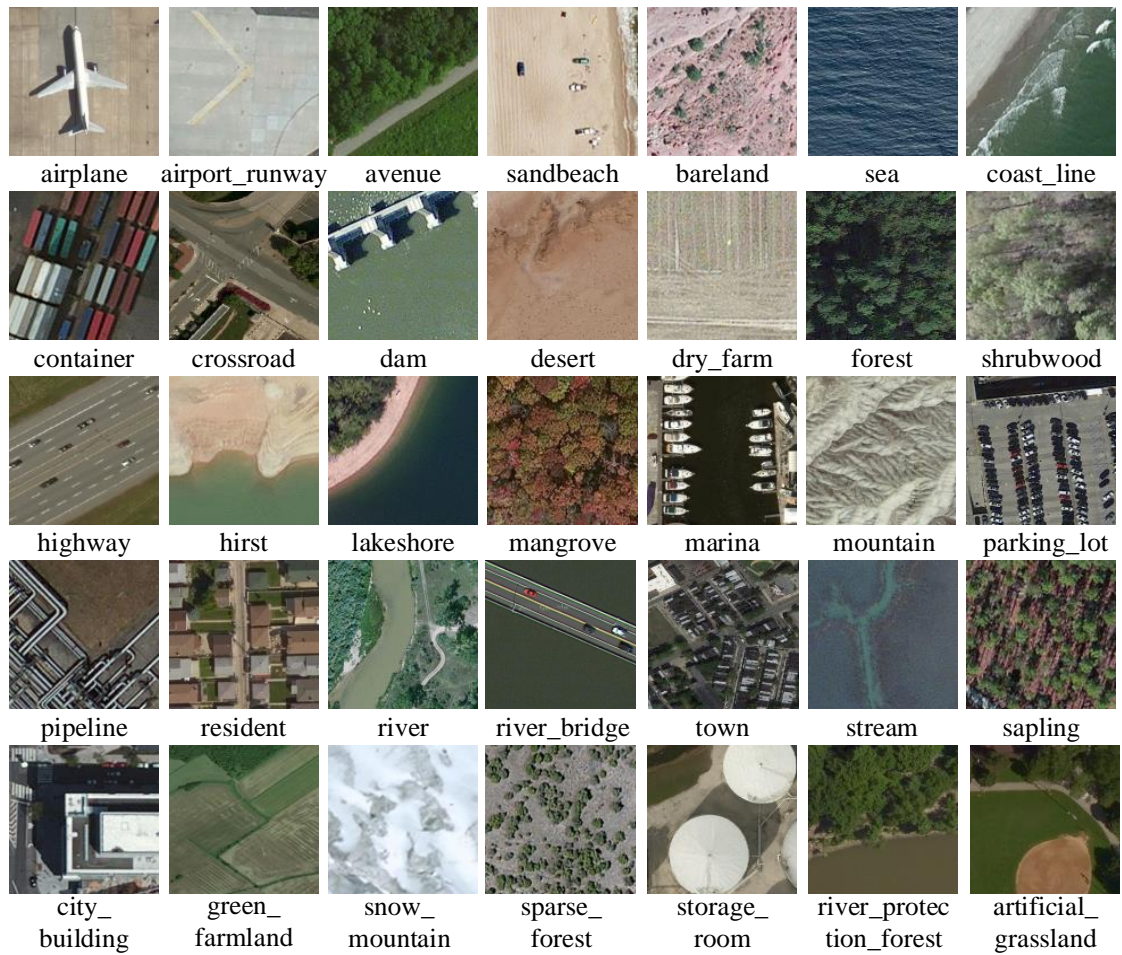

Figure 1. Object samples from RSI-CB-256

AlexNet obtained the first place in the 2012 ImageNet image classification challenge, is a classic deep network model developed in recent years, and has been widely used in image recognition and classification. Cichy et al. (2016) used this model to natural image recognition and concluded that its eight-layer structure can be compared with the hierarchical response of human visual cortex. In the current study, AlexNet is trained on the remote sensing image classification dataset called RSI-CB-256. Experimentally, the dataset is divided into training set, validation set, and testing set in a ratio of 8:1:1. According to the experimental results, the recognition accuracy of this model reaches 94.47%.

Considering the overall feature distribution, 80% of the images in RSI-CB-256 are randomly selected as the samples. All samples are input into the high-performance network AlexNet to acquire the neuronal response shown as the output (termed as feature map) of each convolutional layer, as shown in Fig. 2. The feature map of five convolution layers in the network model are regarded as the initial data.

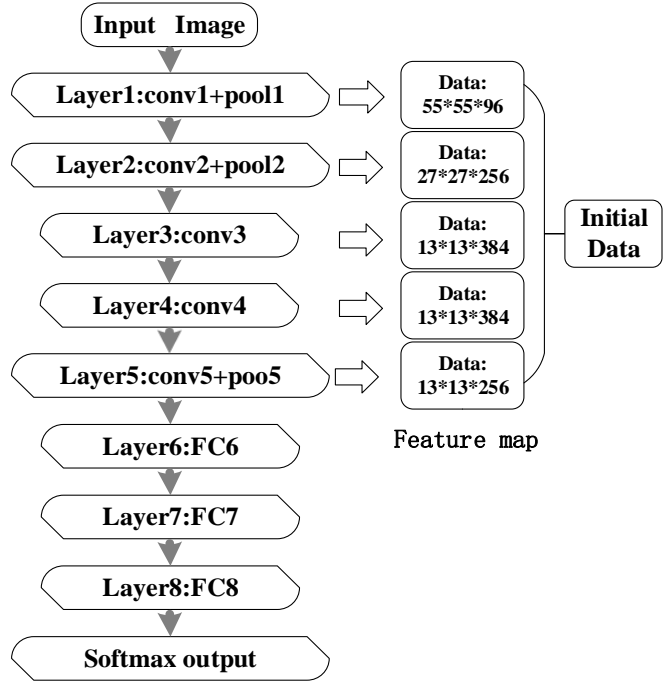

Figure 2. Experimental network model

## 4. Methodology

### 4.1 Statistical analysis on neuronal activations

DCNN has been further optimized and improved to achieve an intrinsic representation of images. Sun et al. (2015) investigated the neuronal activations of DeepID2+ nets in face identification and verification by implementing observations on LFW and YouTube Faces via statistics in the form of activation distributions and histograms. As a result, moderately sparse and highly selective neuron subsets were found in identities and identity-related attributes. The current study is inspired by activity pattern statistics and investigates the representation properties of deep network in recognizing different remote sensing images and ground object characteristics.

The visual stimulation from each image can generate an activity pattern in every convolution layer. The five convolution layers correspond to five activity patterns with the dimensions as the number of the involved neurons. One image is taken as an example to show the data processing for an activity pattern in conv1. The process involves five steps as follows ($i$ is an image in a certain class; $S$ is the number of class; $n$ is the number of neuron in conv1, i.e., $n$ = 1, 2, 3 ... 96):

**Step 1:** Construct a convolutional feature map matrix: $M=[m_1; m_2;\cdots; m_n]$, where $m_n$ is a feature map with a size of 55×55, thereby resulting in $M=55\times55\times96$.

**Step 2:** Normalize the feature map matrix through $G = \frac{M}{\max(M)}$. Values in $M$ are normalized to 0–1, such that $G= [g_1; g_2\ldots; g_n]$, where $g_n$ is a normalized feature map.

**Step 3:** Calculate the neuronal response value: $V_n$ = mean(0.8*$G_n$). Considering the feature distribution of the sample images, 80% of the values in each feature map $m_n$ are used to obtain an average value as the response value of one neuron. Accordingly, $V = [v_1; v_2; ...; v_n]$, $v_n$ is the response value of the $n^{th}$ neuron, and $V=1\times96$ is the activity pattern in conv1.

**Step 4:** Calculate a threshold for determining the neuronal response: $T = \text{mean}(G_i^S)$. $G_i^S$ denotes all the normalized response values of the $i^{th}$ image of class $S$ in conv1. By averaging all the normalized response values, each category in each convolution layer gains a different $T$.

**Step 5:** Determine the effective response value of neuron. The response value of a neuron larger than $T$ is defined as the effective response value (neuronal activation). The other values are assigned as 0 (neuronal inhibition). In this way, effective response value and 0 finally constitute a neuronal activity pattern for an image in each convolution layer.

## 4.2 Neuronal representational similarity analysis

Li et al. (2015) investigated the underlying representations learned on their intermediate convolutional layers to determine whether different neural networks can learn the same feature representation. In the said study, a bipartite matching approach that makes one-to-one assignments between neurons and a spectral clustering approach that finds many-to-many mappings were introduced to approximately align neurons from two networks. The results suggested that the correlation measurement is an adequate indicator of the similarity between two neurons.

Neurophysiology has long interpreted the selectivity of neurons across, which is the pattern of activity representing the content. In a multi-dimensional space taking the neurons as its dimensions, Kriegeskorte and Kievit (2013) used the RDM to measure the dissimilarity of two patterns between their points, which correspond to the set of all possible objects. On the basis of the representation geometry of these points, the RDM can determine which distinctions the stimulus honors or disregards. Furthermore, different representations can be easily compared by computing the correlation between RDMs.

A basic assumption is that the remote sensing images of a same class should stimulate similar neuronal activities of DCNN. In the present study, RDM is adopted to calculate and visualize the correlation among activity patterns and thus survey the neuronal representation for diverse characteristics of different remote sensing images. By calculating and visualizing the dissimilarity among activity patterns of different features, RDM can create an intuitive feature expression on intra- and inter-class images, thereby revealing the representation properties of deep network model in remote sensing image recognition. The dissimilarity between activity patterns is calculated as 1 minus correlation which represents Pearson's correlation coefficient between two different activity patterns. As a result, the diagonal elements in the RDM result in a dissimilarity value of 0 for identical activity patterns and greater than 0 for off-diagonal elements. This finding shows that certain dissimilarity exists between different activity patterns (Fig. 3).

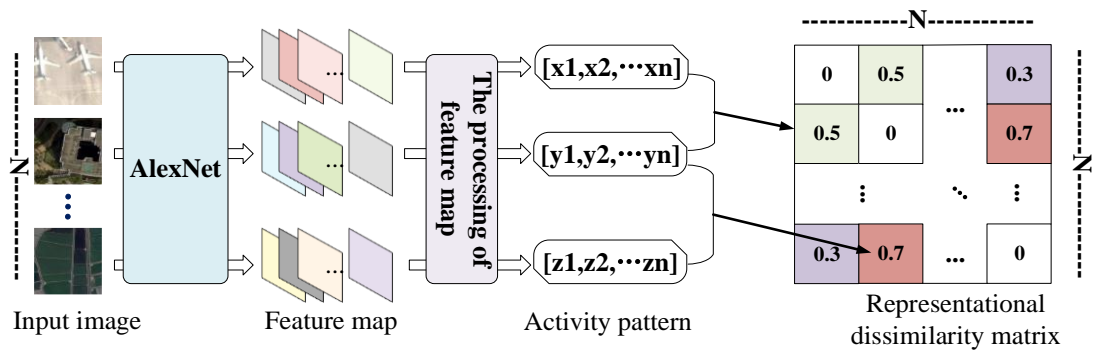

Figure 3. Process of calculation and visualization of RDM. Each stimulus image generates an activity pattern. RDM is an intuitive expression of dissimilarity coefficients between different activity patterns.

## 4.3 Decoding by localizing class-specific image regions

GAP combined with CAM (Zhou et al., 2016) is performed on the convolutional feature maps. Then, the importance of the image regions is identified by projecting back the weights of the output layer on the feature maps. As a result, the discriminative image regions used by the classification-trained CNNs are determined for identifying a particular category, thereby highlighting the discriminative object parts detected by the CNN.

The images from a same class possess similar visual features, and vice versa. In this case, we utilize CAM to visually locate the spatial distribution of essential features. Different from the original AlexNet, the fully connected layer is replaced with a GAP to avoid the loss of spatial information and overfitting of parameters. As shown in Fig. 4, the weights of neurons are evaluated in each convolution layer in accordance with the recognition result. Then, these weights are combined to obtain the activation value for all pixel positions of input image. A simple modification of GAP combined with CAM is linked to every convolution layer to achieve discriminative localization of class-specific image regions and thus ascertain the activities of hierarchical neurons caused by individual visual stimuli (Fig. 5).

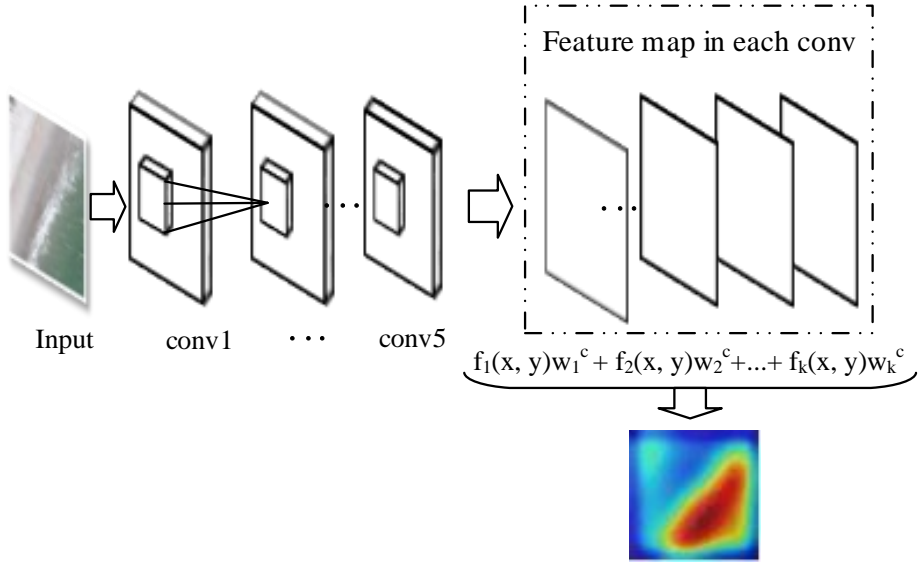

Figure 4. Calculation and realization of CAM

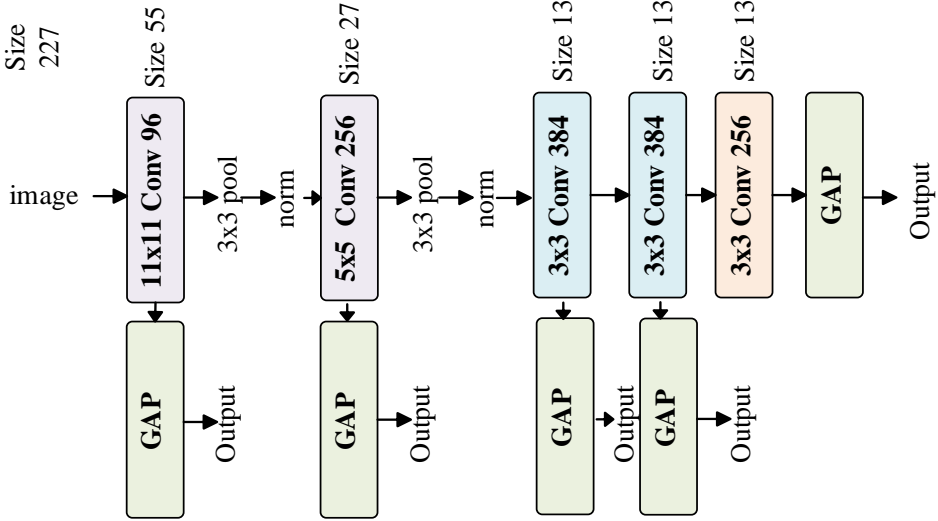

Figure 5. Improved design of CAM network structure

# 5. Results and discussion

## 5.1 Selective responses to visual stimuli

We aim to observe the responses of neurons to remote sensing images in the convolution layers. For this purpose, we access the neuronal responsive properties by performing statistical analysis from three aspects: 1) statistics of neuronal response patterns, 2) comparative histogram statistics between the number of activated neurons and the number of visual stimuli, and 3) histogram statistics of typical neuronal responses.

**A. Statistics of neuronal response patterns**

Neurons respond distinctly to either different images or diverse objects, thereby generating a different activity patterns in the convolution layers. As shown in conv1–conv5 of Table 5 in the Appendix, statistic of neuronal responses is carried out to understand the different activity patterns caused by dissimilar classes of images in the network. This task can be mainly achieved from the following two aspects:

(1) Dissimilarity in neuronal response intensity

The neurons from different convolution layers respond with dissimilar intensities to an identical stimulus, and different response intensities are produced by dissimilar objects in the same convolution layer. For example, the neuronal response intensities in conv1, conv2, and conv5 for airplane images are generally small and range from 0 to 0.1. On the contrary, approximately 15% of neurons in conv3 and conv4 exhibit response intensity larger than 0.1 for airplane images. This trend also is also observed in other classes.

(2) Neuronal activation or inhibition

Different classificatory images containing dissimilar objects and different visual characteristics of classificatory image or object lead to specific neuronal activation or inhibition. The activation or inhibition acted by the particular neurons to the characteristics of images or objects occurs in each of the five convolution layers, showing moderate sparseness. To further observe this sparseness, the histogram statistics on the number of activated neurons and the number of images involved are compared in the next section.

## B. Quantitative histogram statistics of activated neurons and visual stimuli

Table 1 shows the histogram statistics between the number of neurons in each layer and all the sampled images of 35 categories. The left side of the table shows the number of activated neurons on each image, and the right side shows the number of images on each activated neuron.

As shown in the left side of Table 1, around $30 \pm 18$ neurons in conv1 respond to approximately 17,200 images (accounting for nearly 94.3% of all the sampled images), around $70 \pm 20$ neurons in conv2 respond to approximately 17,100 images (accounting for nearly 93.7%), around $120 \pm 25$ neurons in conv3 respond to approximately 17,300 images (accounting for nearly 94.8%), around $120 \pm 30$ neurons in conv4 respond to approximately 17,100 images (accounting for nearly 93.7%), and around $60 \pm 28$ neurons respond to approximately 17,500 images in conv5 (accounting for nearly 95.8%). As shown in conv1–conv5 of Table 6 in the Appendix for a single category, 40, 80, 120, 120, and 80 neurons in conv1, conv2, conv3, conv4, and conv5, respectively, respond to most images. Most different classificatory images or diverse image features awake the small proportion of different neurons in each convolution layer, i.e., sparse neuronal responses to visual stimuli.

The sparse property of the neuronal representation is also illustrated at the right side of Table 1. In particular, partial images stimulate most of the neuronal responses in each convolution layer. For example, nearly $6,500 \pm 1,500$ images stimulate approximately 48 neurons in conv1 (accounting for around 50% of all neurons in conv1), nearly $5,500 \pm 2,500$ images stimulate approximately 148 neurons in conv2 (accounting for around 58%), nearly $5,500 \pm 2,500$ images stimulate approximately 300 neurons in conv3 (accounting for around 78%), nearly $6,000 \pm 2,500$ images stimulate approximately 320 neurons in conv4 (accounting for around 58%), and nearly $4,000 \pm 2,000$ images stimulate approximately 235 neurons in conv5 (accounting for around 58%). Furthermore, the response property is reflected not only in all the sampled images but also in each of the related 35 classes (conv1–conv5 of Table 7 in the Appendix).

From the above-mentioned histogram statistics, two response properties are obtained as follows: a) partial neurons respond to most of the images (left side of Tables 1 and 6) and b) partial images stimulate most of the neuronal responses. In other words, most of the neurons respond to partial images (right side of Tables 1 and 7). They reveal the sparse representation of DCNN, i.e., the stimulus response is performed by the effective stimulation of partial neurons, with different images stimulating distinct neuronal responses.

Table 1. Relationship between the number of neurons and the number of images

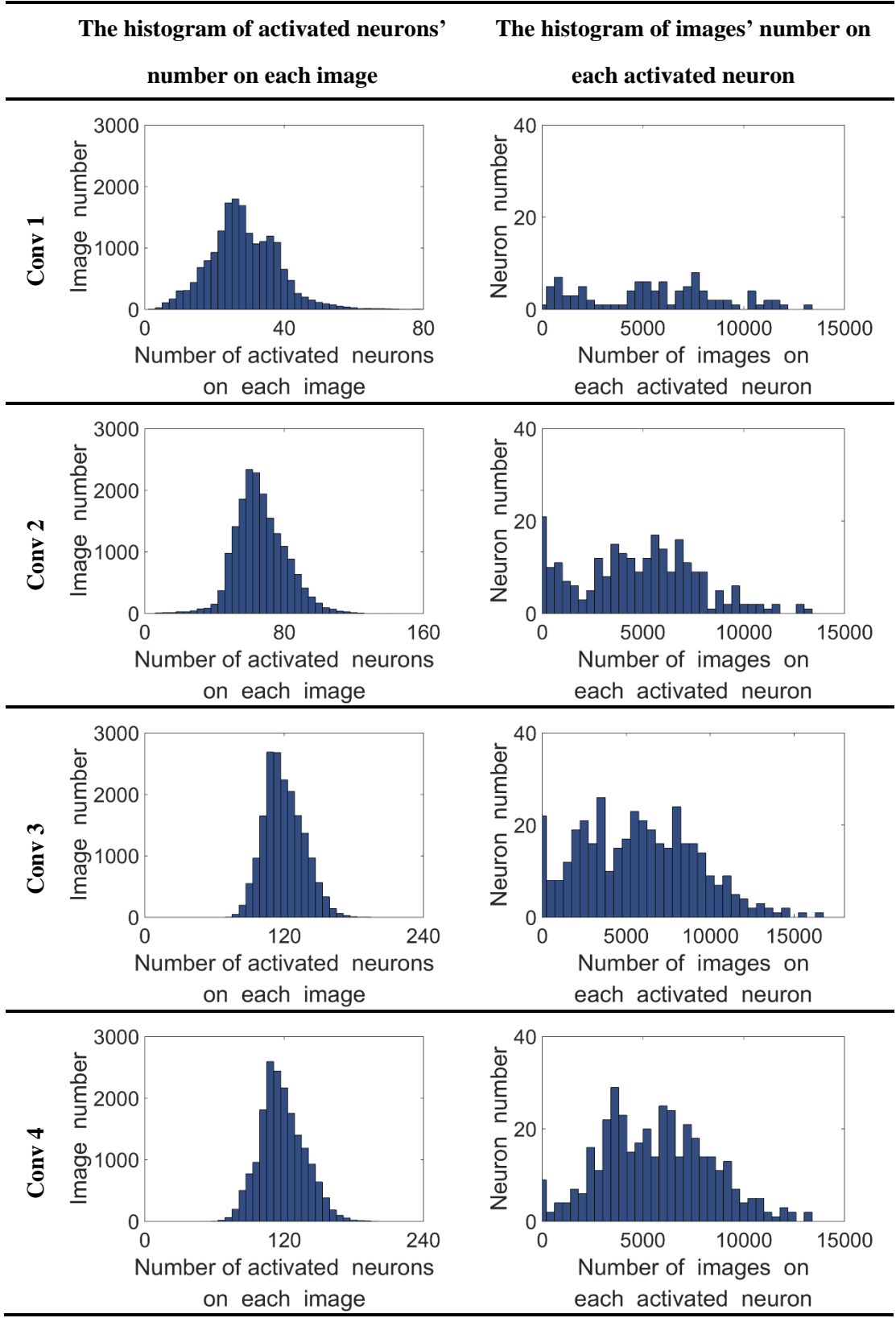

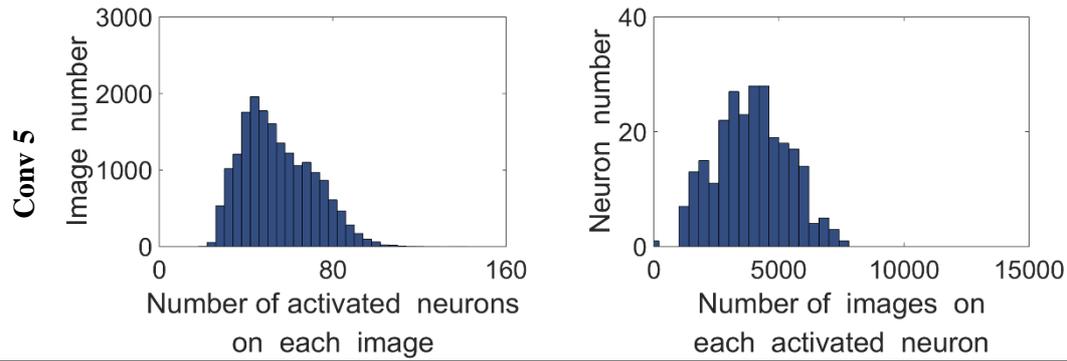

Table 1. Left: Histogram of the number of activated neurons on each image. The abscissa in the graph indicates the number of activated neurons on each image, and the ordinate shows the number of images. Right: Histogram of the number of images on each activated neuron. The abscissa indicates the number of images on each activated neuron, and the ordinate shows the number of neurons.

## C. Histogram statistics of typical neuronal responses

The dissimilarity of neuronal activity patterns depends on the selective responses of neurons, i.e., sparse neuronal activation or inhibition. Several typical neurons in the five convolution layers are observed statistically to calculate the response rate of involved images in related class and thus investigate the selective response. For example, neuron35 in conv1 shows constant activation to airplane and bareland images but constant inhibition to stream images. Unlike the response characteristic of neuron35, neuron55 presents constant activation to stream images but constant inhibition to airplane and bareland images. All these typical neurons from the five convolution layers generate sustained activation to most or even all images of a specific class whereas sustained inhibition to images of other categories. As shown in Table 2, the individual neuron in each layer shows a significant dissimilar response to different categories or different intra-class images with various features, such as shape, color, size, and position.

The above-mentioned investigations suggest that neuronal activation or inhibition is not only evidently sparse but also strongly selective, i.e., typical neurons generate activation or inhibitory effects on specific image features. This phenomenon reveals the strong selective response in the feature representation of DCNN.

Table 2. Typical neuronal responses to different object classes

| Conv 1 | | |
|---|---|---|
| **airplane** | **bareland** | **stream** |

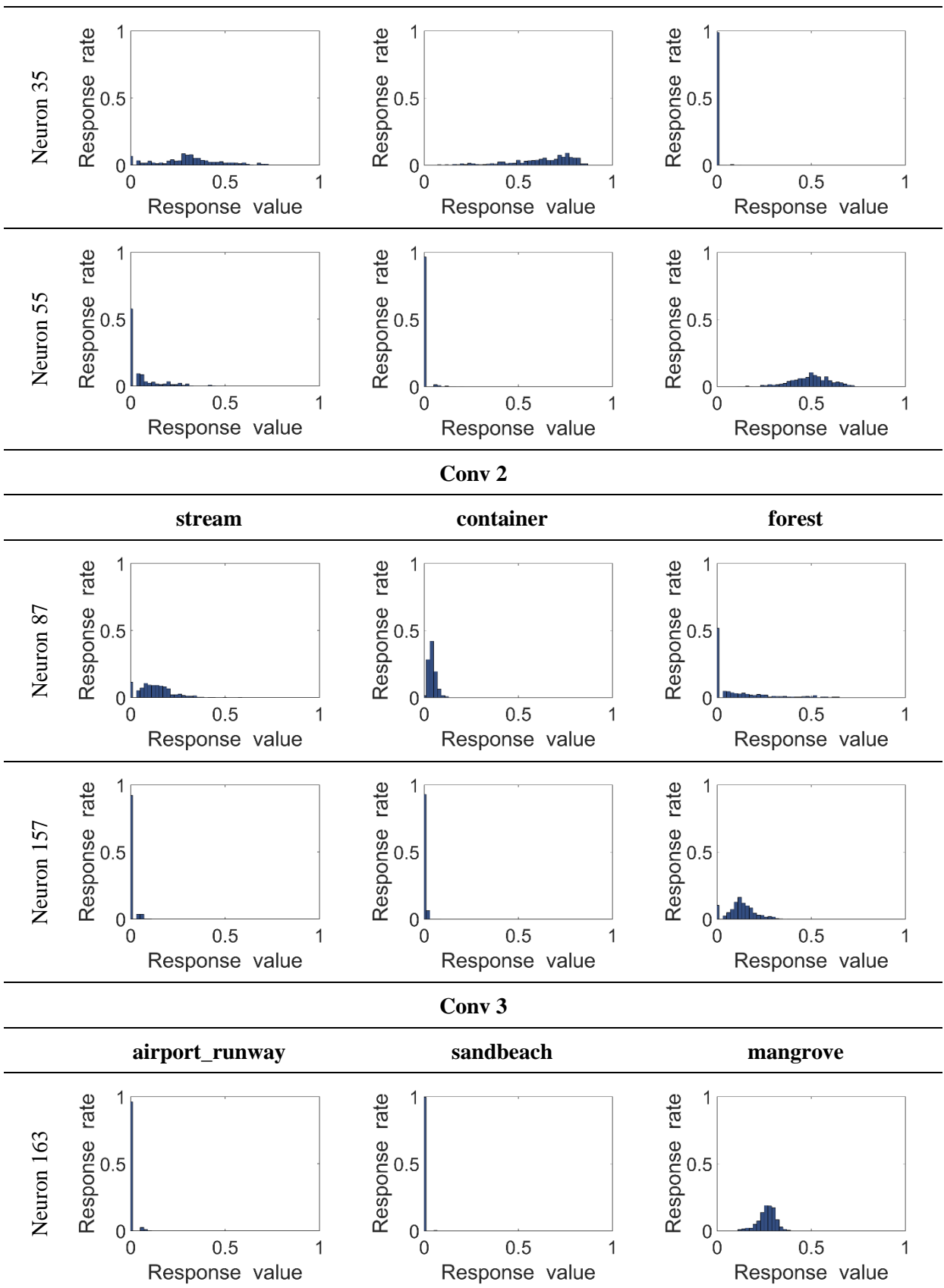

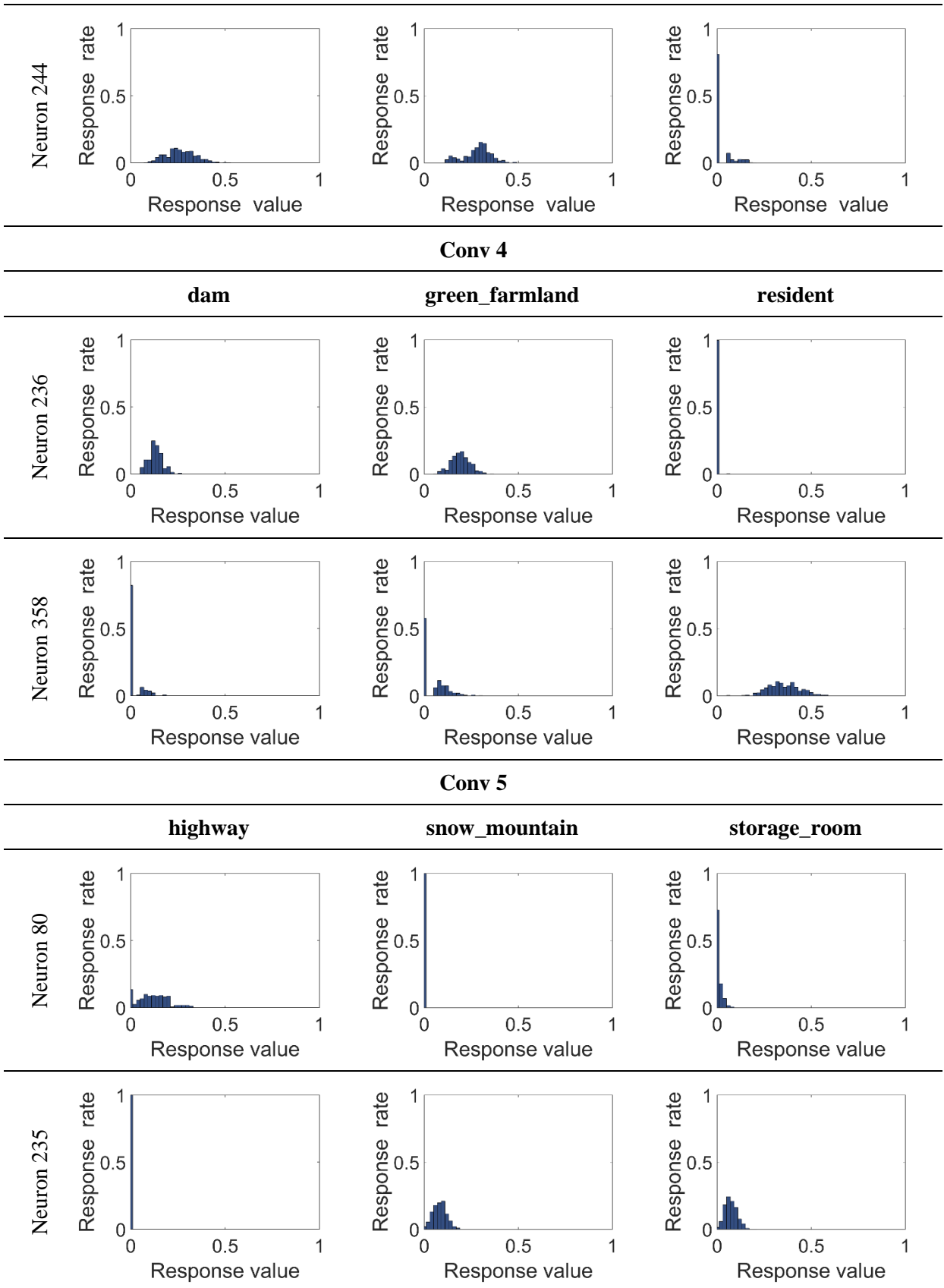

Table 2. Statistics of response values of typical neurons in each convolution layer. The graphs show the response value of typical neurons to different objects. The abscissa indicates the response value of the neuron to the intra-class images, and the ordinate indicates the response ratio of the neuron to the intra-class images.

## 5.2 Contribution of selective response to intra-class consistency and inter-class dissimilarity

RDM is used to calculate and visualize the dissimilarity of activity patterns between intra- and inter-class images and thus obtain the dissimilar activity patterns of neuronal response to different remote sensing images. Accordingly, we preprocess the data using the following four steps:

**Step 1:** Divide the intra-class images into 12 groups after sorting all images in each category in descending order according to the Top-1 recognition accuracy, thereby resulting in 12 groups for each category.

**Step2:** Select the first 12 images from each group of every category to constitute a subset, i.e., each subset contains 12 images from 35 classes, thereby resulting in 420 images for each subset. In each subset, the 420 images keep the accurate sort in their own category and present their corresponding activity patterns.

**Step3:** Calculate the RDM for each subset to obtain the dissimilarity representation between every pair of activity patterns and to visualize their representational geometry in the response space.

As a result, all RDMs of all 12 subsets in five convolution layers are $420 \times 420$ matrixes (left side of Table 3). Combined with the processing flow in Fig. 3, the visualization diagrams in Table 3 show the dissimilarity between different pairs of activity patterns in dissimilar subsets. As shown in the legend, the dissimilarity coefficient is reflected by the color.

To obtain a clearer visual observation of RDM than all the 35 classes involved, 10 classes of different features are selected for the RDM calculation and visualization. The selected 10 categories include airplane, bareland, city_building, storage_room, parking_lot, airport_runway, green_farmland, coast_line, highway, and resident. As in the previous steps, we select the first 12 images from each group in 10 classes to build a subset. We calculate and visualize RDM for all 12 subsets in each convolution layer, thereby resulting in RDM with $120 \times 120$ matrix (right side of Table 3).

In the neuronal response space, the difference between inter- or intra-class images is reflected by the dissimilarity coefficient (visualized by different colors) of RDM. The bluish color in the RDM diagonal region indicates the dissimilarity between intra-class images, while the yellowish color in the RDM off-diagonal region reflects the dissimilarity between inter-class images.

Comparing these RDMs shows that the representational geometries of intra- and inter-class images exhibit the following properties:

a) Although the 12 subsets have been sorted by recognition accuracy (the image with the highest recognition accuracy in subset 1), the discrimination of inter-class image still exists in all subsets and is evident in the high-level convolutional layer.

b) The difference in colors shows that the dissimilarity of intra-class activity patterns is lesser than that of inter-class activity patterns.

c) Some similar colors (with different depths) appear in the off-diagonal regions of RDM in conv1 and conv2. Therefore, the low-level convolutional layer likely generates similar feature representation, such as edge, color, direction, or texture. This response is selective but slightly weak.

d) The off-diagonal region of RDM in conv3, conv4, and conv5 does not show any blue bulk. Therefore, DCNN generates great dissimilarity to different activity patterns in the upper level wherein meaningful representations are generated.

Table 3. RDM of different subsets

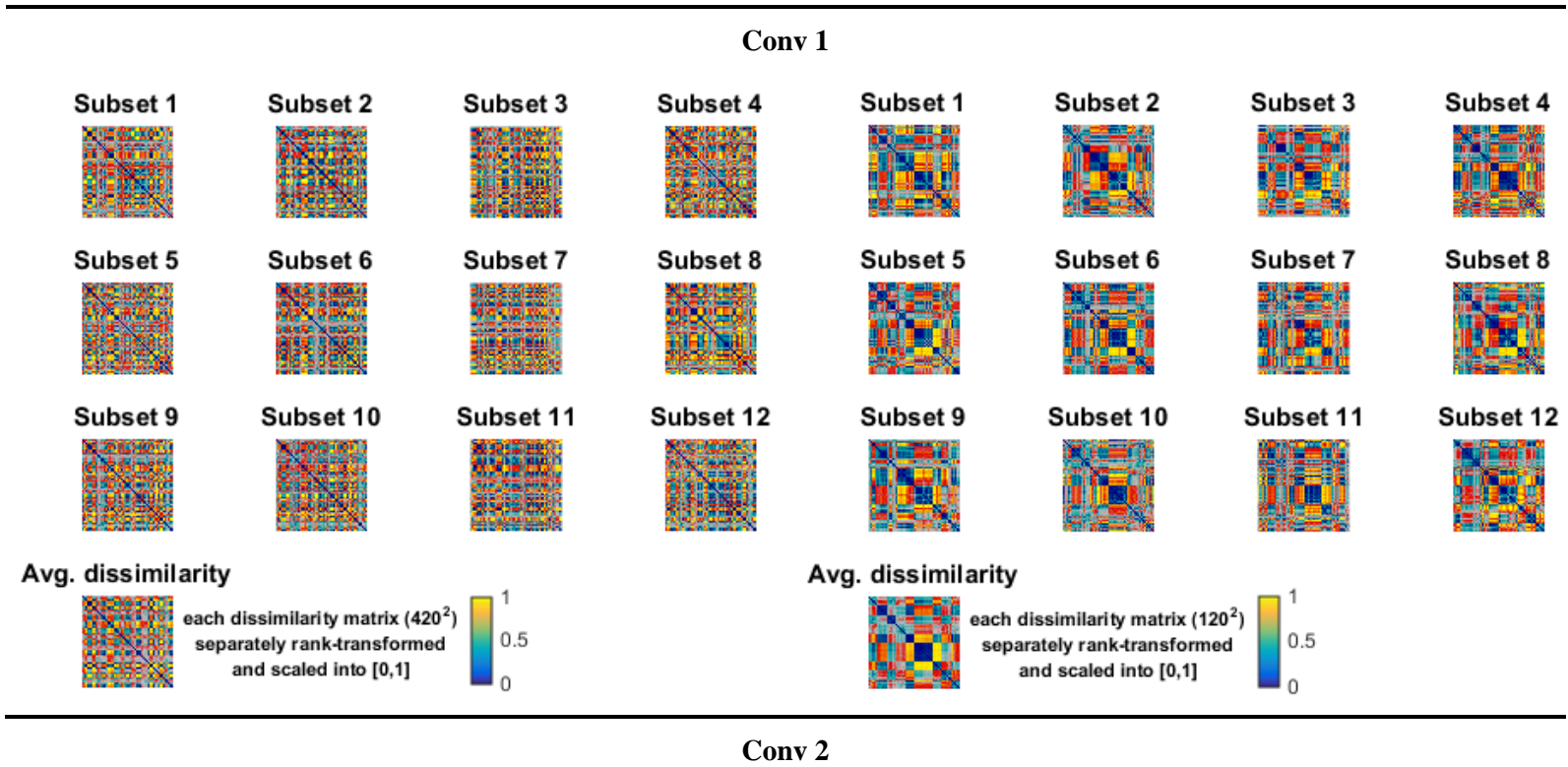

**Conv 2**

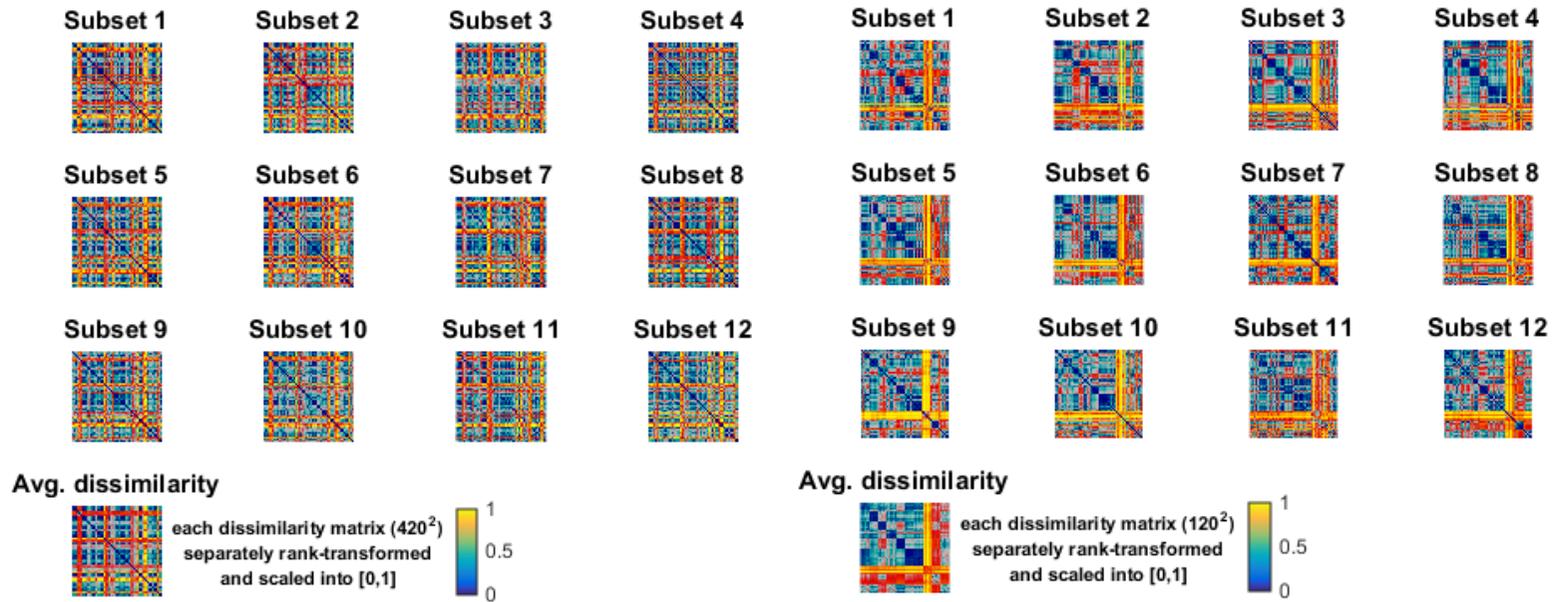

Conv 3

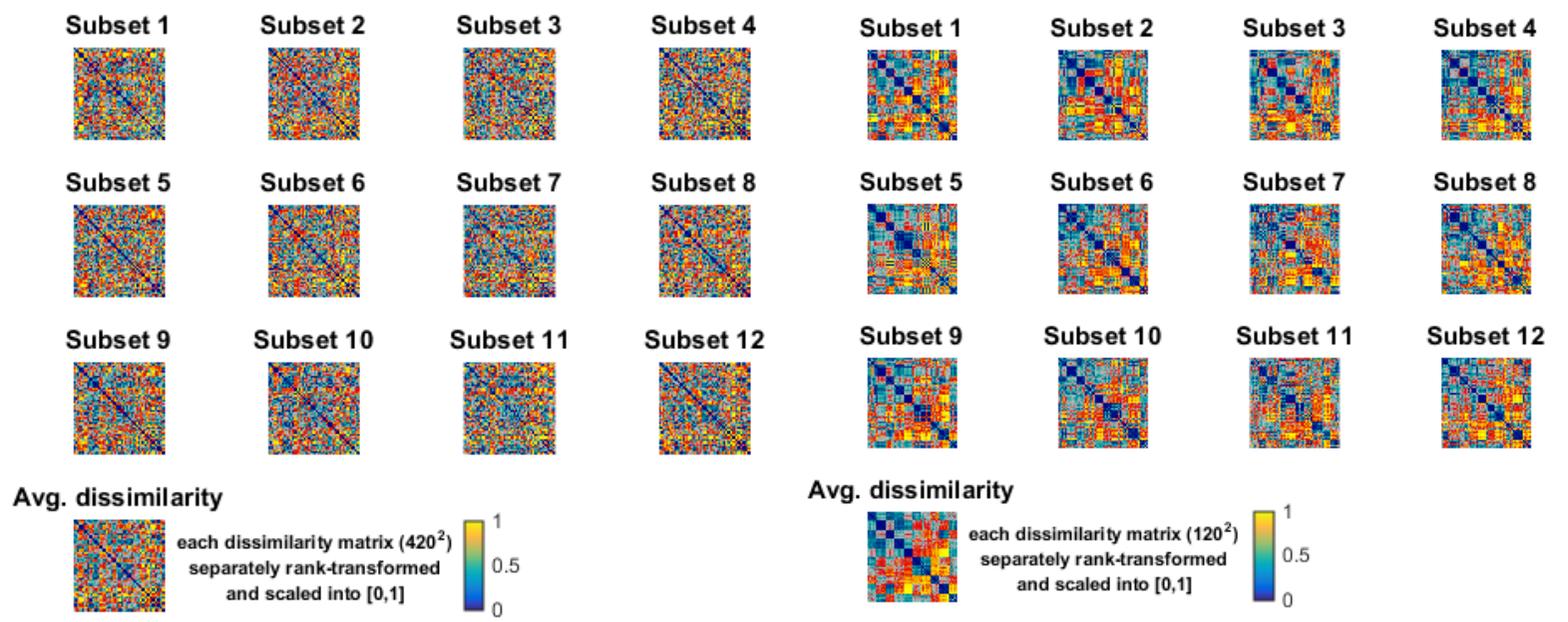

**Conv 4**

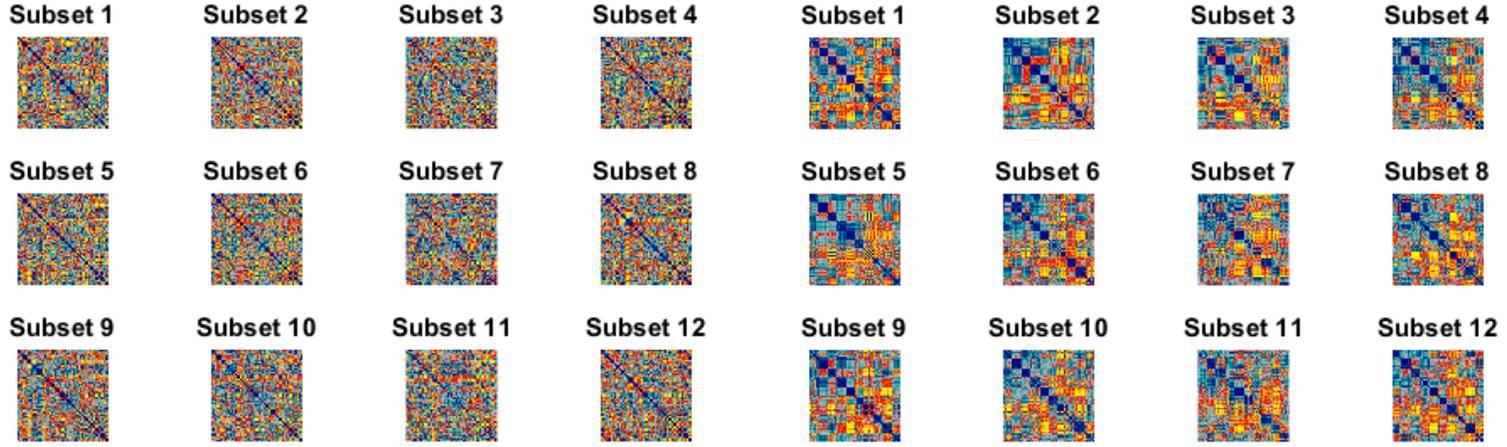

**Conv 5**

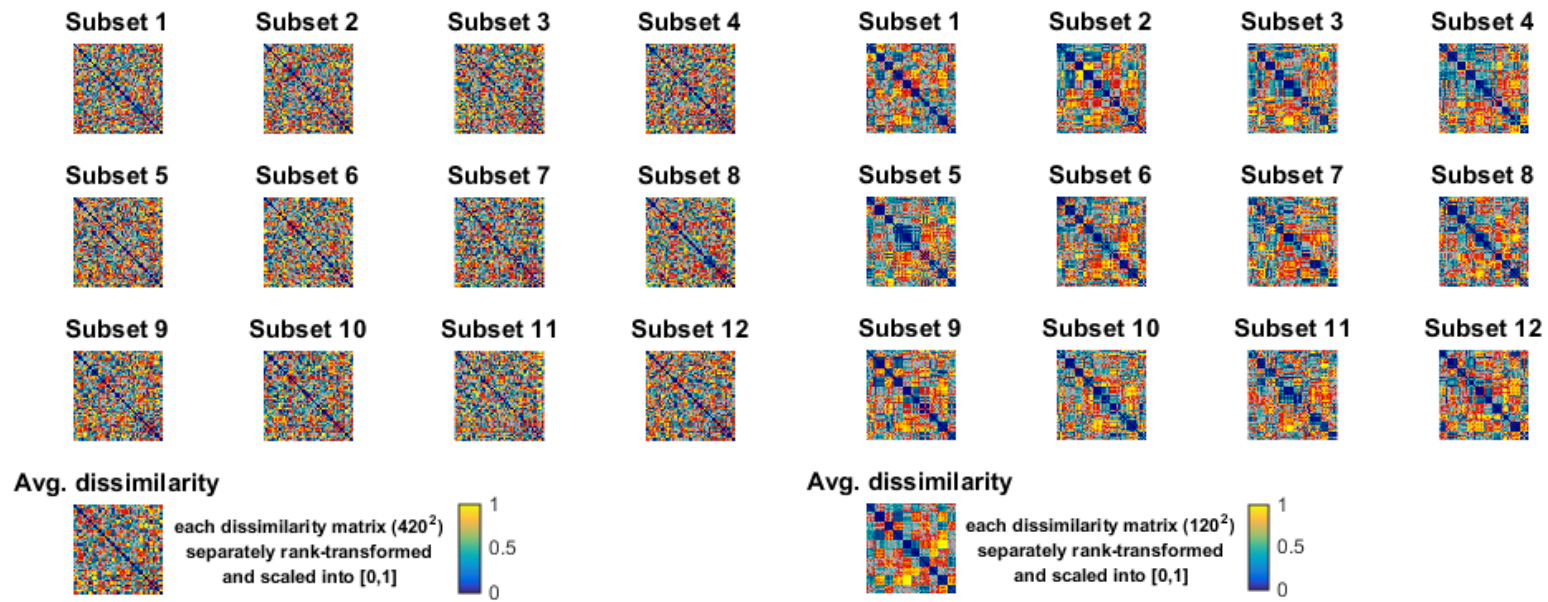

Table 3. Left: RDM visualization of different subsets including 35 categories of objects. Right: RDM visualization of different subsets including 10 categories of objects. A total of 10 categories of objects are selected to perform the same RDM visualization (left graph) and thus verify the experimental results.

## 5.3 Invariant representation of diverse visual stimuli

Deep network is capable of hierarchical feature learning to produce an abstraction of feature from low level to high level. DCNN shows a certain invariant representation in image recognition because of its feature learning capability, but this invariant representation has not been fully understood in remote sensing image recognition.

To further reveal the invariant representation in remote sensing image recognition, we perform calculation and visualization of RDM for several selected images of 10 previous categories. These images contain objects of highly complex spatial relationships and diverse visual features; thus, they are suitable for examining whether the various features of a same object or a same class generate similar activity patterns. According to different characteristics, such as object background, size, shape, and shadow (Table 4), we select 10 images for 3 feature groups, which represent the various intra-class images. By calculating and analyzing the RDM of 300 images, the response property of different visual features of identical stimulus can be obtained.

Table 4. A total of 10 categories of remote sensing images

| Object categories | Different features of object | Images example |
|---|---|---|
| airplane | different scenes and object size | 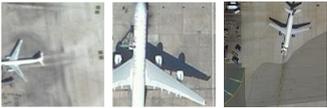 |
| city_building | different shapes, colors and shadows | 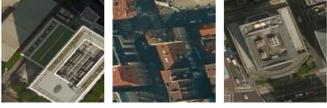 |
| storage_room | different colors, shadows and object size | 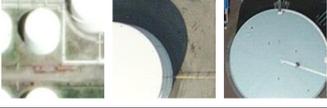 |
| parking_lot | different scenes and number of vehicles | 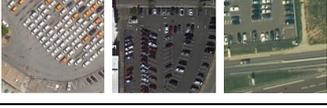 |
| resident | different colors, shadows and housing density | 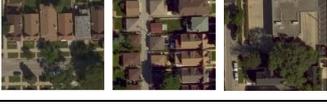 |
| **airport_runway** | different scenes | 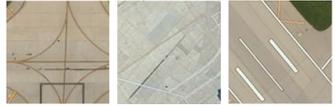 |
| **bareland** | different scenes | 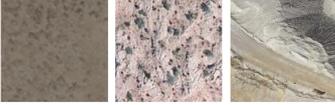 |
| **green_farmland** | different shapes and colors | 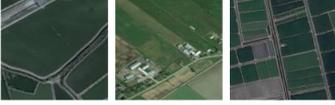 |
| **coast_line** | different scenes and shapes | 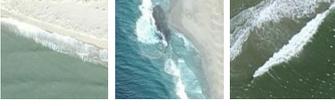 |
| **highway** | different scenes and number of vehicles | 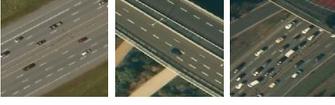 |

Table 4. Different features of the 10 categories of objects and the corresponding samples of images. A total of 30 images are selected for each category of object on the basis of the different features to explore the feature responses of identical category with different visual characteristics.

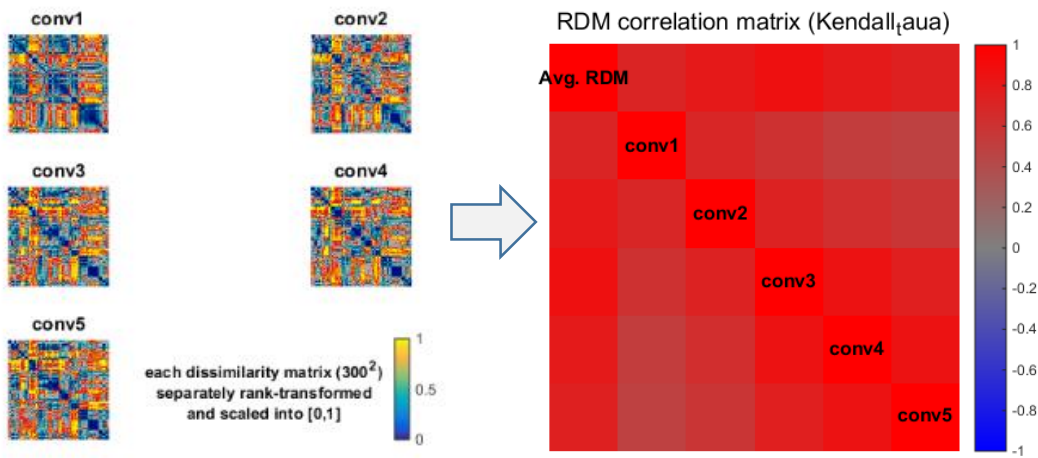

Figure 6. Left: RDM of 10 categories. Right: RDM. The figure illustrates the RDM correlation among five convolution layers. The color in the figure corresponds to the correlation coefficient.

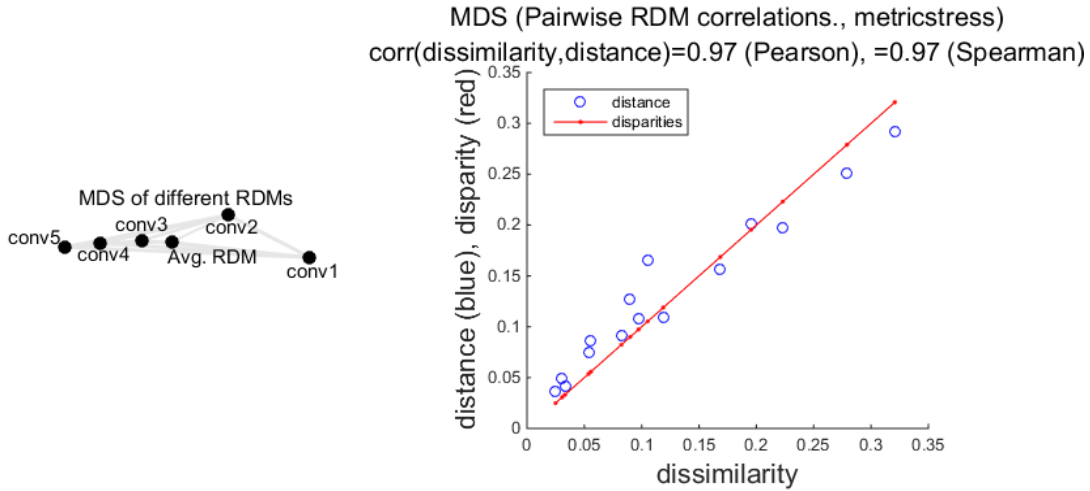

Figure 7. Relationships among five convolution layers. Left: Multi-dimensional scale analysis of different RDMs. The figure illustrates the distance of RDM between different convolution layers. Right: Correlation between the dissimilarity and distance of RDM (with Pearson's and Spearman's correlation coefficient of 0.97).

Fig. 6 illustrates the correlation coefficients between RDMs in different convolution layers visually using different colors. As shown in Fig. 6, the RDMs of conv3, conv4, and conv5 are highly similar. The left picture in Fig. 7 shows the distance between the RDMs in five convolution layers using the multi-dimensional scale analysis method. Each point represents the RDM of 10 categories in each convolution layer, and the distance between these points indicates the dissimilarity between RDMs in five convolution layers. The right picture in Fig. 7 shows that the correlation coefficient between the distance and dissimilarity of RDM is 0.97, indicating that the dissimilarity of the RDMs between the convolution layers can be expressed as the visual distance between RDMs to a large extent. The distance among conv3, conv4, and conv5 is small, which is consistent with the correlation coefficient matrix in Fig. 6. Therefore, the three convolution layers exhibit highly similar response to the images. By contrast, the distance between conv1 and conv2 is large; in particular, the gradually increasing distance between conv1 and conv3, conv4, and conv5 corresponds to the gradually decreasing correlation coefficient (right side of Fig. 6). This condition indicates that low- and high-level convolution layers present a large dissimilarity in image feature representation. Combined with the previous geometry representation visualization in Table 3, we further conclude that neurons show dissimilarity representation to different images in each convolution layer and the distinction in evident for images in high-level convolution layers.

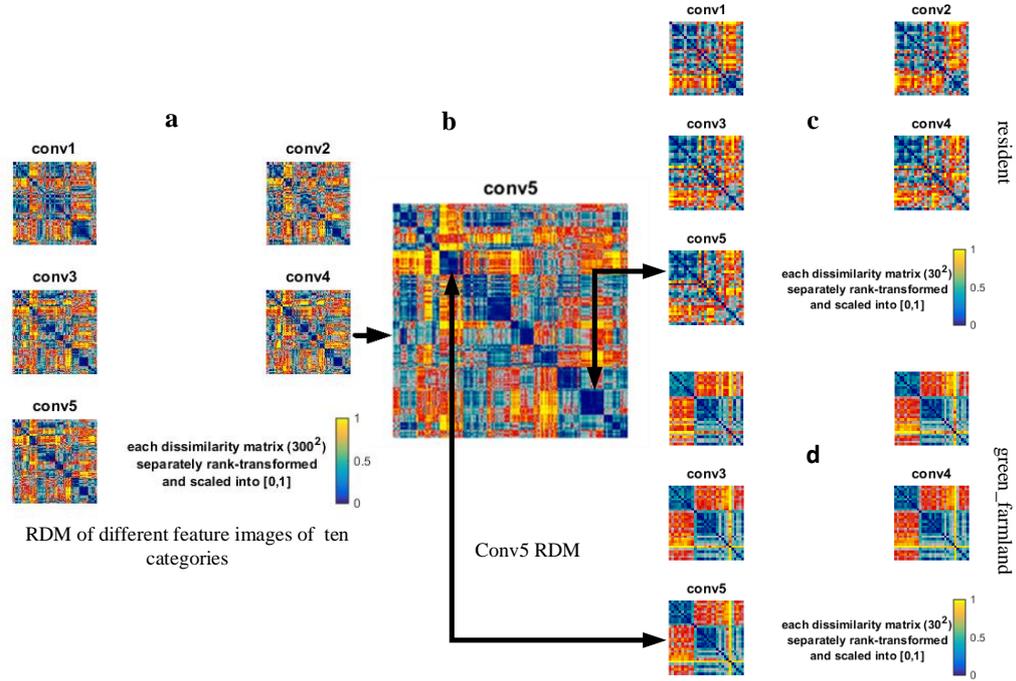

Figure 8. RDM of 10 categories and different features. (a) RDM visualization results of different feature activity patterns among 10 categories in five convolution layers. The size of each RDM is 300×300. (b) Extended RDM on conv5. (c) RDM visualization results of different feature activity patterns for resident images. The size of each RDM is 30×30. (d) RDM visualization results of different feature activity patterns for green_farmland images. The size of each RDM is 30×30. Black arrows indicate the two categories of RDM levels.

Fig. 8(a) shows the visual representation of the 300 selected images of 10 categories in each convolution layer. The RDM diagonal regions in each convolution layer are generally small (bluish) whereas off-diagonal regions are generally large (reddish and yellowish), indicating that the dissimilarity coefficient between the inter-class activity patterns is considerably greater than that between the intra-class ones. Therefore, the inter-class activity patterns are distinguishable. The distribution of the dissimilarity coefficient of inter- and intra-class activity patterns (columns 1 and 2 of Fig. 9) shows that the dissimilarity coefficients of inter-class activity patterns in each convolution layer are mostly from 0.6 to 1.2 and those of intra-class activity patterns are from 0 to 0.6. This finding proves that the network structure strongly represents image features, resulting in different activity patterns to different images. The experimental results further show that the inter-class activity patterns differ and the intra-class activity patterns are highly similar.

Fig. 8(b) illustrates the amplified RDM of conv5 shown in Fig. 8(a) and shows that the dissimilarity coefficients between intra-class activity patterns in the diagonal region of the RDM are small (diagonal bluish areas). Figs. 8(c) and (d) show the RDMs of resident and green_farmland images corresponding to different objects of diverse features, indicating that dissimilarity between a pair activity patterns also exists in intra-class images. The dissimilarity coefficients of activity patterns in green_farmland images are larger than those in resident images. The difference of activity patterns of intra-class images is generally lesser than inter-class images. As shown in columns 3 and 4 of Fig. 9, the dissimilarity coefficient of activity patterns among residents is in the range of 0–0.2 and the dissimilarity coefficient of activity patterns in green_farmland images is in the range of 0–0.6. These values correspond to the RDM visualization in Figs. 6(c) and (d). Considering that the characteristics of resident images are similar, the DCNN model produces relevant activity patterns. Although the selected green_farmland images are highly diverse with significantly distinctive activity patterns, their correlation still exists.

Figs. 8 and 9 show that the various features (e.g., object background, size, and shape) of intra-class images can induce different activity patterns. However, a strong correlation still exists between their activity patterns (despite different correlation coefficients for different categories). This result reveals that the invariant representation is significant for DCNN to gain high identification and classification performance in HRRS image recognition.

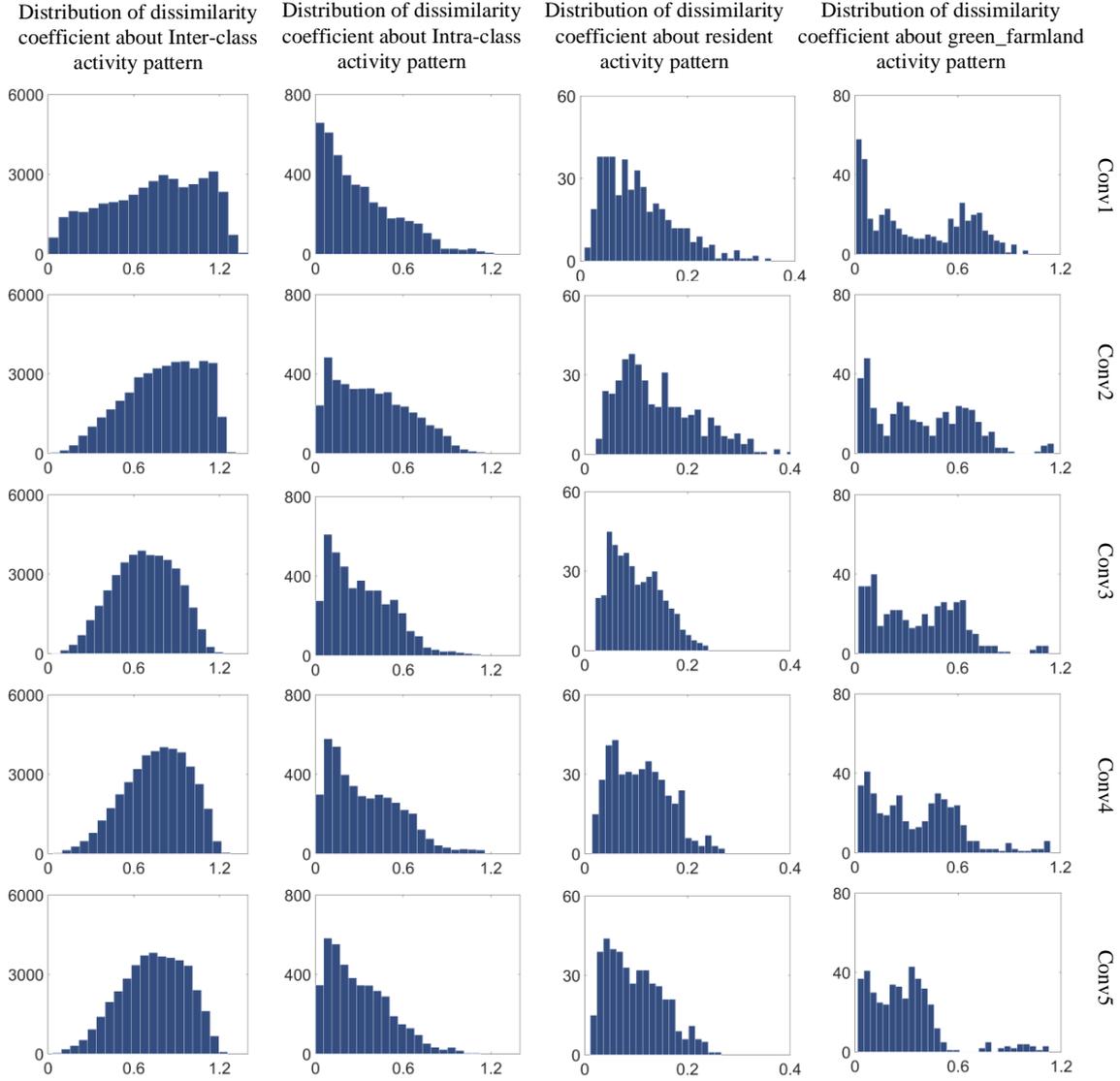

Figure 9. Distribution of dissimilarity coefficient. The first, second, third, and fourth columns show the distribution of dissimilarity coefficient of inter-class, intra-class, resident, and green_farmland activity patterns. The intra-class similarity representation and inter-class dissimilarity representation can be explained through the distribution of dissimilarity coefficient.

## 5.4 Decoding the intrinsic feature from neuron coding

We further analyze whether selectivity and invariance are embodied in the feature representation of DCNN. For this purpose, CAM is applied to visually investigate the recognition mechanism from following questions: 1) Is the feature distribution of intra-class samples invariant? 2) How does the selectivity affect the classification results? 3) How does the invariant feature perform from low level to high level in DCNN?

**A. Extracting the invariant features of intra-class images**

Despite the difference between intra-class samples, the alignment between semantic feature and visual feature provides the samples with a common feature subspace, which is attributed to the invariant semantic feature of the class. This invariance is reflected in the consistent distribution of feature response for the intra-class stimuli.

Fig. 10 shows examples of CAMs for the same data in Fig. 10(c). Notably, the strong response areas are similar for the intra-class objects in every convolution layer. For example, the heat maps of different airplane images in conv1 exhibit intensive responses to the aircraft fuselage, and the heat maps of all resident samples in conv5 only respond to the building structure. The strong responses are concentrated in some specific areas, which are the essential elements to compose the intra-class samples. As shown in Fig. 10, an intrinsic activity pattern exists for the intra-class samples in each convolution layer. Therefore, in remote sensing image recognition based on DCNN, the neurons can learn the common characteristics of intra-class samples while ignoring their different characteristics.

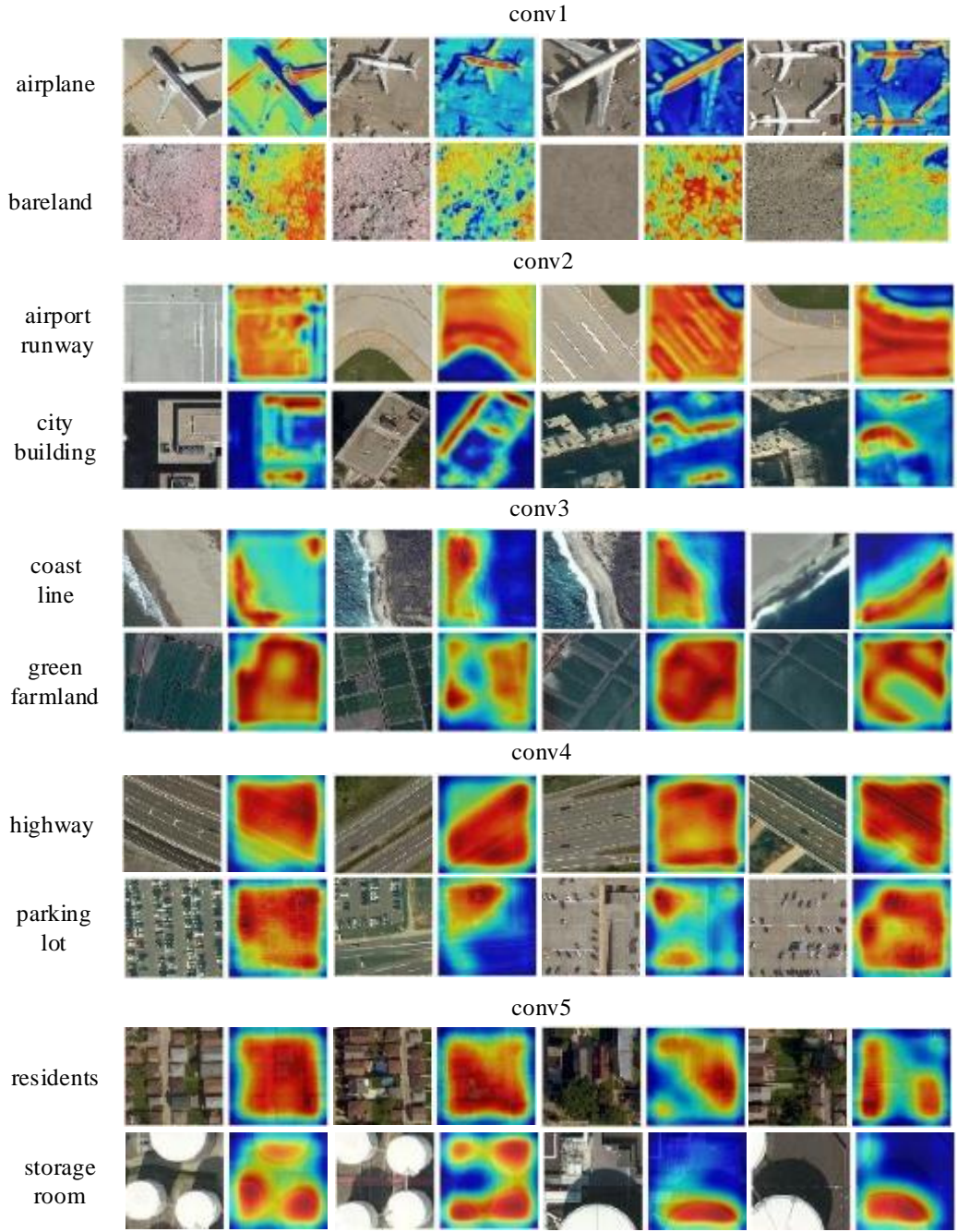

Figure 10. CAMs of 10 categories in different convolution layers. The top to bottom parts show different convolution layers corresponding to the different CAMs. The left to right parts displays the CAMs of the same category. The warmer the hue means the greater the contribution of the region to the recognition. By contrast, the colder the hue means the smaller the contribution of the region to the recognition.

**B. Determining classification results by use of local selective response**

The dissimilarity of feature distribution in different remote sensing images is reflected not only in the global features but also in the local ones. The convolutional neurons code the feature by layers to achieve the semantic attribution of stimuli, while the DCNN model of most of the ground objects especially in the high-level convolutional layer selectively respond to the partial specific regions.

Fig. 11 shows the Top-1 to Top-5 CAMs of three categories. The first row is airport_runway, and the Top-2 classification result is pipeline because its corresponding CAM intensively responds to the region covering white lines (i.e., zebra marking). Therefore, the regional features affect the final classification results to a large extent. Corresponding to the pipeline in the second row, the Top-1 CAM strongly responds to the linear texture area, reflecting its similarity with airport_runway. The fourth row shows the same airport runway sample without the linear texture feature. In this case, its Top-2 classification is no longer the pipeline, further indicating the impact of the regional feature on the classification results. The third row shows the parking_lot of Top-1 CAM responding to car areas, indicating that the process of understanding parking_lot is the same as abstracting the concept of car. The Top-2 result is shown as cityroad, because the model focuses on the local road areas of the cityroad. The above-mentioned analysis of the specific regional response strongly supports that the selective representation of neurons plays a significant role in remote sensing image recognition and classification task based on DCNN regardless of the classification results. In other words, the feature learning of DCNN strongly emphasizes the selective response on the crucial elements of an object that varies for each individual.

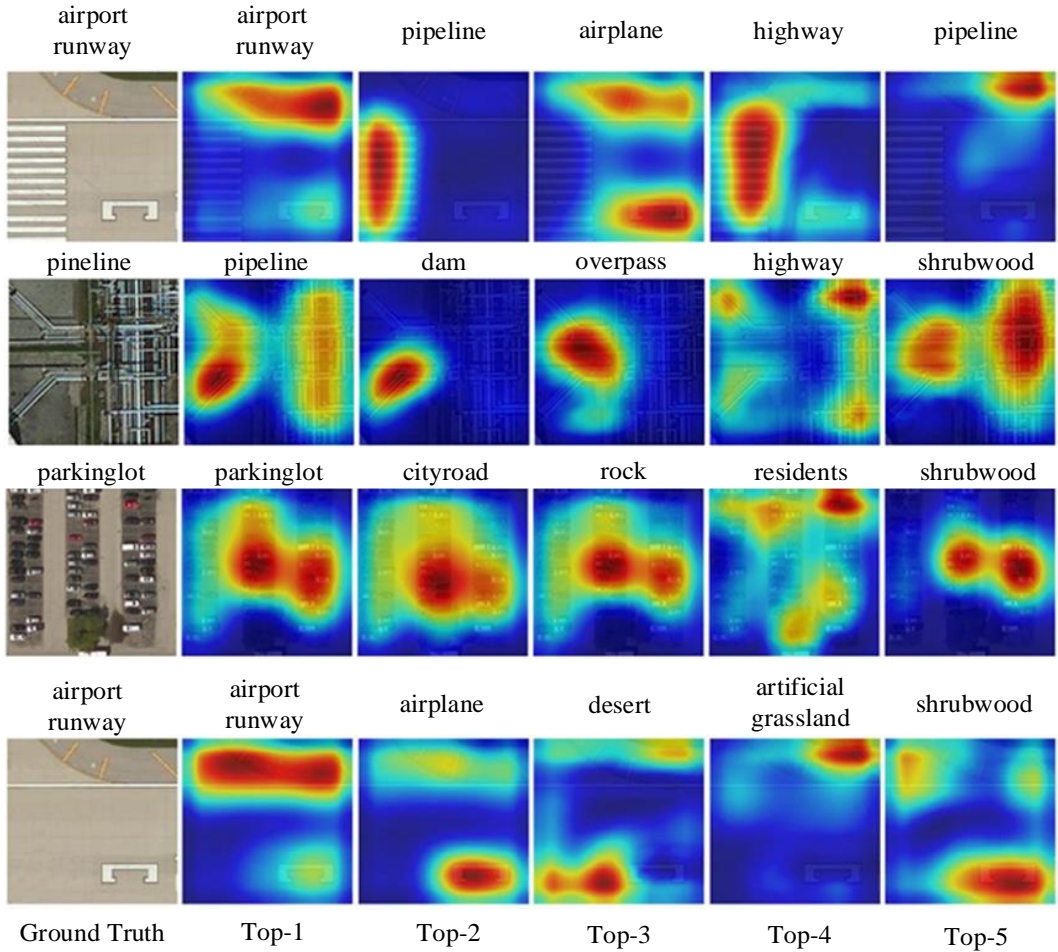

Figure 11. Model for studying the key features of CAMs

## C. Strong selective coding in high-level convolutional layer

Analyzing the learned characteristics in different convolutional layers shows that, although the responsive distributions slightly differ, the neurons in each layer can respond to the specific common region of intra-class images. Fig. 12 shows the visualization of the responsive distribution to the same objects in each layer. For example, all layers exhibit strong response to the fuselage of airplane or the building structures of resident, which indicates that all layers have learned the key characteristics of different objects. Furthermore, the feature response distribution in lower layer is divergent, and the intensity of most responsive areas is weak. On the contrary, the responses become increasingly concentrative on specific areas. Therefore, the hierarchical feature representation across convolution layers retains a notable invariant property.

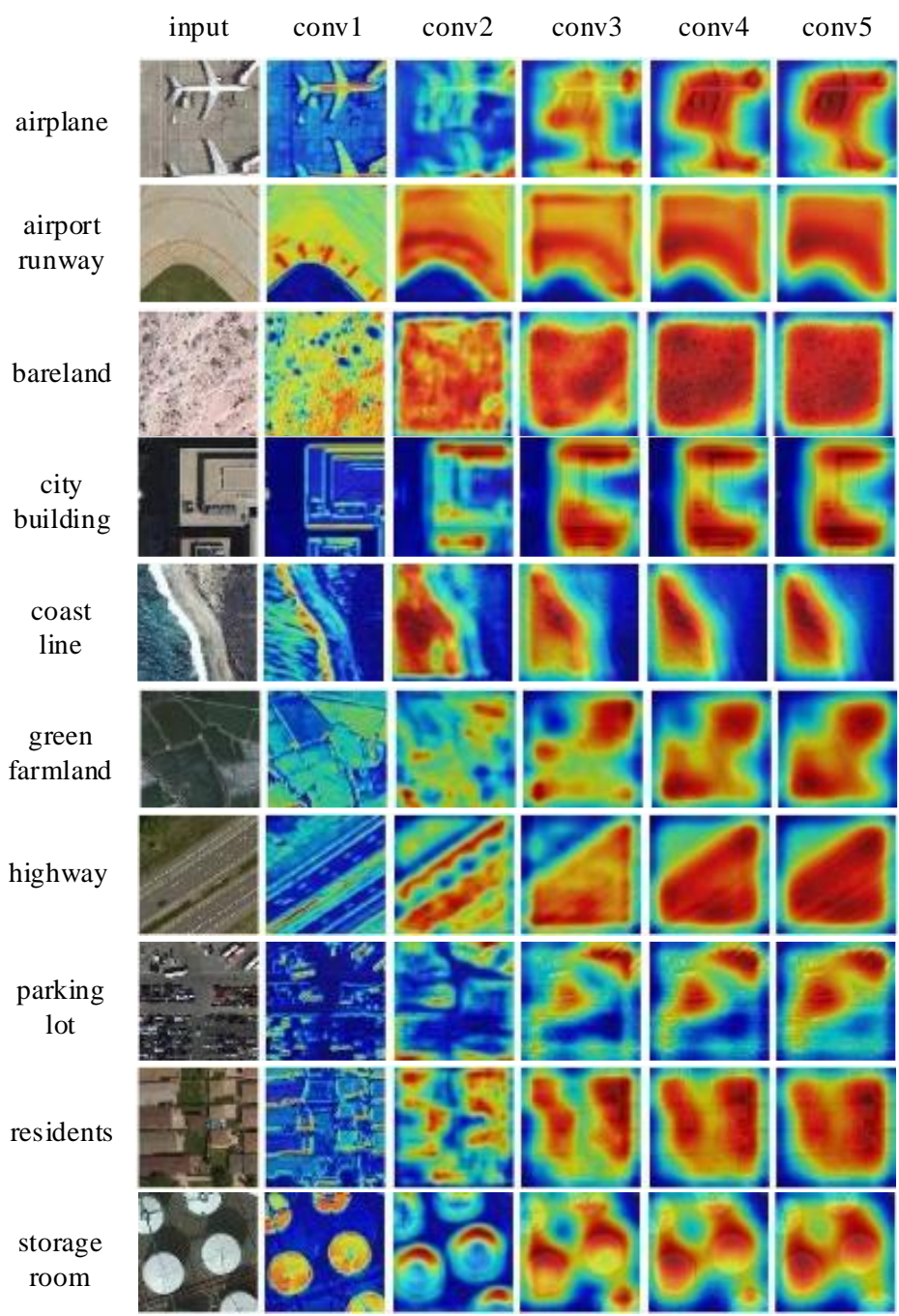

Figure 12. CAMs of different layers to dissimilar categories. The horizontal part displays the CAMs of different convolution layers, and the vertical part displays the CAMs of different categories.

The high-level feature representation maps the intra-class samples to the same semantic space by selectively coding the low-level feature; as a result, the output of the upper layer contains much semantic information. The CAMs of different layers shown in Fig. 13 indicate that the misidentification in the lower layers for the intra-class samples can be correctly identified in the upper layers. Therefore, the high-level feature representation possesses a stronger abstractive capability than the low-level one and can thus improve the recognition accuracy and the model robustness. As long as a lower layer achieves a correct identification for a stimulus, the output of the upper layer can also be correctly obtained. This finding shows that a shallow network with reduced computing cost is potentially feasible for some categories.

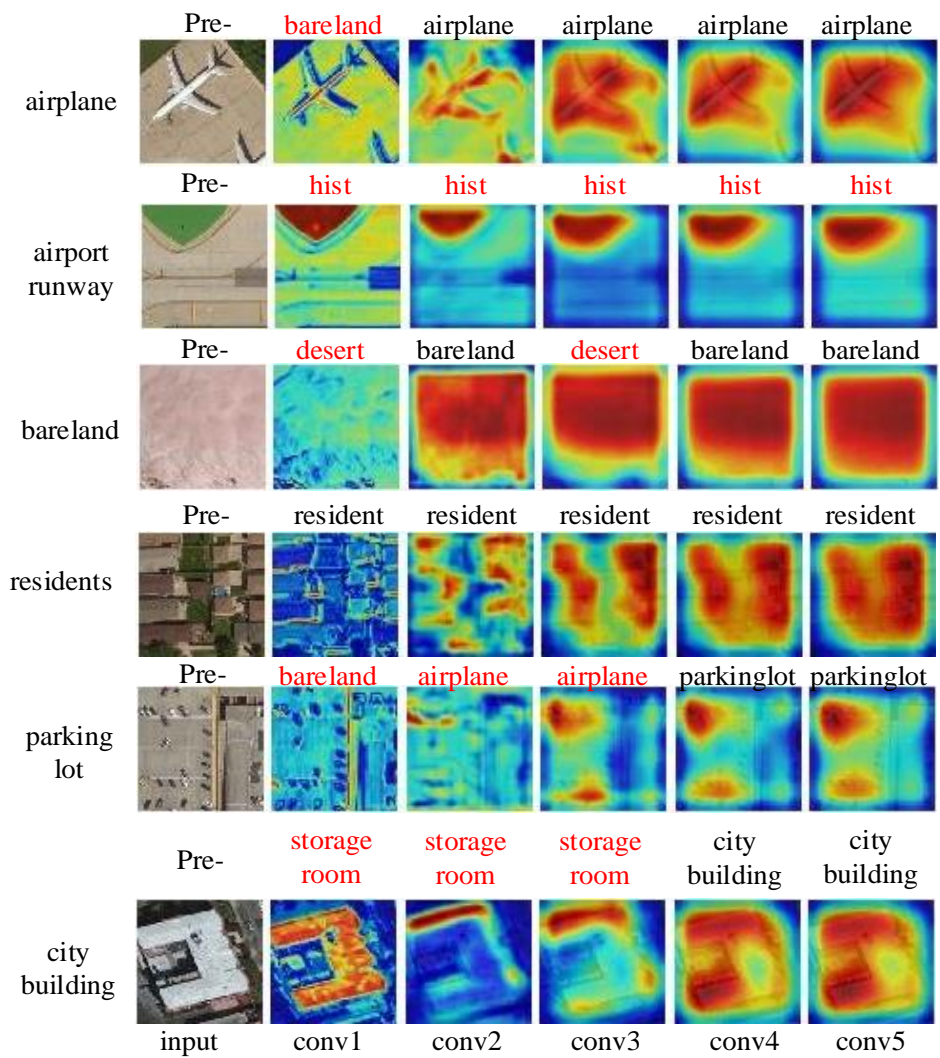

Figure 13. Visualization of low- to high-level feature coding on the impact of classification results. The top to bottom parts show the different categories. The left to right parts display the CAMs of different convolution layers. The top of each picture shows the predicted results in different layers of the model. The red font indicates the incorrect recognition, and the black font indicates the correct recognition. For example, airplane is mistakenly recognized as bareland in conv1 but correctly recognized in conv2–conv5.

## 5.5 Significant

The remote sensing domain relies heavily on the accurate and rapid identification of ground objects. However, diverse objects can integrally contain numerous images due to the variation in the position, scale, pose, and illumination of objects and the presence of visual clutter. The encoding mechanism of neurons for different percepts is a major topic of discussion in remote sensing image interpretation. This study analyzes and understands the neuronal response behavior in remote sensing image recognition task by providing a fundamental cognition. This work can be used in learning transformations by considering the mechanisms similar to those performed in the human visual ventral stream. Further considerations can also be proposed, such as what enables a student DNN to learn from the internal representational spaces of a reference model and why generative models are capable of generating data which can compensate a large number of images to the shortage of real images.

## 6. Conclusion and future works

DCNN performs comparable to human vision in image recognition and can thus accomplish recognition task in natural and remote sensing images. Ascertaining the fundamental principles of DCNN is an important task. This study is inspired by the works concerning the cognitive mechanism of visual cortex and proposes an investigative framework of "visual stimulation–characteristic response combined with feature coding–classification decoding." On the basis of a large-scale remote sensing image labeling database and the classic deep neural network AlexNet, we ascertain the selective and invariant properties in neuronal feature representation by combining statistical calculation and visual analysis. The study conclusions are valuable for understanding the intrinsic properties of deep convolution network in remote sensing image recognition.

From the results, we observe that the DCNN neurons present a strong selectivity for visual stimulation of remote sensing images. These neurons exhibit varying degrees of activation or inhibition depending on the different categories of remote sensing images and the different characteristics of the same categories. The statistics conducted on a large number of samples reveal that the activation only occurs on partial neurons in each convolution layer. In other words, a small part of the neurons effectively activate most images of a same category. Therefore, DCNN exhibits sparseness property in stimulus response for remote sensing image recognition. RDM is used to calculate and visualize the dissimilarity of activity patterns between the intra- and inter-class images in each convolution layer to further understand the selective response regularity of neurons between different classes and images. The lower layer represents the low-level features and the upper layer represents the high-level features. Notably, the selective representation is evident in the upper layer. The selective response is also reflected in CAM. The sparse response of the neurons guarantees intra-class consistency and inter-class dissimilarity, which provide further understanding on the selective representation of neurons in remote sensing image recognition.

The DCNN neurons can still accomplish objects recognition tasks efficiently and accurately in case of different shapes, sizes, and shadows of objects. In this study, we use a strategy of "feature encoding–response decoding" to carry out the experimental analysis. The numerical calculation and visualization of RDM show that object features (e.g., object background, size, and shape) differ among intra-class images, but neurons can generate similar encoding representations for identical objects. Meanwhile, the classification and visualization of CAM reveal that feature differences exist in intra-class samples. However, the low-level visual characteristics and high-level semantics of these samples are highly consistent, thereby addressing the so-called "semantic gap" in the computer vision domain. The DCNN neurons encode the different images of a same class into the common feature subspace, thereby resulting in consistency of intra-class responsive distribution and invariant semantic representation.

Future studies can extend this work by adopting new measurement methods to analyze and reveal the representation property of the neural network for remote sensing image recognition and utilizing the neuronal representation to guide the transfer learning and generative learning, as the two aspects have attracted much interest in the remote sensing domain.

# 7. Appendix

Table 5. In each diagram, the abscissa indicates the serial number of neuron, and the ordinate indicates the effective responses of the neuron to the object.

**Table 5.** Neuronal stimulus-responses

**Table 5-Conv1**

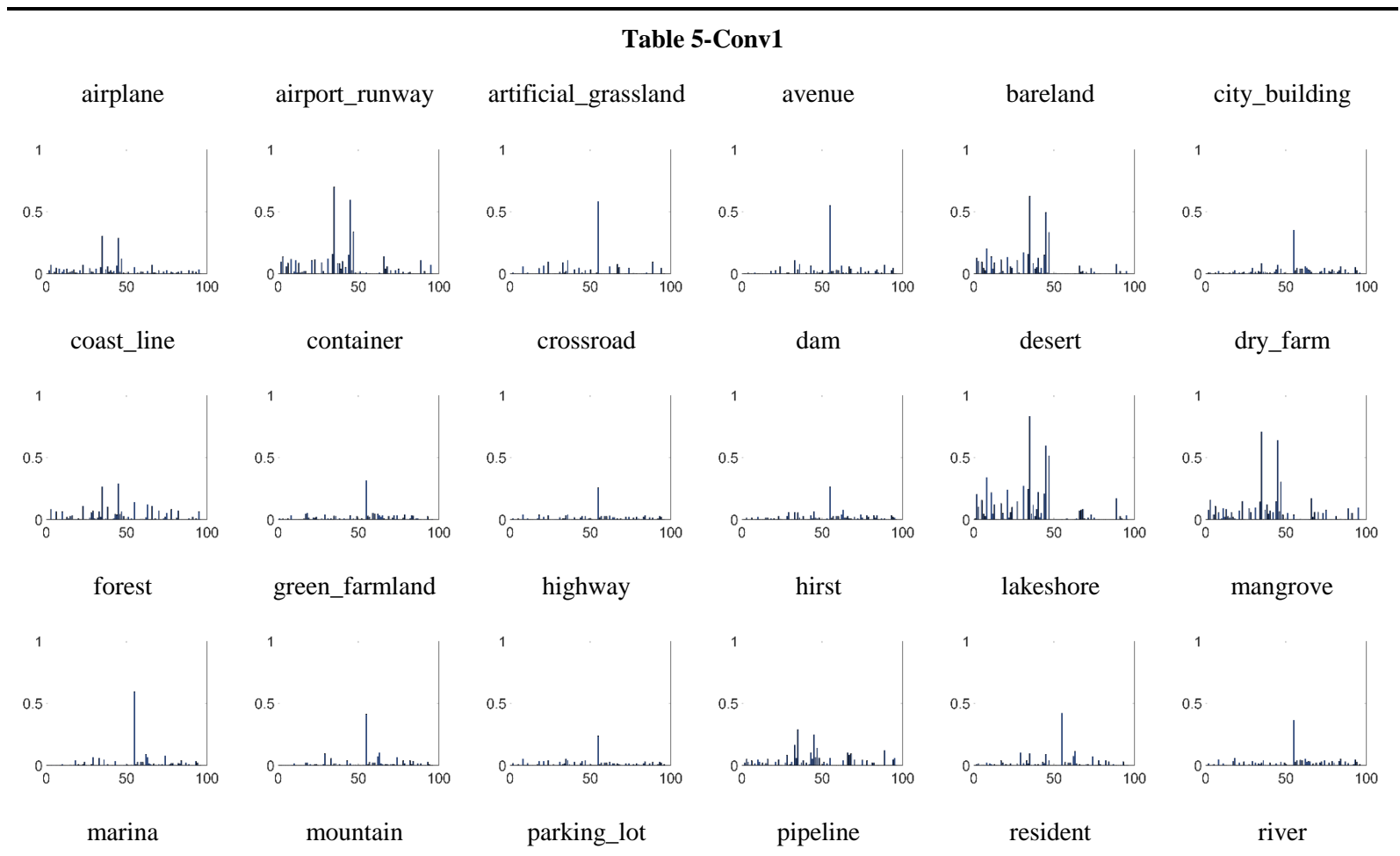

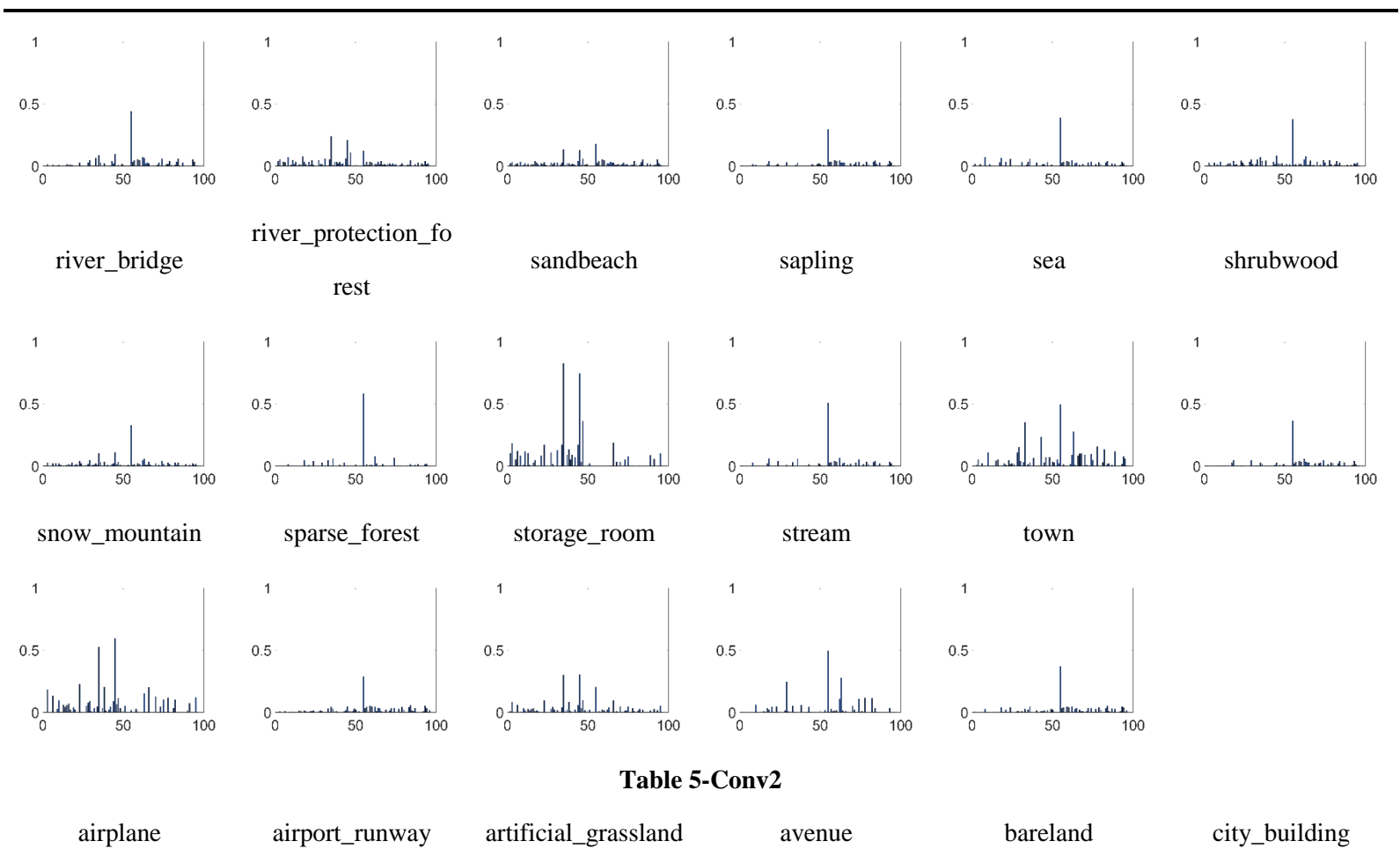

Table 5-Conv2

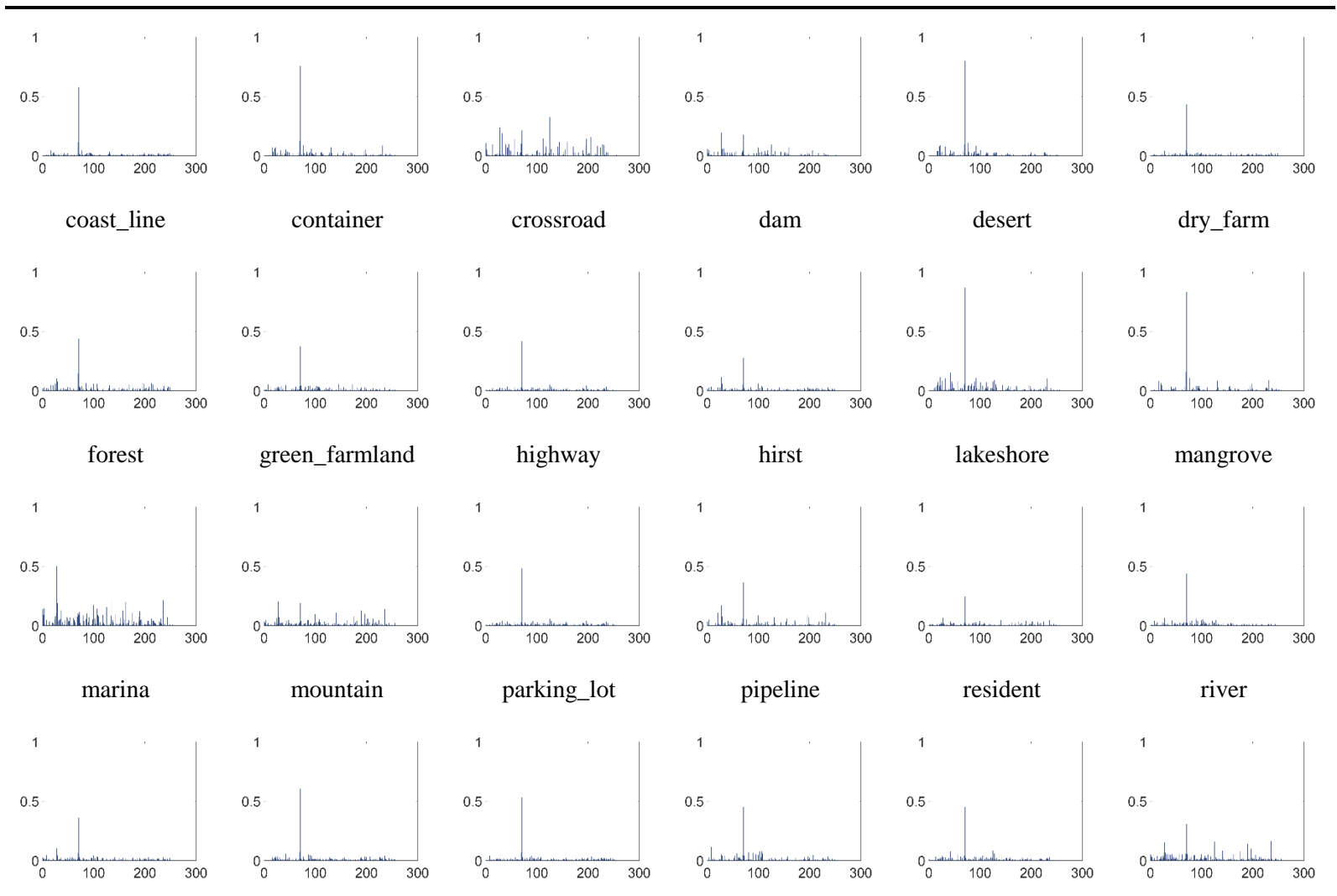

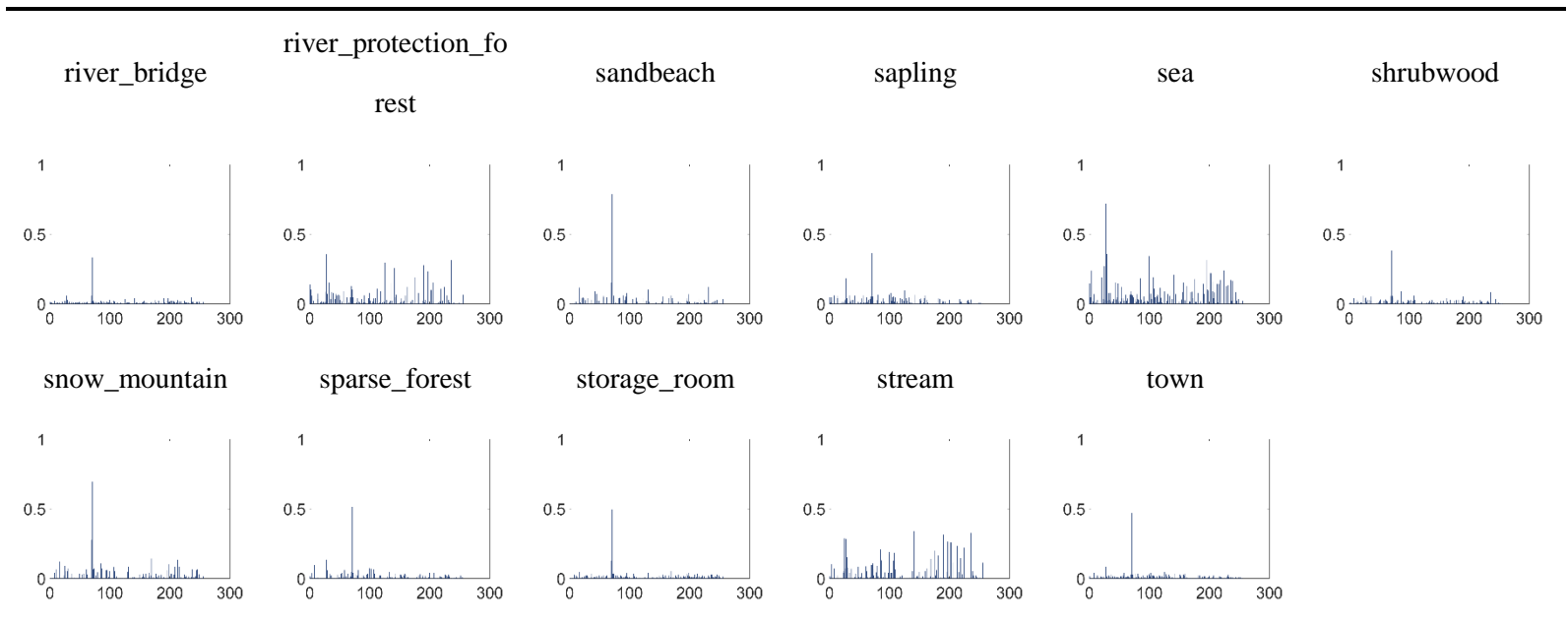

**Table 5-Conv3**

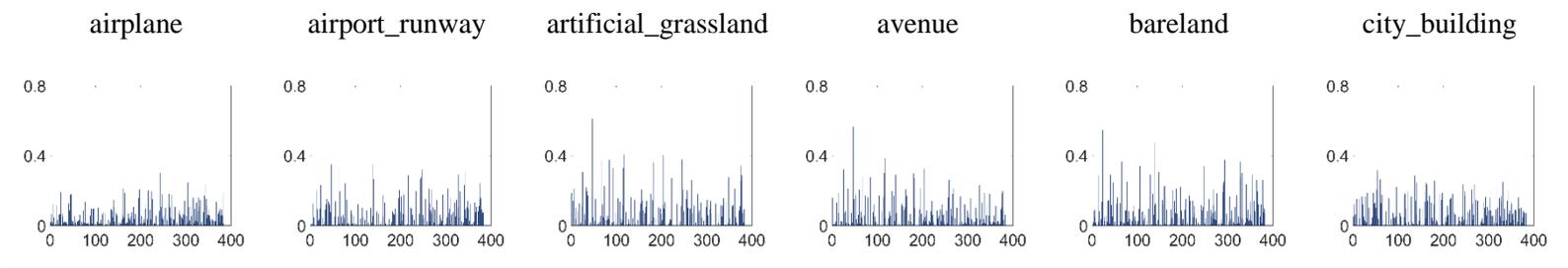

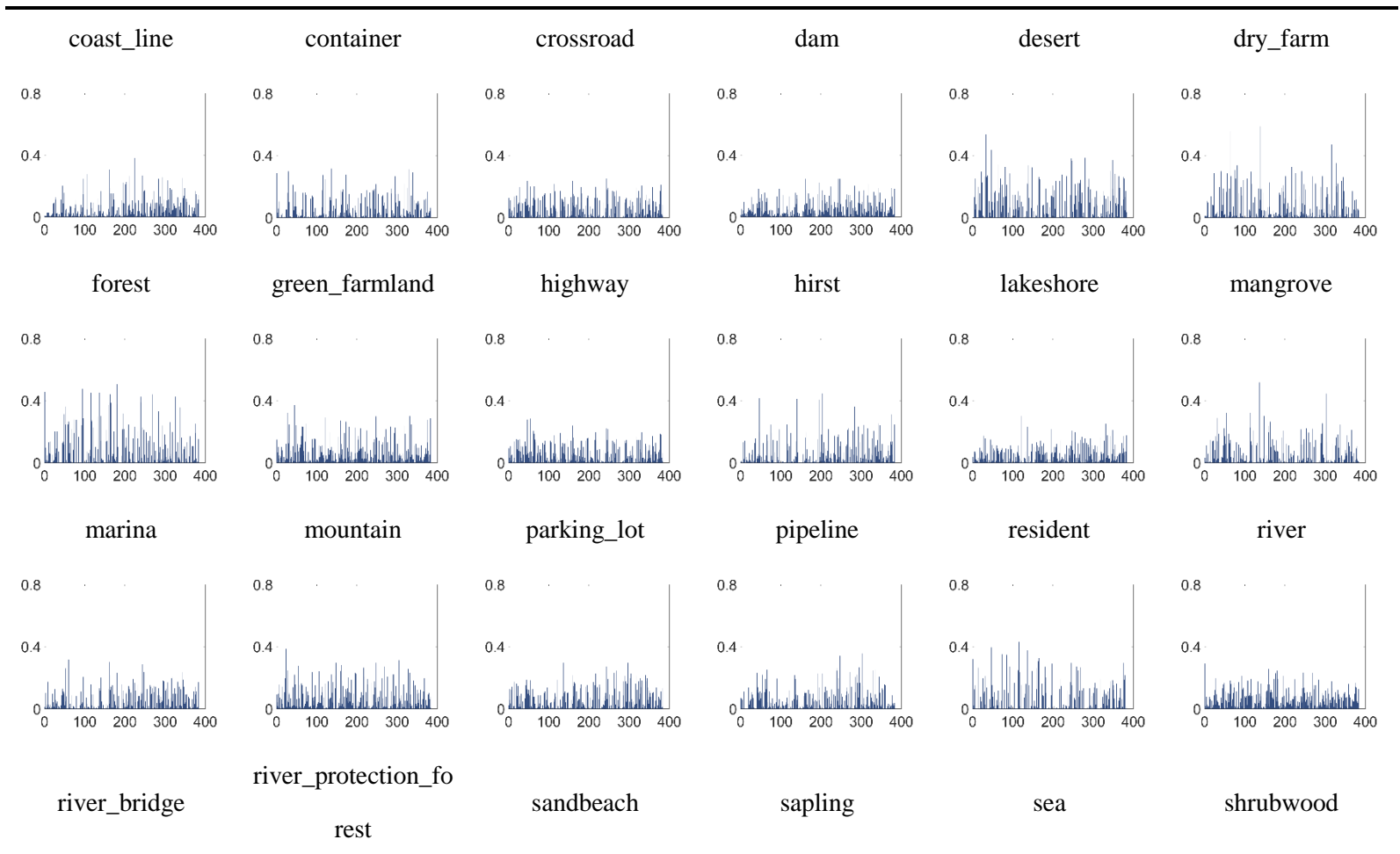

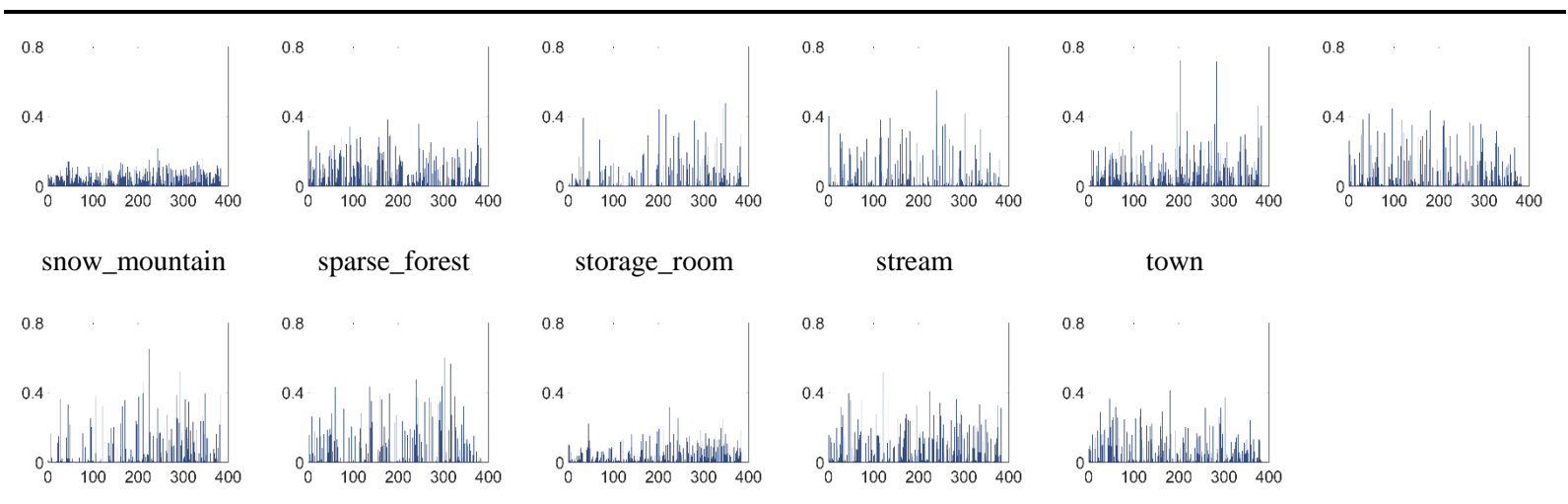

snow_mountain | sparse_forest | storage_room | stream | town

**Table 5-Conv4**

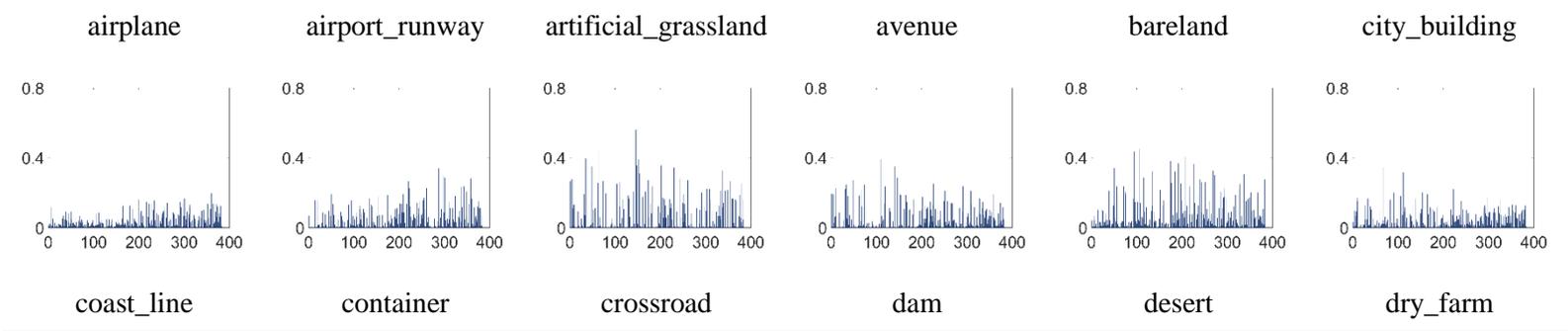

airplane | airport_runway | artificial_grassland | avenue | bareland | city_building

coast_line | container | crossroad | dam | desert | dry_farm

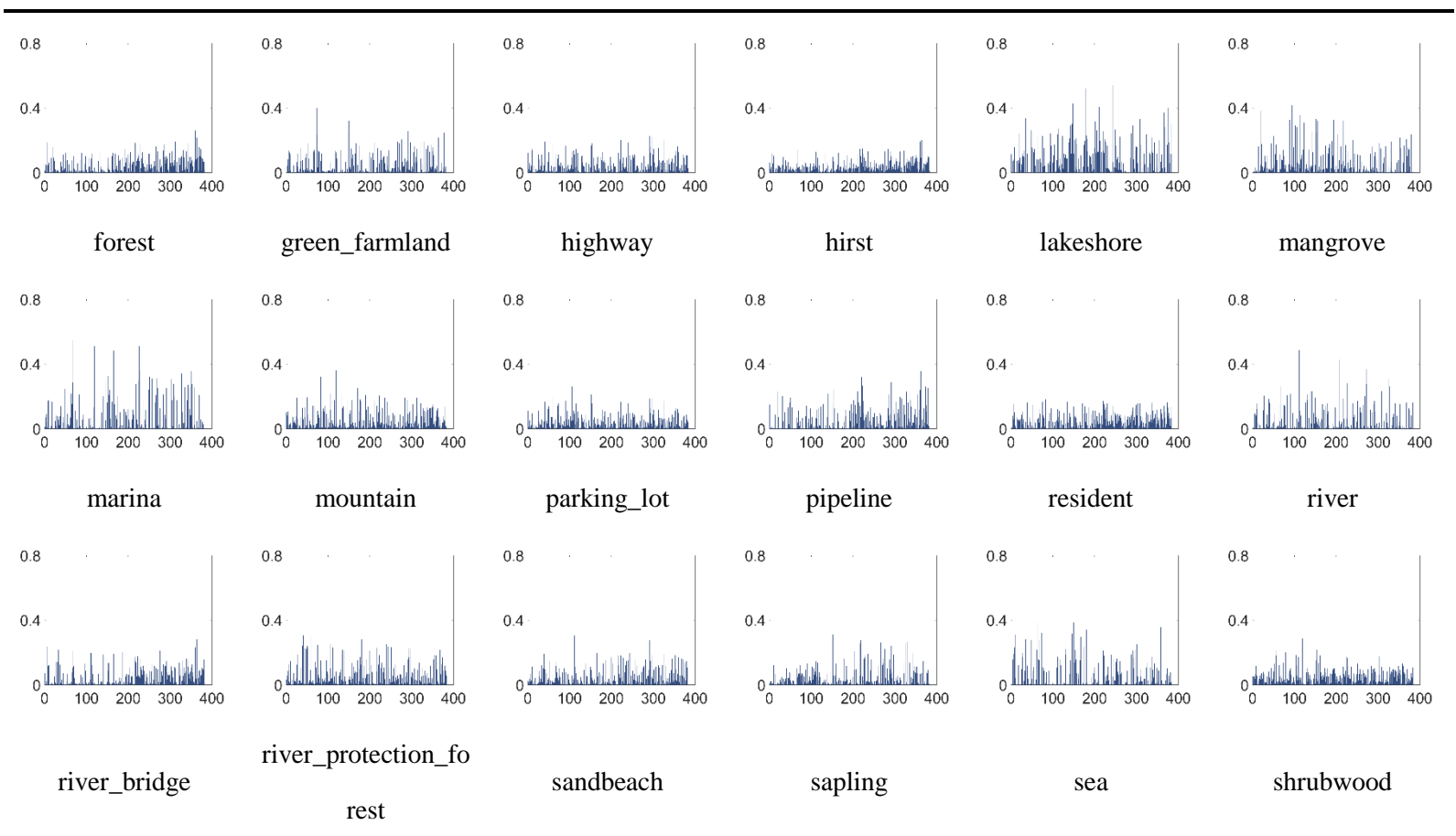

| forest | green_farmland | highway | hirst | lakeshore | mangrove |

| marina | mountain | parking_lot | pipeline | resident | river |

| river_bridge | river_protection_forest | sandbeach | sapling | sea | shrubwood |

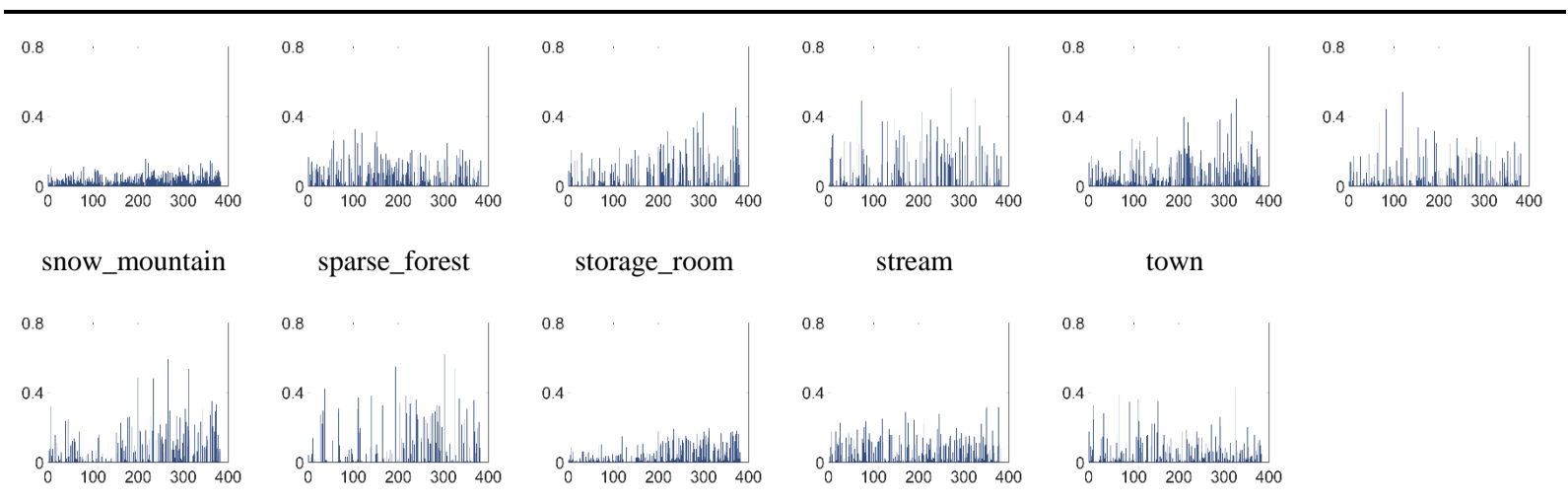

| snow_mountain | sparse_forest | storage_room | stream | town | |

**Table 5-Conv5**

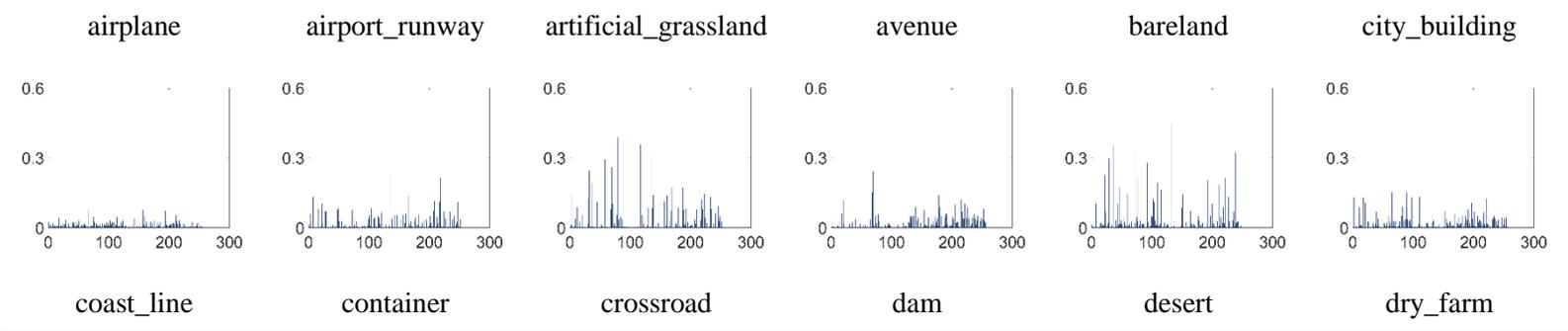

| airplane | airport_runway | artificial_grassland | avenue | bareland | city_building |
| coast_line | container | crossroad | dam | desert | dry_farm |

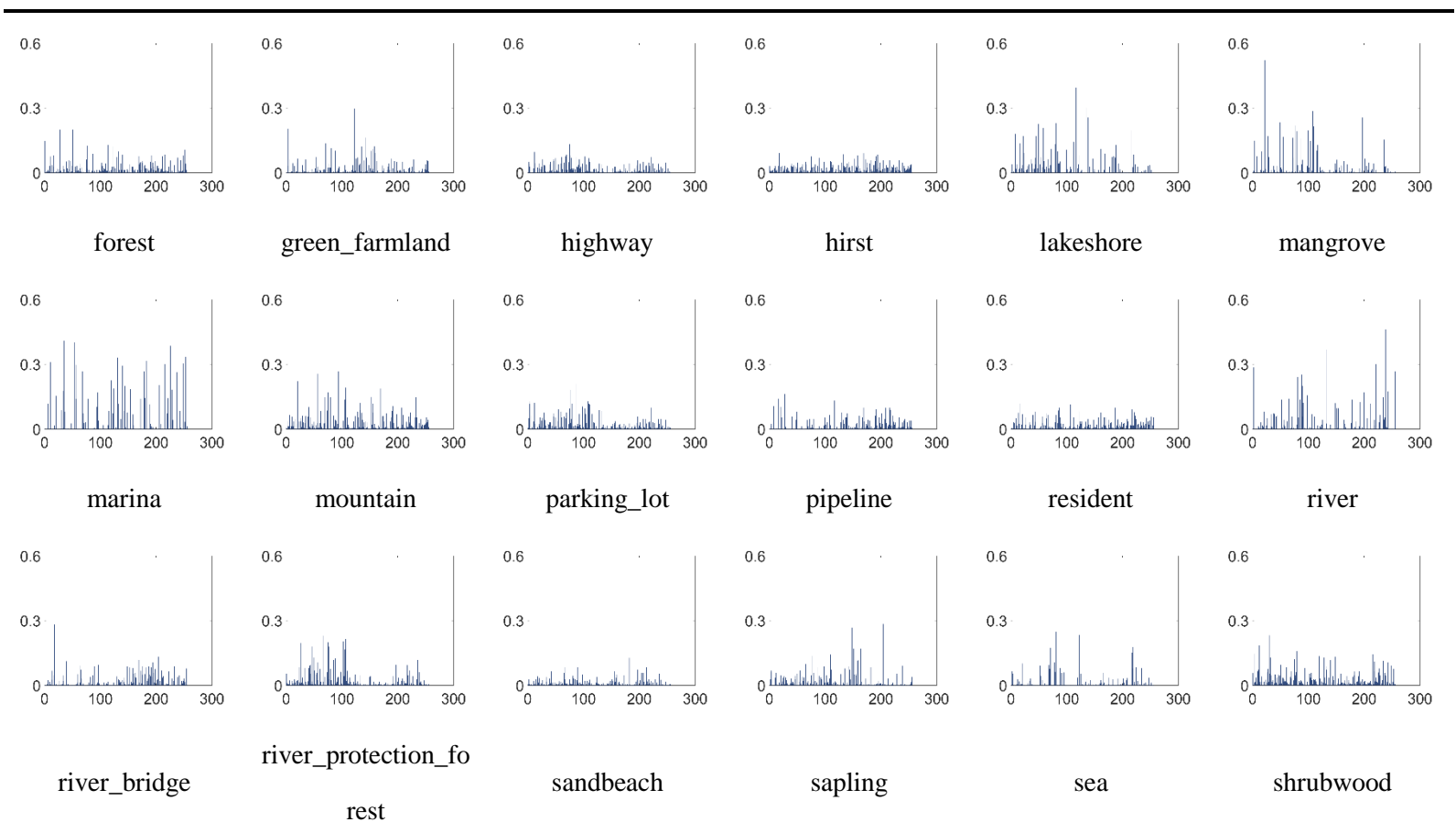

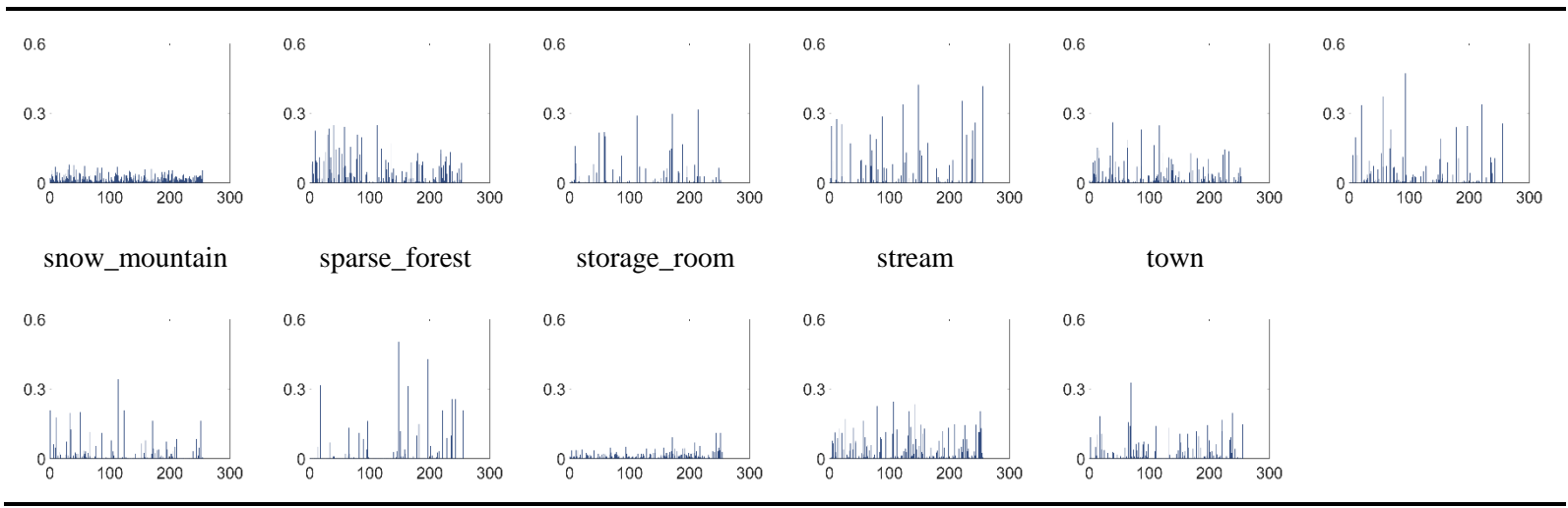

|           |              |                    |           |            |               |
|-----------|--------------|--------------------|-----------|------------|---------------|
| snow_mountain | sparse_forest | storage_room | stream | town | |

Table 6. The histogram of activated neurons number on each image. The abscissa in the graph indicates the activated neurons number on each image, and the ordinate shows the number of images.

**Table 6-Conv1**

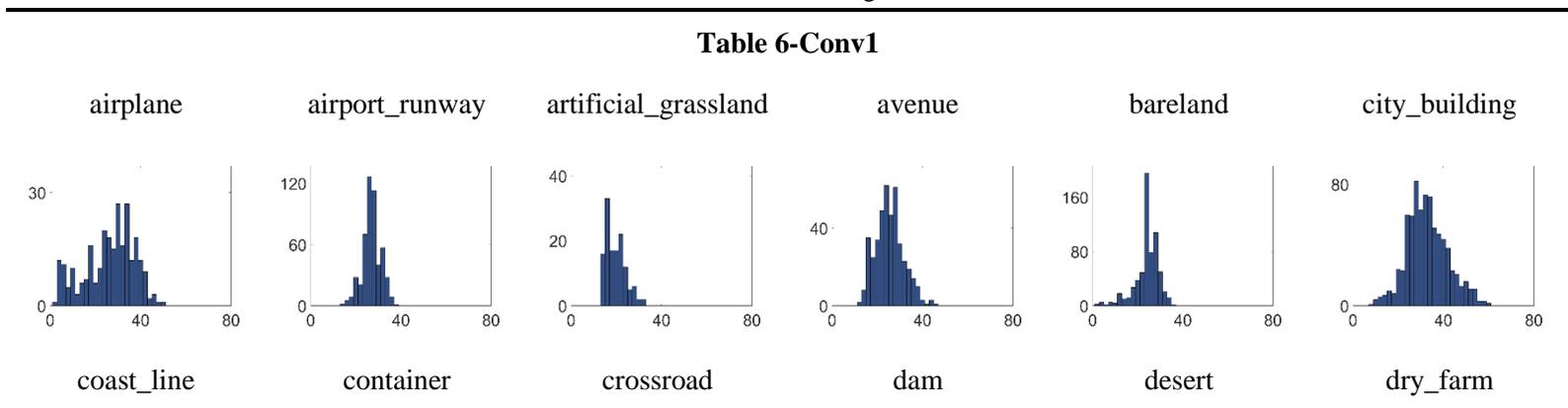

| airplane | airport_runway | artificial_grassland | avenue | bareland | city_building |
|----------|----------------|----------------------|--------|----------|---------------|
| coast_line | container | crossroad | dam | desert | dry_farm |

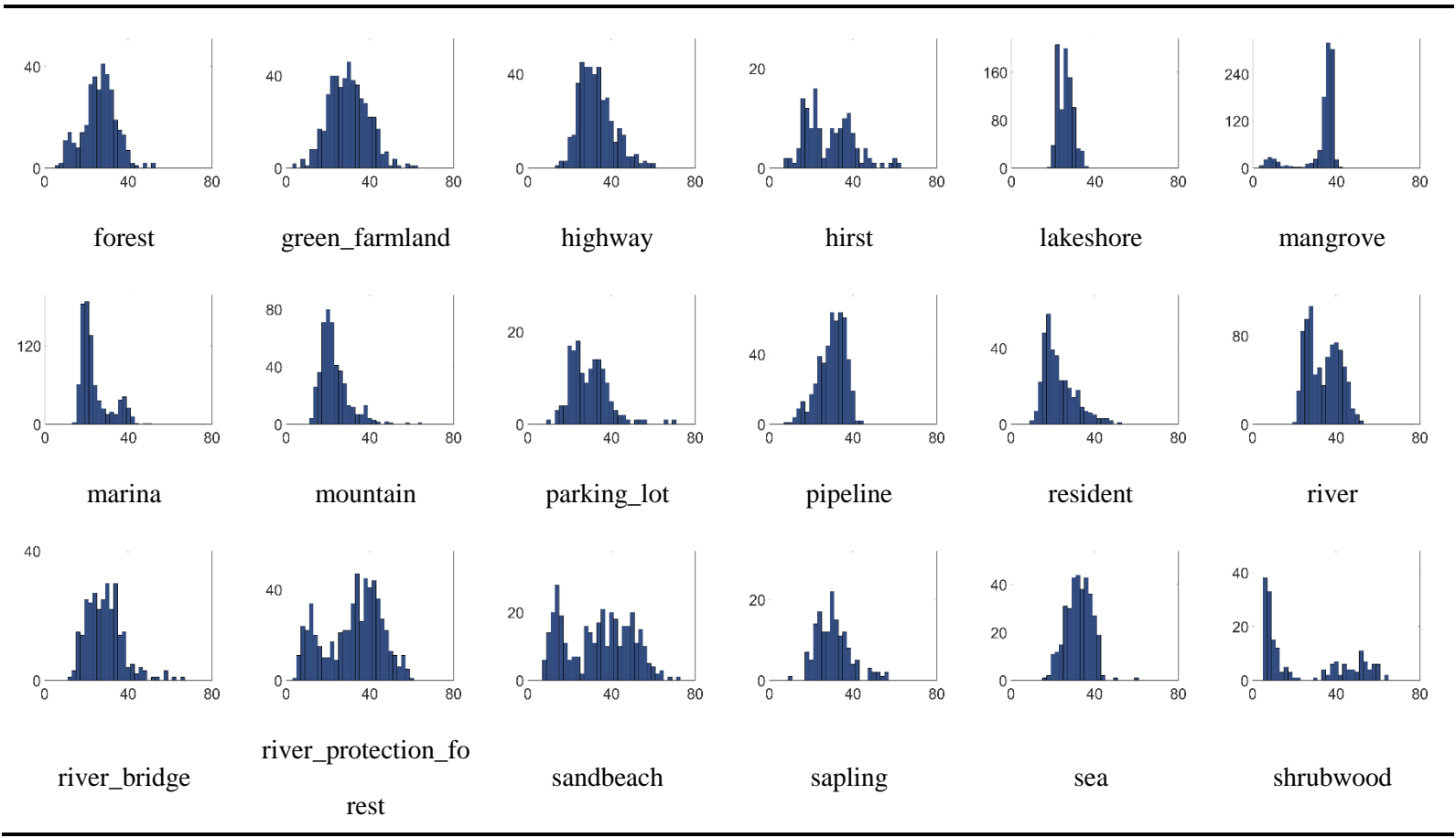

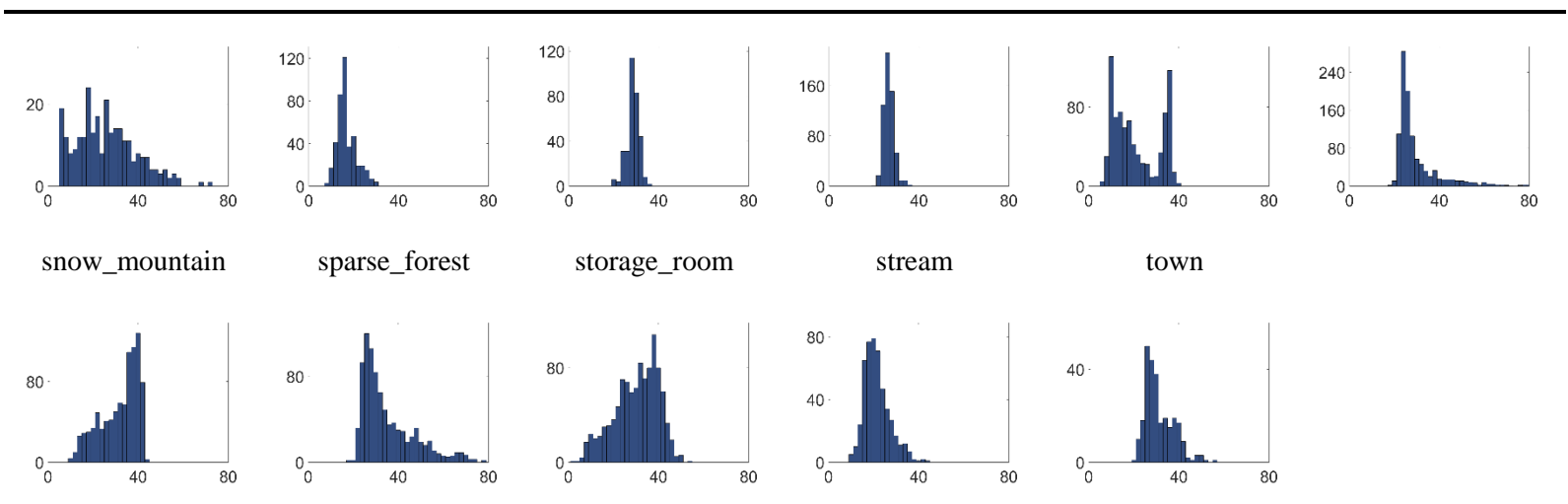

**Table 6-Conv2**

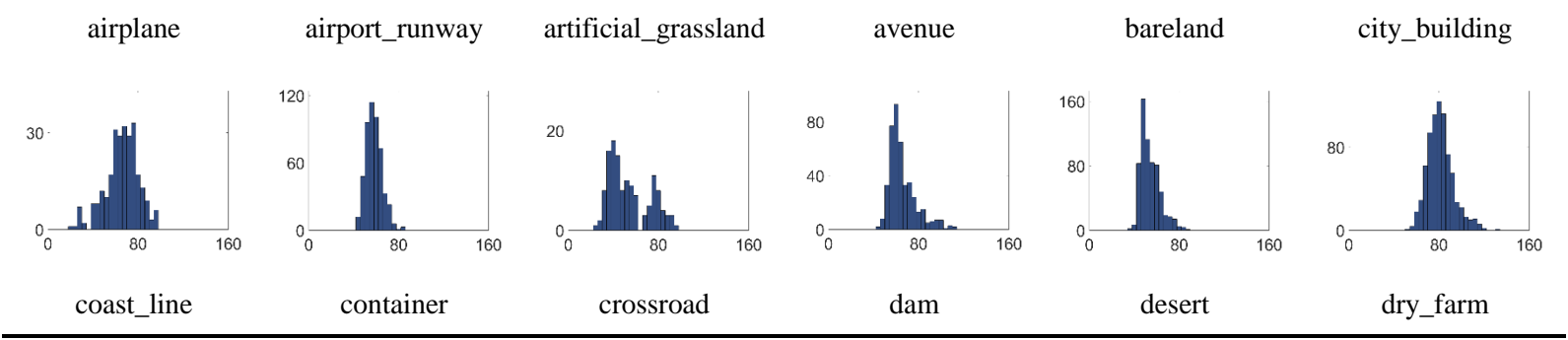

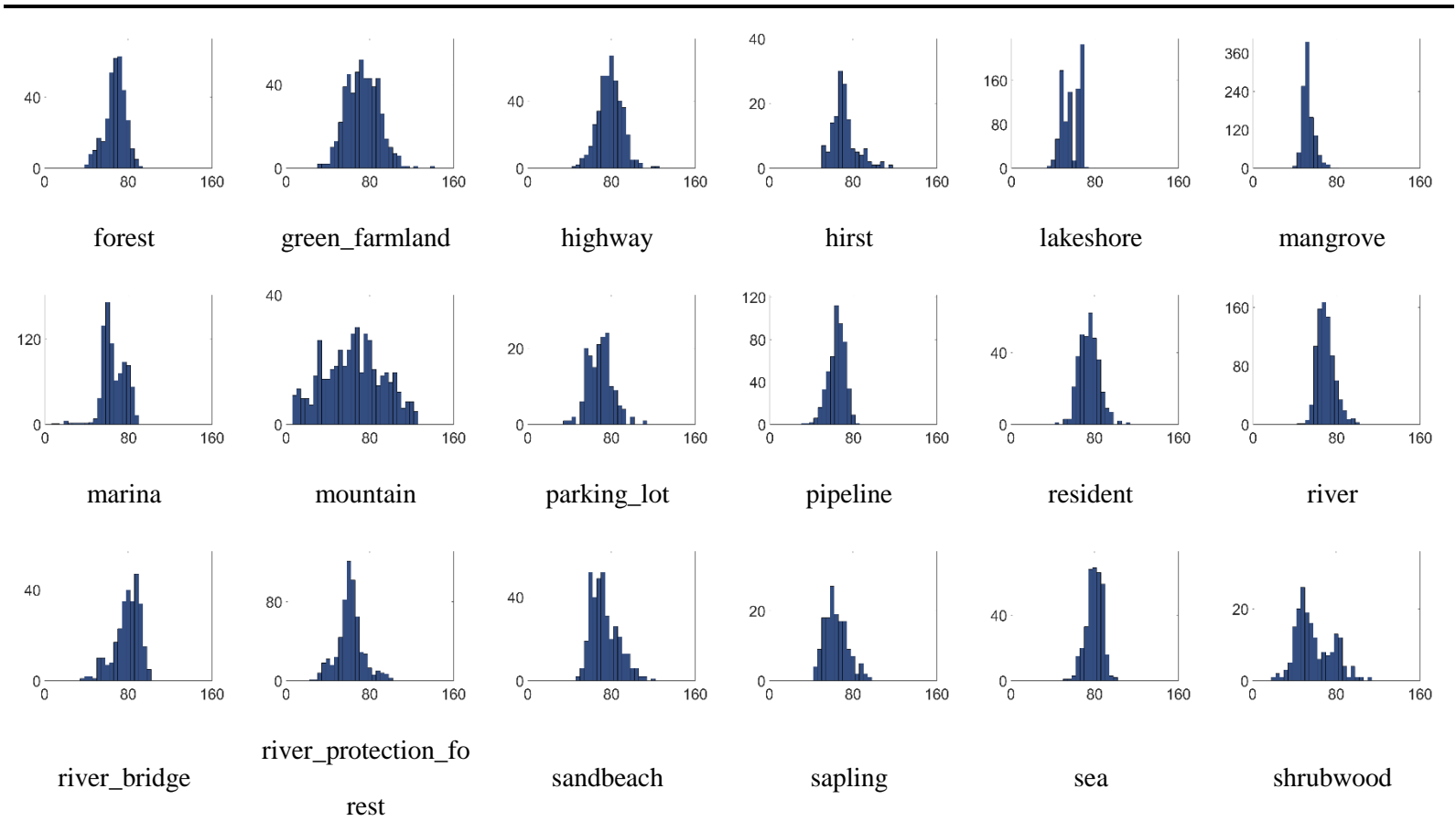

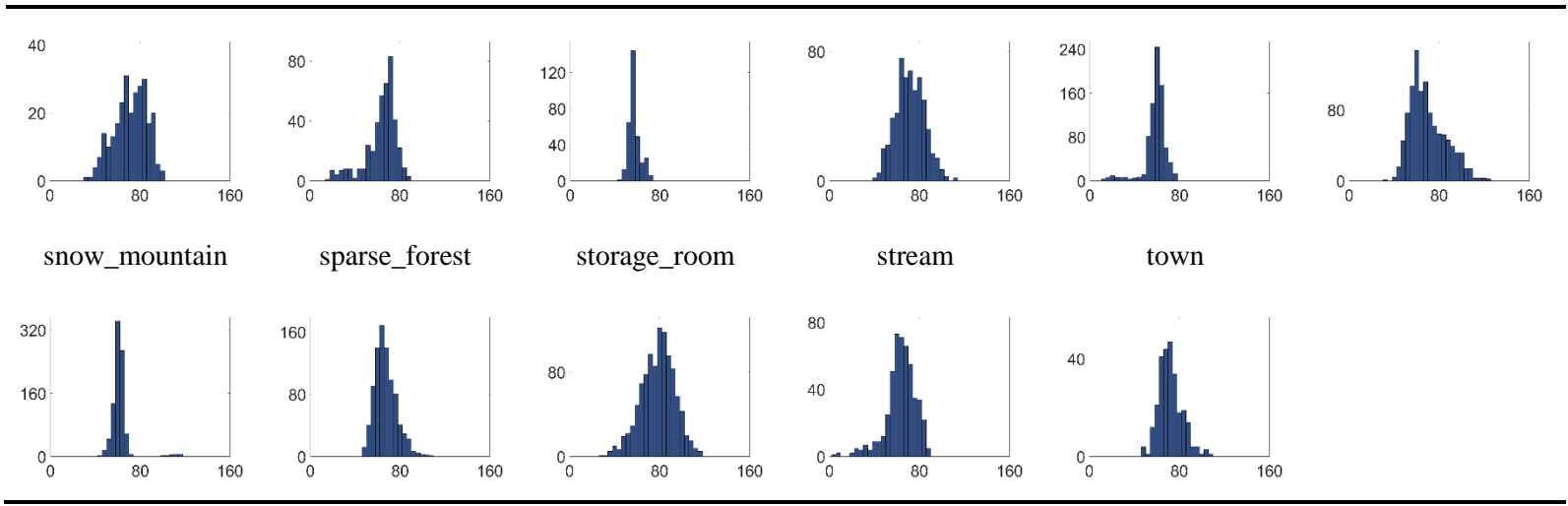

| snow_mountain | sparse_forest | storage_room | stream | town | |

### Table 6-Conv3

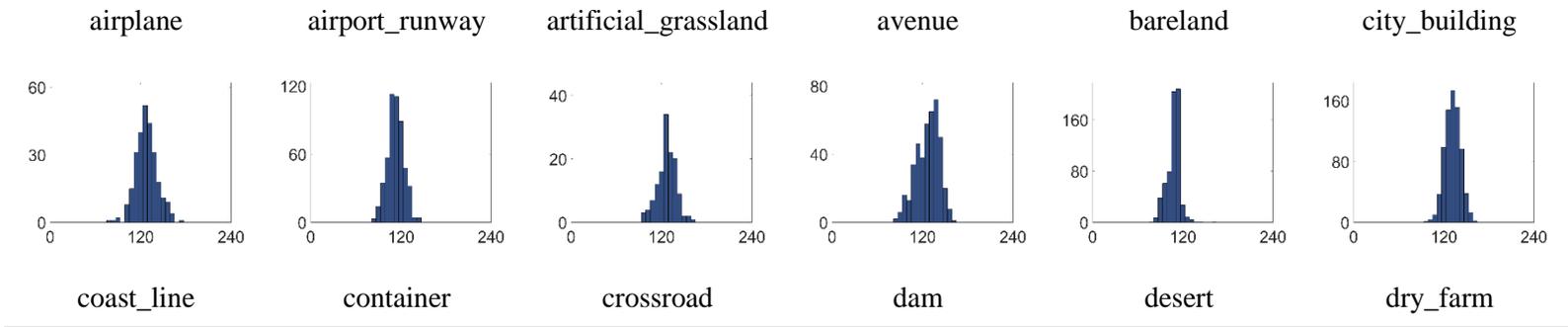

| airplane | airport_runway | artificial_grassland | avenue | bareland | city_building |
| coast_line | container | crossroad | dam | desert | dry_farm |

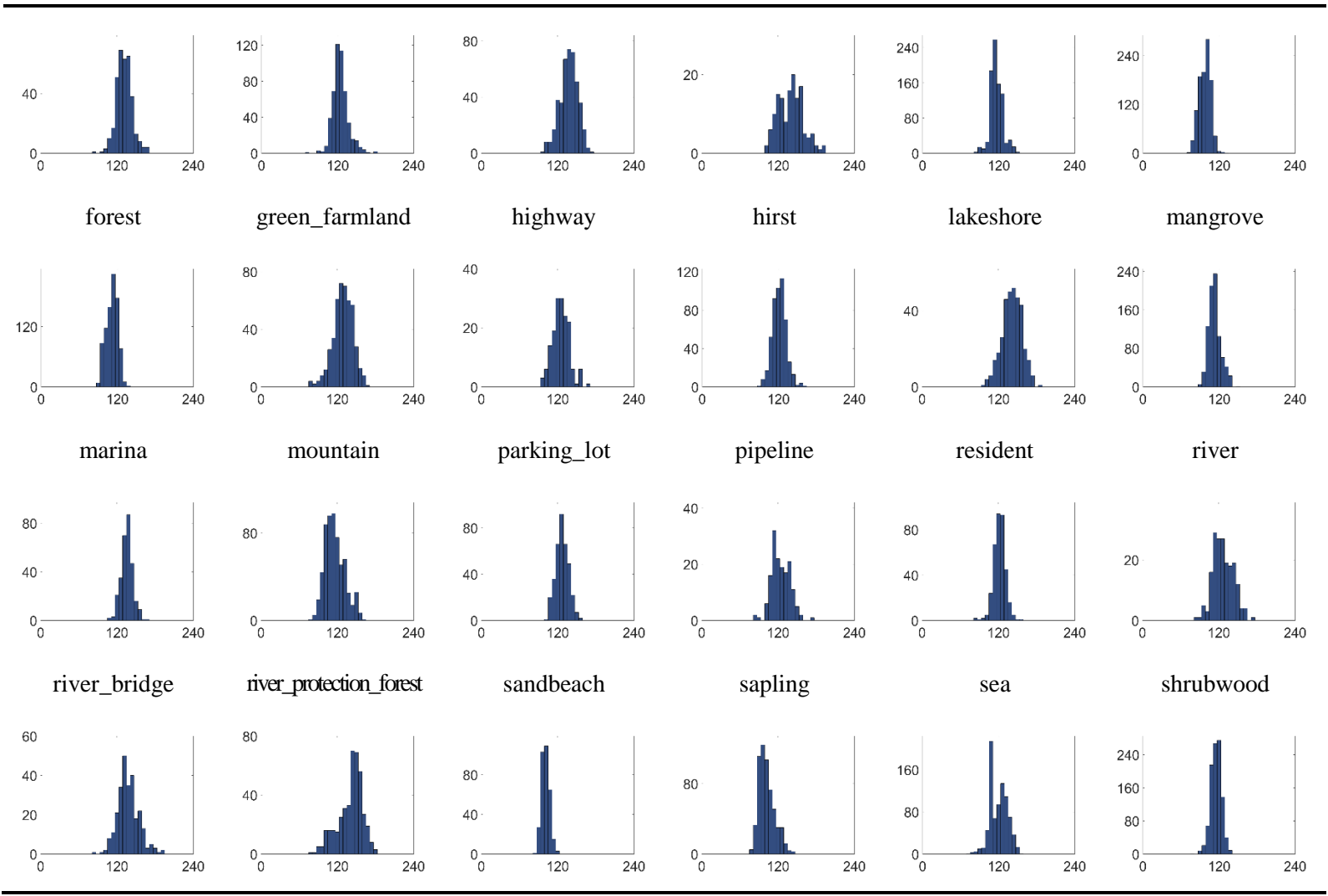

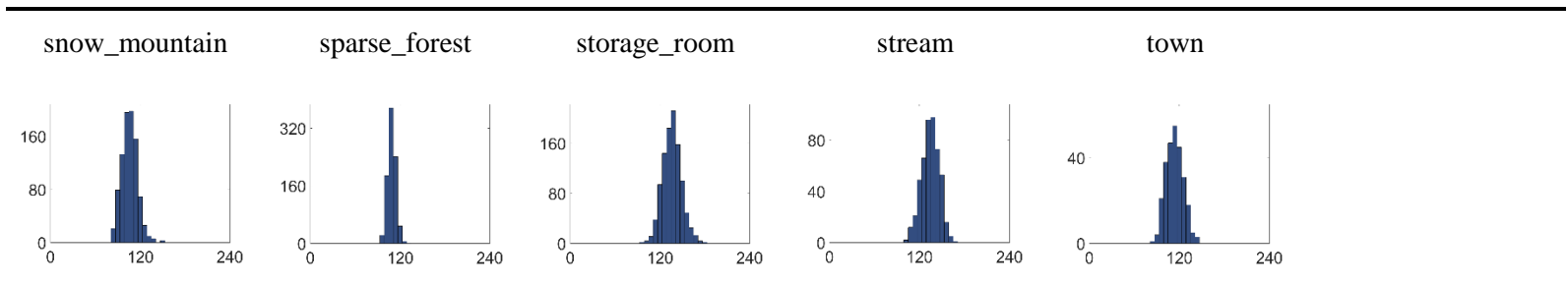

**Table 6-Conv4**

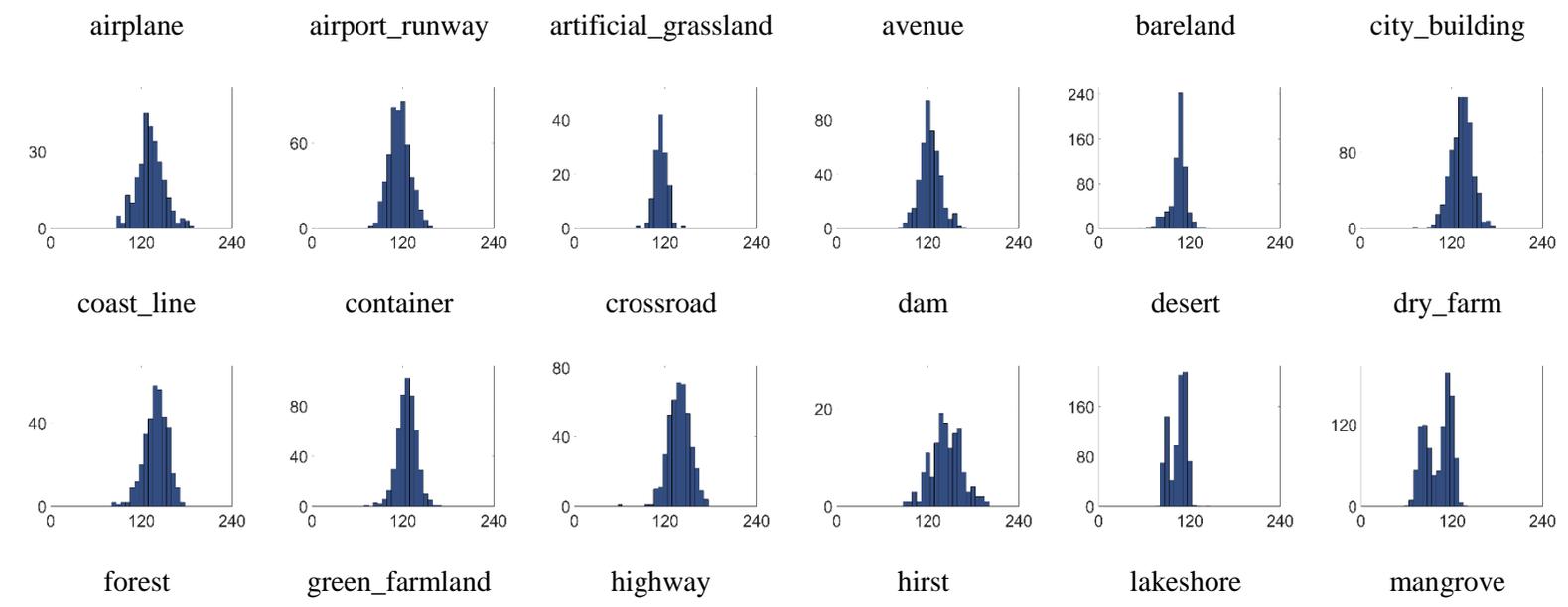

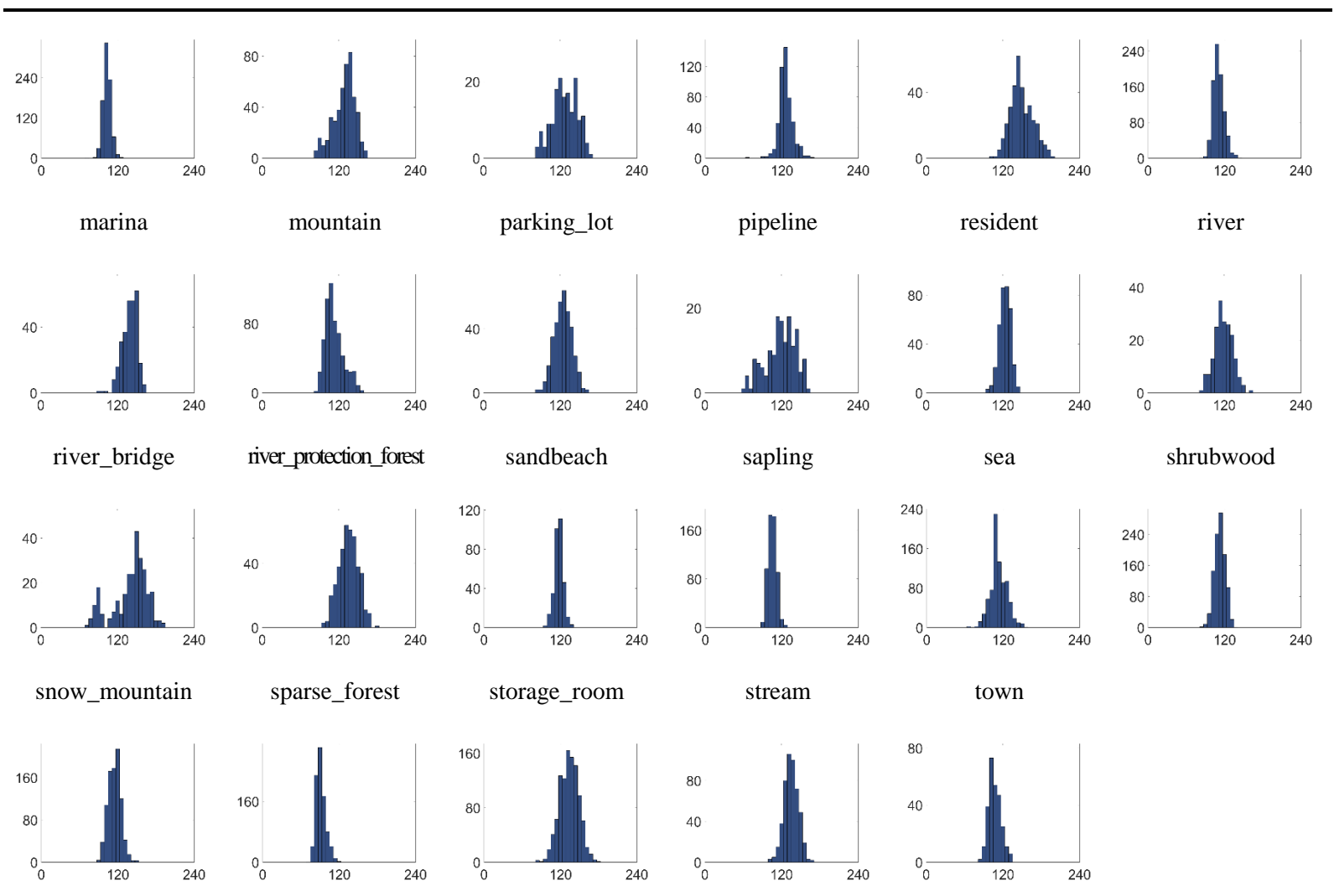

**Table 6-Conv5**

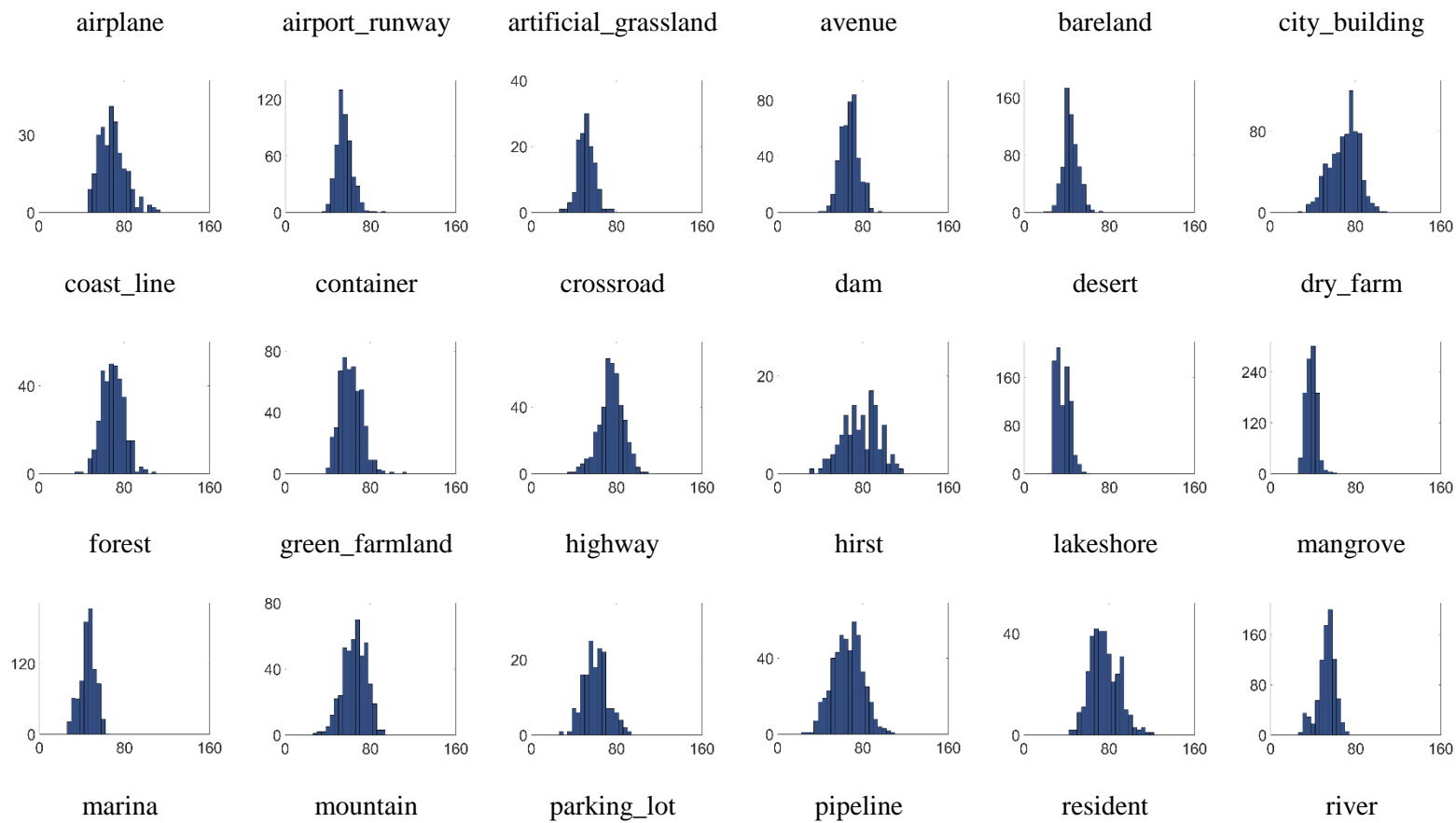

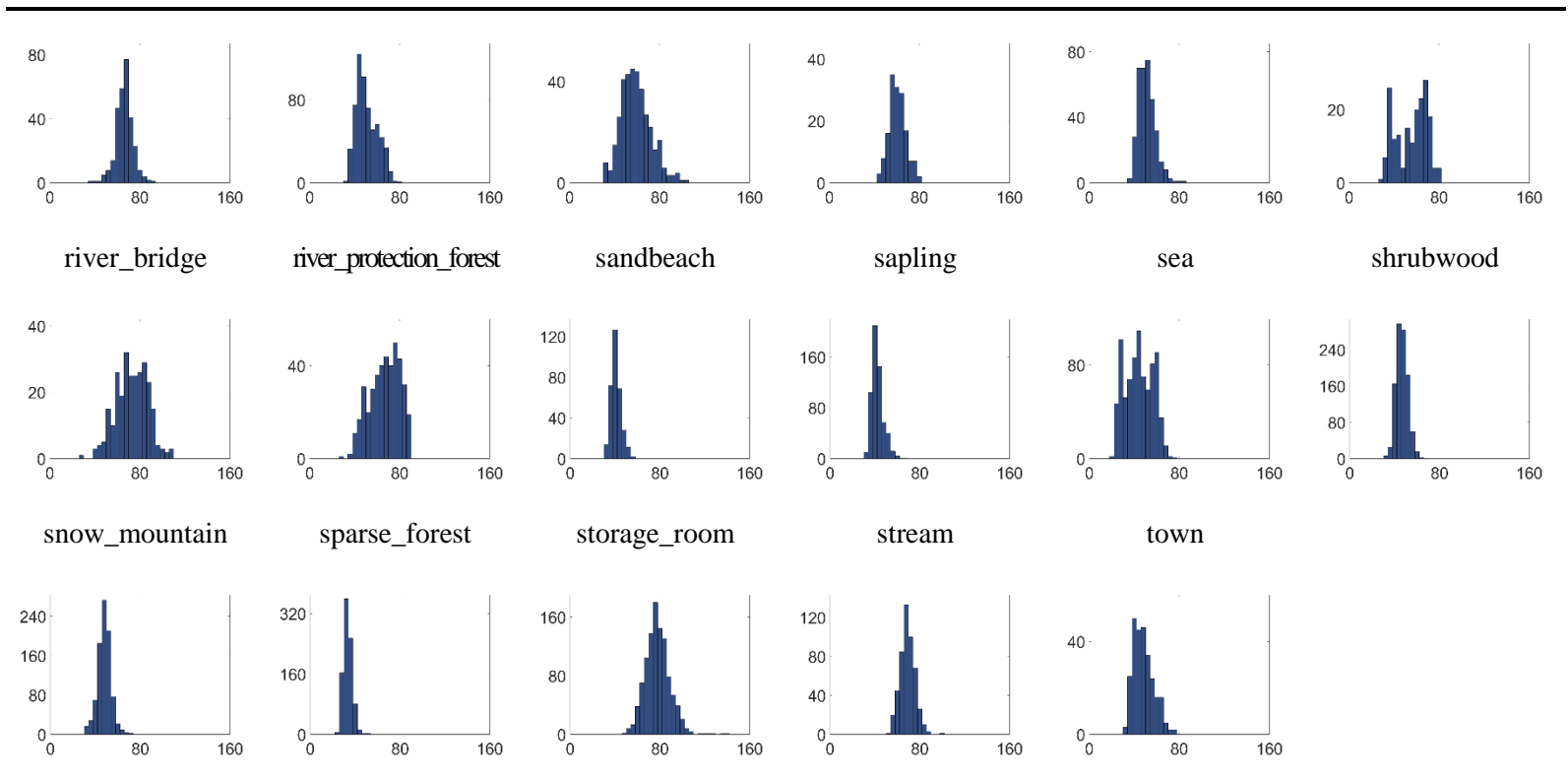

Table 7. The histogram of images number on each activated neuron. The abscissa indicates the images number on each activated neuron, and the ordinate shows the number of neurons.

**Table 7-Conv1**

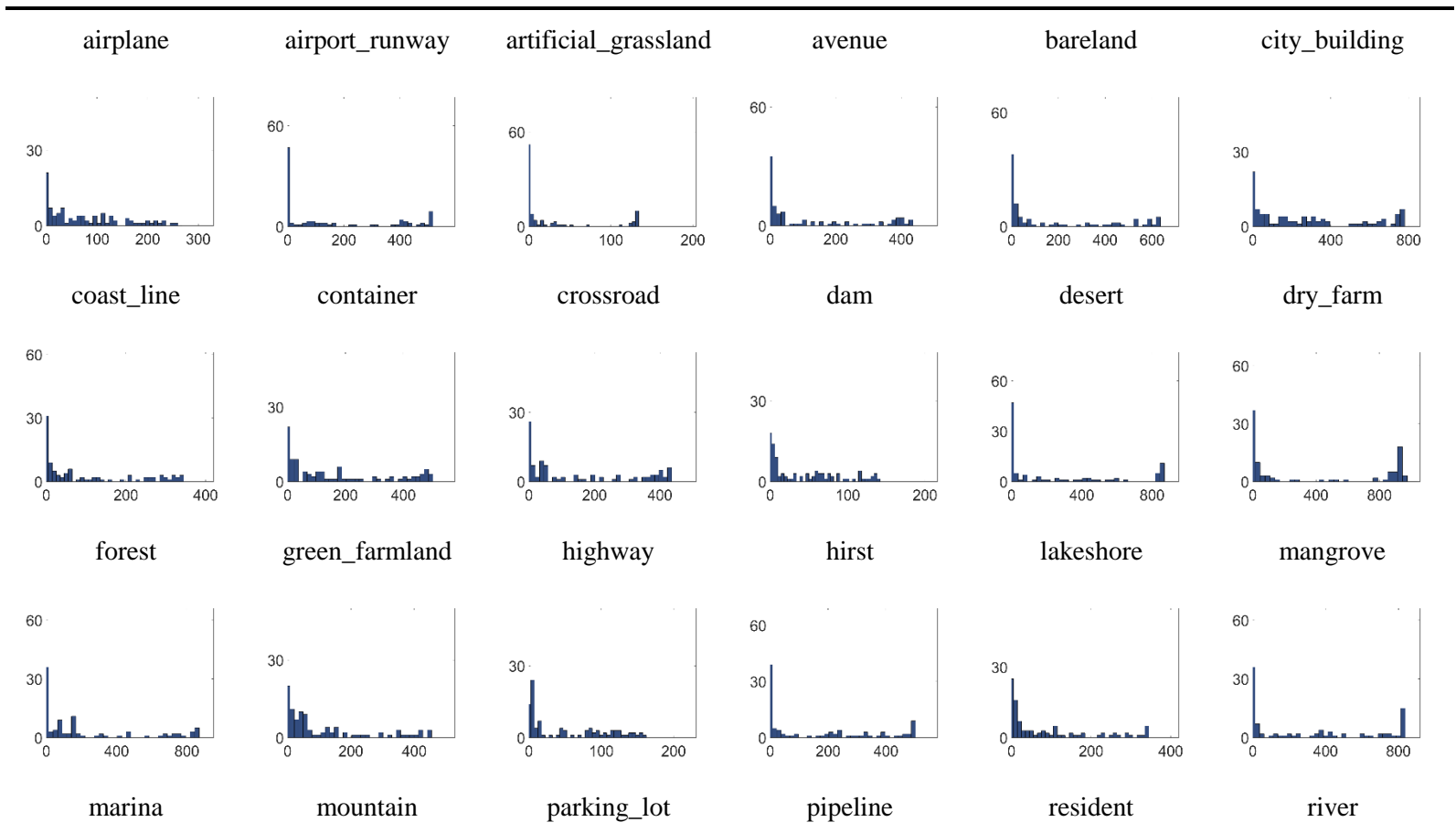

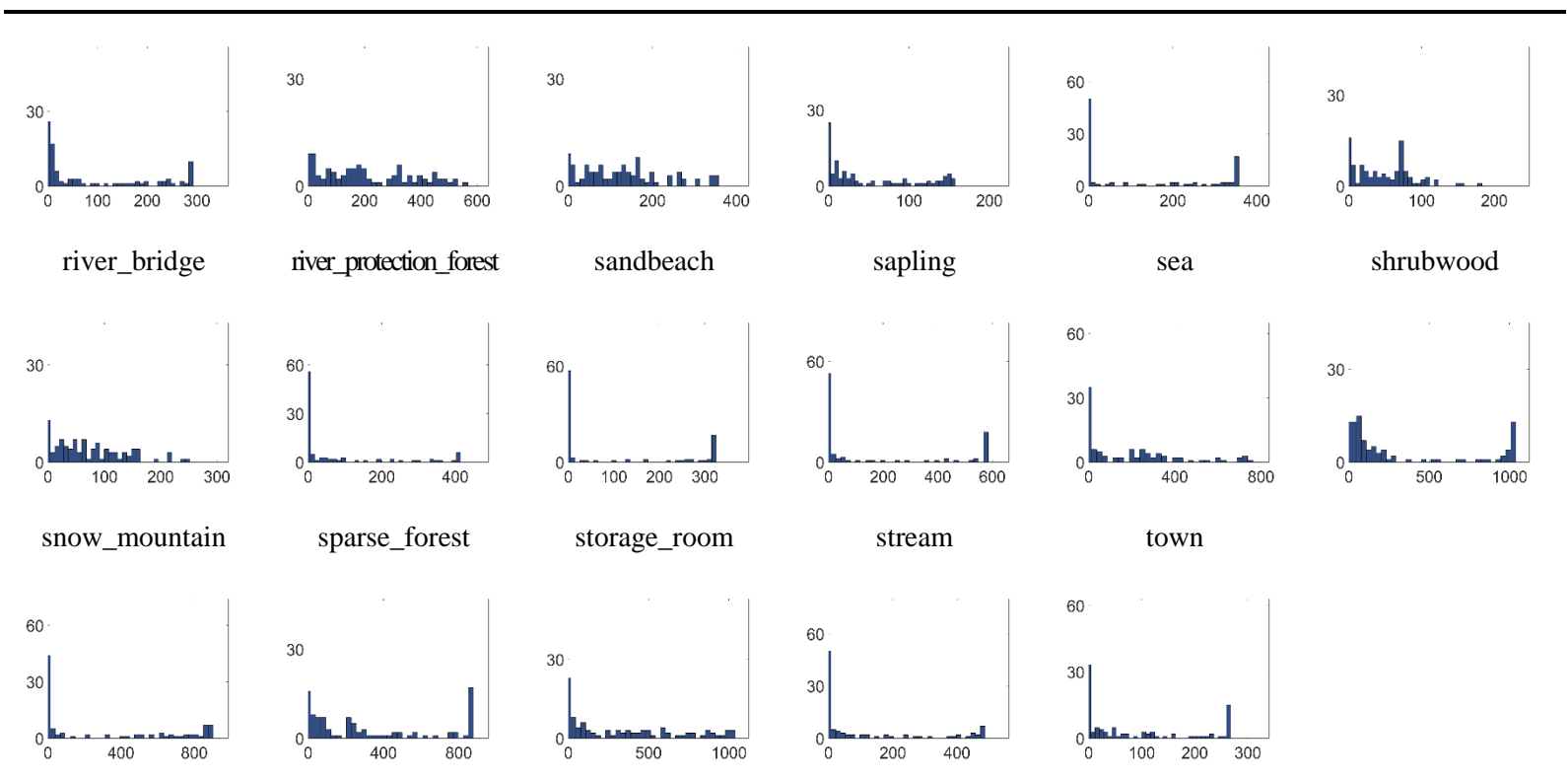

**Table 7-Conv2**

| airplane | airport_runway | artificial_grassland | avenue | bareland | city_building |

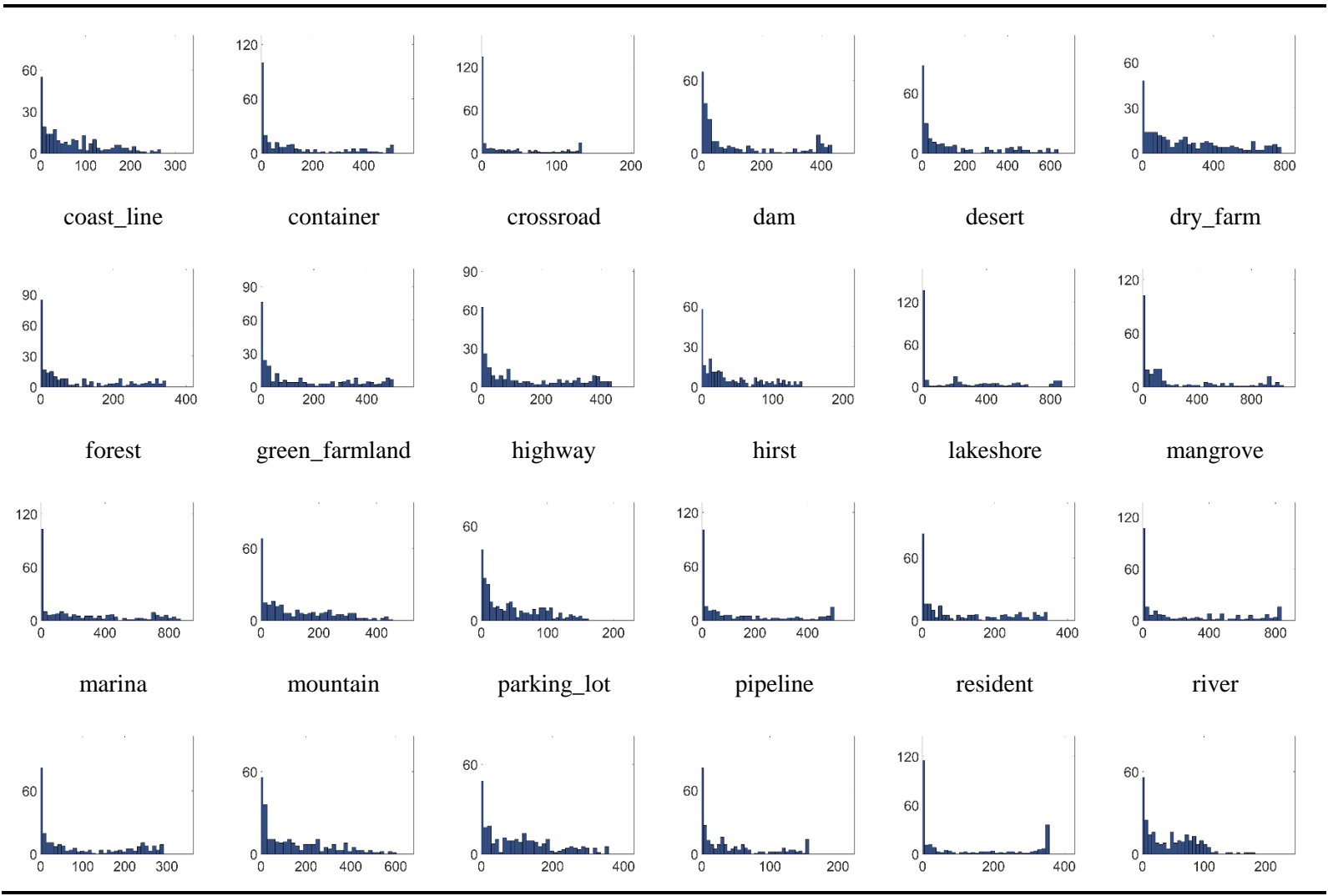

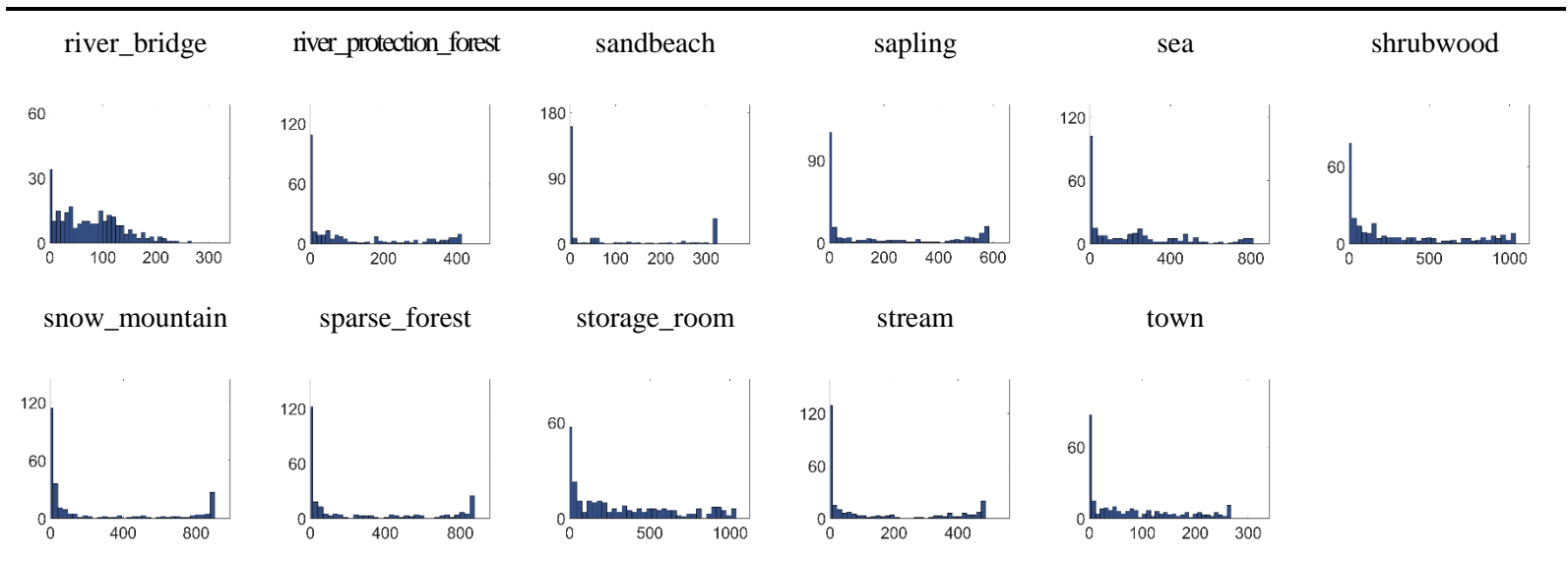

Table 7-Conv3

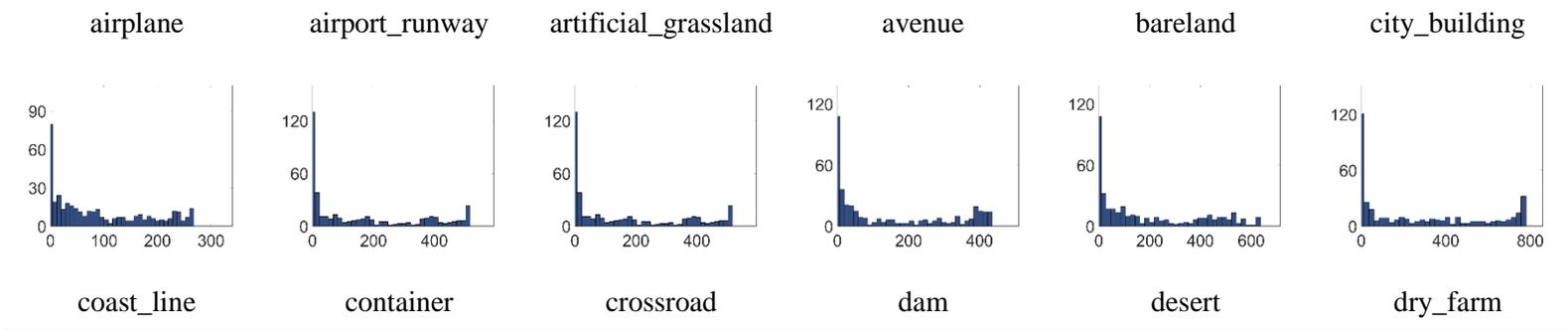

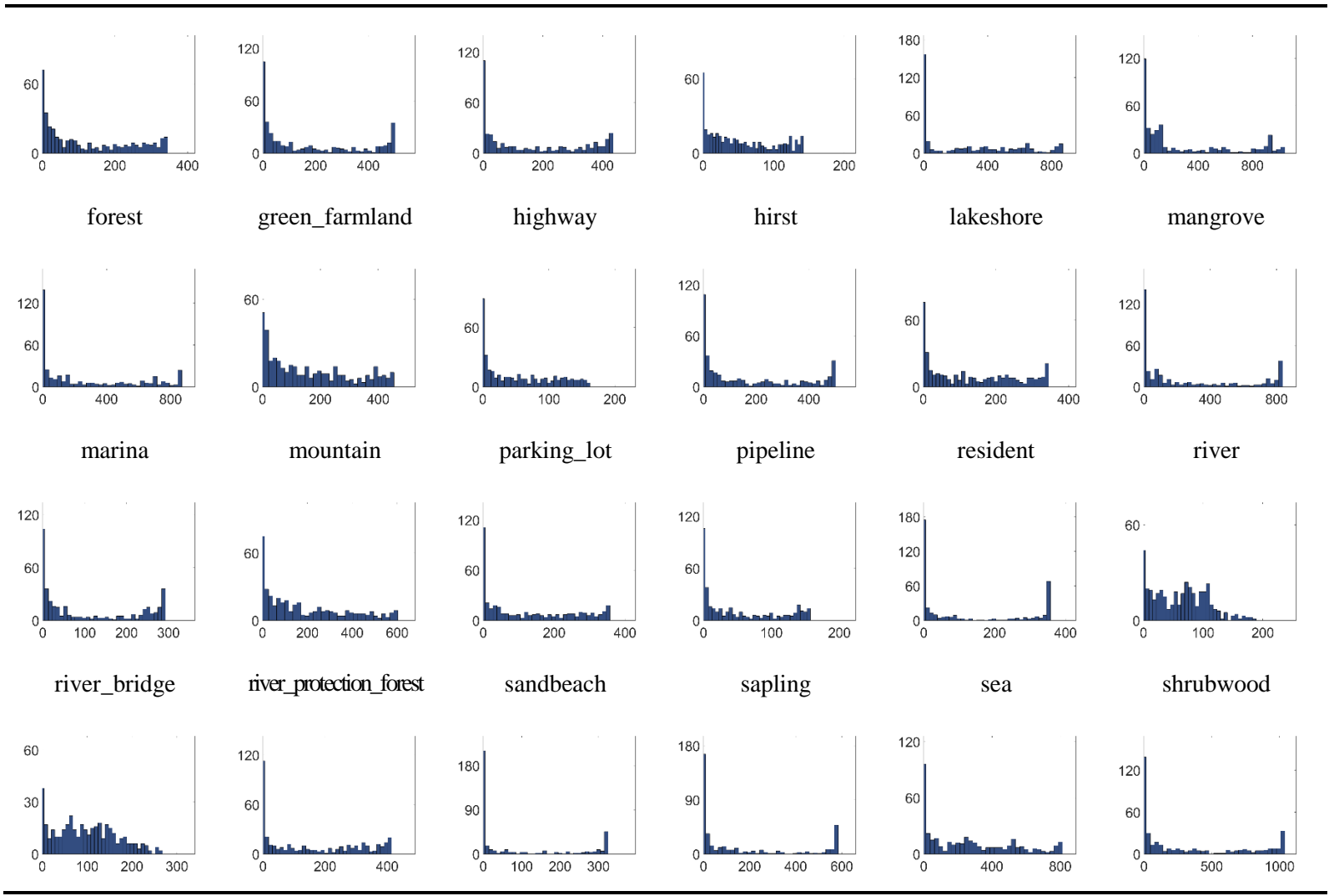

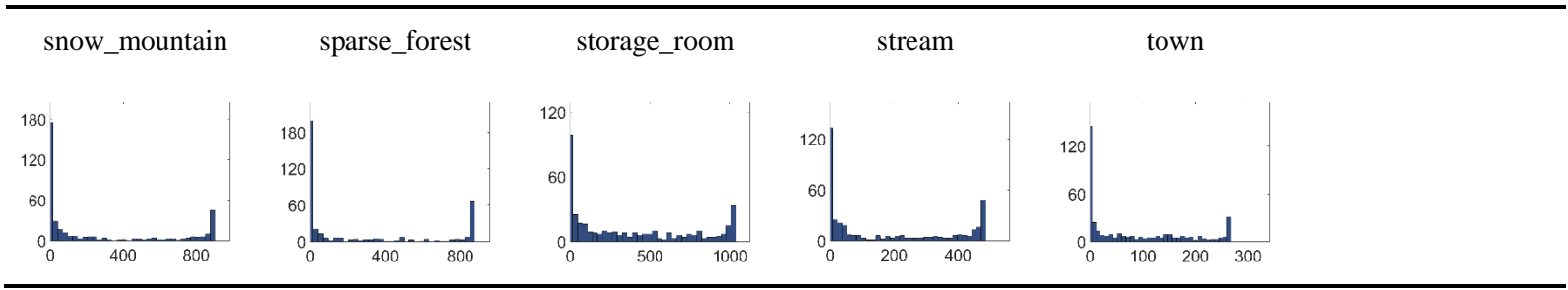

**Table 7-Conv4**

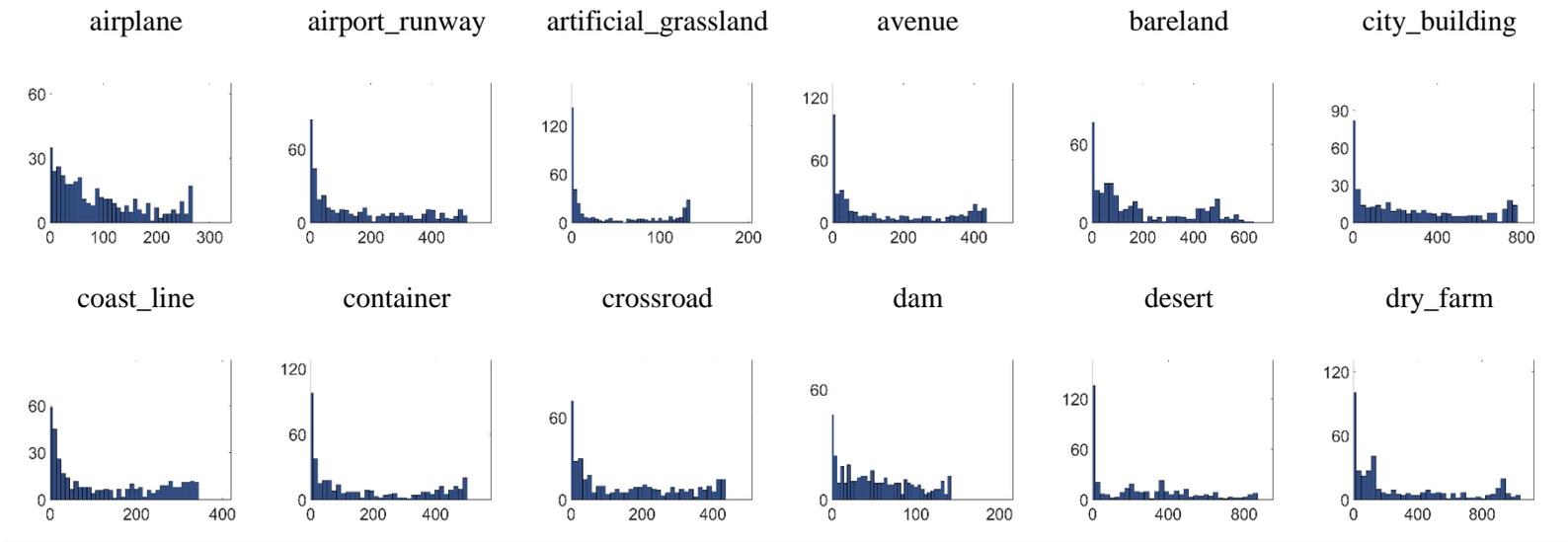

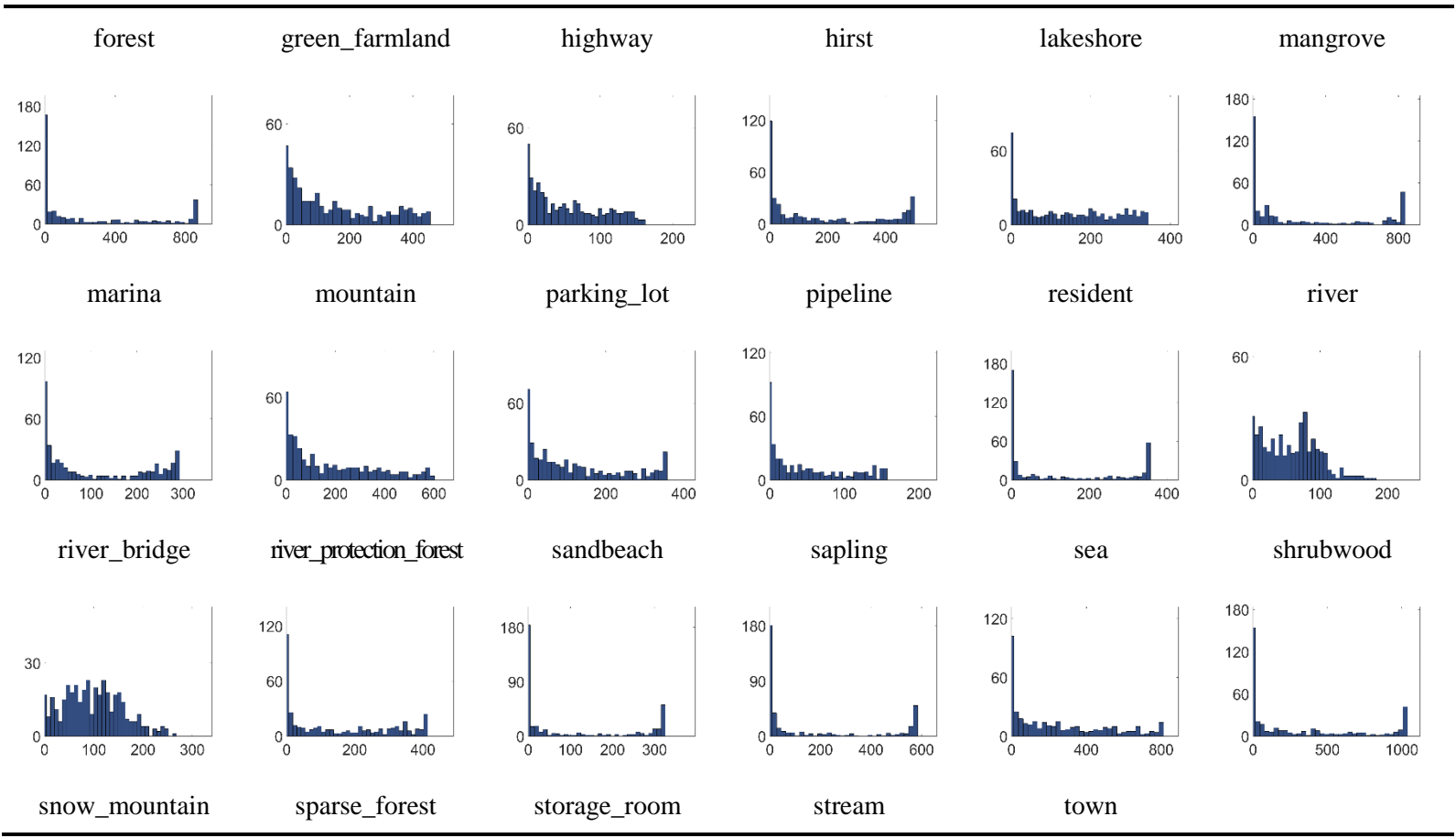

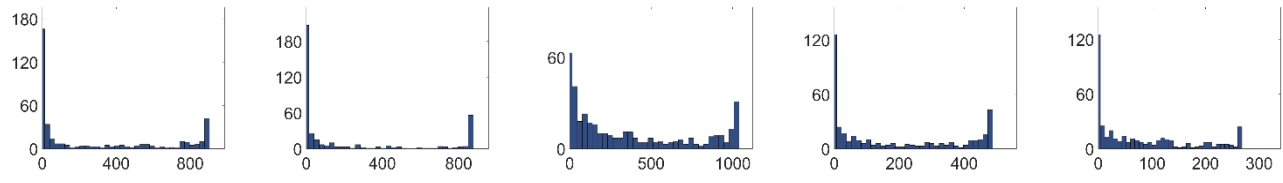

**Table 7-Conv5**

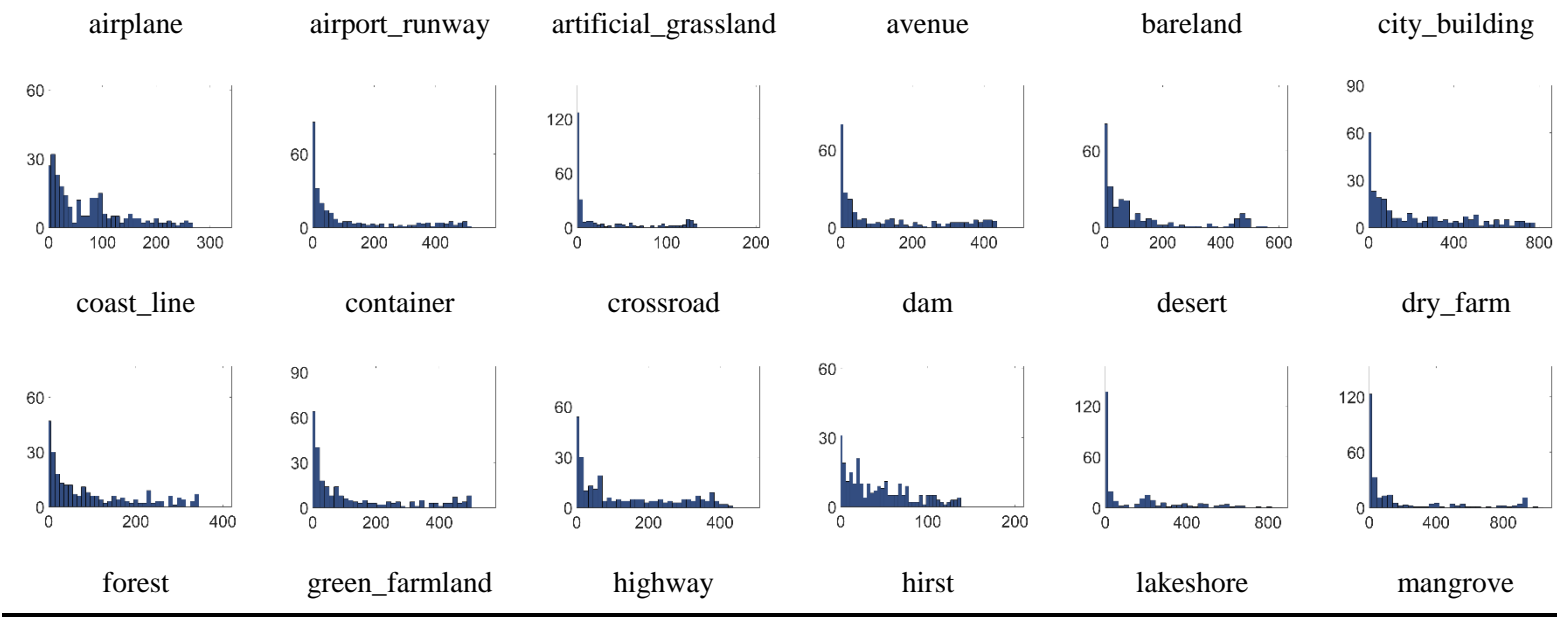

| airplane | airport_runway | artificial_grassland | avenue | bareland | city_building |
| coast_line | container | crossroad | dam | desert | dry_farm |
| forest | green_farmland | highway | hirst | lakeshore | mangrove |

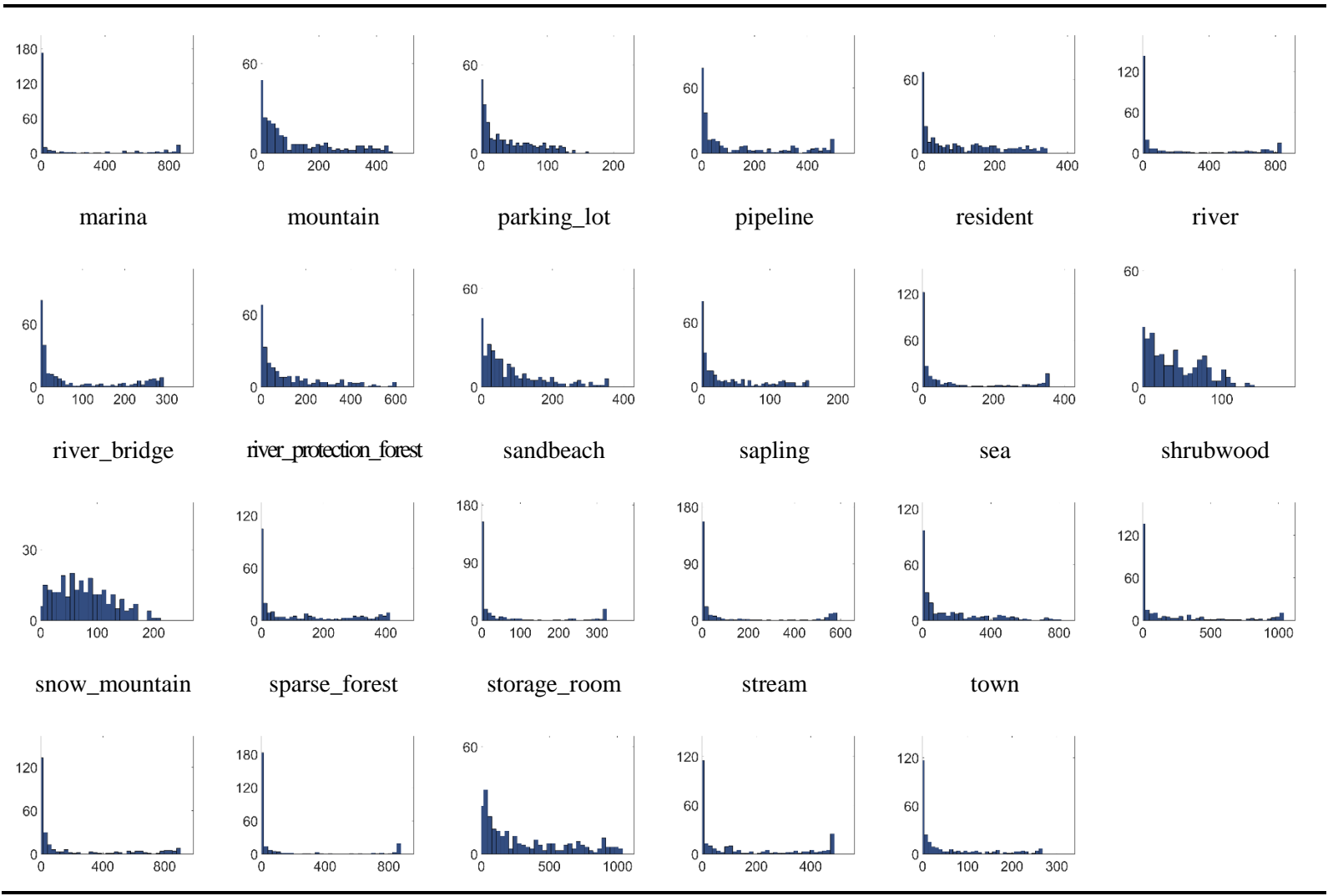

**Acknowledgments:**

This research has also been supported by National Natural Science Foundation of China (41671357, 41571397, and 41501442) and Natural Science Foundation of Hunan Province (2016JJ3144).

This research has also been supported in part by the Open Research Fund Program of Shenzhen Key Laboratory of Spatial Smart Sensing and Services (Shenzhen University), and the Scientific Research Foundation for the Returned Overseas Chinese Scholars, State Education Ministry (50-20150618).

# References

Alain, G., Bengio, Y., 2016. Understanding intermediate layers using linear classifier probes. arXiv preprint arXiv:1610.01644.

Benediktsson, J.A., Chanussot, J., Moon, W.M., 2013. Advances in Very-High-Resolution Remote Sensing. Proceedings of the IEEE 101, 566-569.

Castelluccio, M., Poggi, G., Sansone, C., Verdoliva, L., 2015. Land Use Classification in Remote Sensing Images by Convolutional Neural Networks. Acta Ecologica Sinica 28, 627-635.

Chen, F.X., Roig, G., Isik, L., Boix, X., Poggio, T., 2017. Eccentricity Dependent Deep Neural Networks: Modeling Invariance in Human Vision, AAAI Spring Symposium Series, Science of Intelligence.

Chen, X., Xiang, S., Liu, C., Pan, C., 2014. Vehicle Detection in Satellite Images by Hybrid Deep Convolutional Neural Networks. IEEE Geoscience and Remote Sensing Letters 11, 1797-1801.

Cheng, G., Zhou, P., Han, J., 2016. Learning rotation-invariant convolutional neural networks for object detection in VHR optical remote sensing images. IEEE Transactions on Geoscience and Remote Sensing 54, 7405-7415.

Cheriyadat, A.M., 2014. Unsupervised feature learning for aerial scene classification. IEEE Transactions on Geoscience and Remote Sensing 52, 439-451.

Cichy, R.M., Khosla, A., Pantazis, D., Torralba, A., Oliva, A., 2016. Deep neural networks predict hierarchical spatio-temporal cortical dynamics of human visual object recognition. arXiv preprint arXiv:1601.02970.

Cimpoi, M., Maji, S., Vedaldi, A., 2015. Deep filter banks for texture recognition and segmentation, Proceedings of the IEEE Conference on Computer Vision and Pattern Recognition, pp. 3828-3836.

Deng, L., 2014. A tutorial survey of architectures, algorithms, and applications for deep learning. APSIPA Transactions on Signal and Information Processing 3, e2.

Fingas, M., Brown, C., 2014. Review of oil spill remote sensing. Marine pollution bulletin 83, 9-23.

Han, J., Zhang, D., Cheng, G., Guo, L., Ren, J., 2015. Object detection in optical remote sensing images based on weakly supervised learning and high-level feature learning. IEEE Transactions on Geoscience and Remote Sensing 53, 3325-3337.

He, K., Zhang, X., Ren, S., Sun, J., 2015. Delving deep into rectifiers: Surpassing human-level performance on imagenet classification, Proceedings of the IEEE international conference on computer vision, pp. 1026-1034.

Hu, F., Xia, G.S., Hu, J., Zhang, L., 2015. Transferring Deep Convolutional Neural Networks for the Scene Classification of High-Resolution Remote Sensing Imagery. Remote Sensing 7, 14680-14707.

Kriegeskorte, N., Kievit, R.A., 2013. Representational geometry: integrating cognition, computation, and the brain. Trends in cognitive sciences 17, 401-412.

LeCun, Y., Bengio, Y., Hinton, G., 2015. Deep learning. nature 521, 436-444.

Li, H., 2017. RSI-CB: A Large Scale Remote Sensing Image Classification Benchmark via Crowdsource Data. https://github.com/lehaifeng/RSI-CB.

Li, Y., Yosinski, J., Clune, J., Lipson, H., Hopcroft, J., 2015. Convergent Learning: Do different neural networks learn the same representations? In Proceedings of International Conference on Learning Representation (ICLR).

Liu, Q., Hang, R., Song, H., Li, Z., 2016. Learning Multi-Scale Deep Features for High-Resolution Satellite Image Classification. arXiv preprint arXiv:1611.03591.

Long, Y., Gong, Y., Xiao, Z., Liu, Q., 2017. Accurate Object Localization in Remote Sensing Images Based on Convolutional Neural Networks. IEEE Transactions on Geoscience and Remote Sensing 55, 2486-2498.

Luus, F.P., Salmon, B.P., van den Bergh, F., Maharaj, B.T.J., 2015. Multiview deep learning for land-use classification. IEEE Geoscience and Remote Sensing Letters 12, 2448-2452.

Ma, X., Geng, J., Wang, H., 2015. Hyperspectral image classification via contextual deep learning. EURASIP Journal on Image and Video Processing 2015, 20.

Mahendran, A., Vedaldi, A., 2015. Understanding deep image representations by inverting them, Proceedings of the IEEE Conference on Computer Vision and Pattern Recognition, pp. 5188-5196.

Nogueira, K., Dalla Mura, M., Chanussot, J., Schwartz, W.R., dos Santos, J.A., 2016. Learning to semantically segment high-resolution remote sensing images, Pattern Recognition (ICPR), 2016 23rd International Conference on. IEEE, pp. 3566-3571.

Nogueira, K., Penatti, O.A., dos Santos, J.A., 2017. Towards better exploiting convolutional neural networks for remote sensing scene classification. Pattern Recognition 61, 539-556.

Paul, A., Venkatasubramanian, S., 2014. Why does Deep Learning work?-A perspective from Group Theory. arXiv preprint arXiv:1412.6621.

Penatti, O.A., Nogueira, K., dos Santos, J.A., 2015. Do deep features generalize from everyday objects to remote sensing and aerial scenes domains?, Proceedings of the IEEE Conference on Computer Vision and Pattern Recognition Workshops, pp. 44-51.

Scott, G.J., England, M.R., Starms, W.A., Marcum, R.A., Davis, C.H., 2017. Training Deep Convolutional Neural Networks for Land-Cover Classification of High-Resolution Imagery. IEEE Geoscience & Remote Sensing Letters 14, 549 - 553.

Serre, T., Wolf, L., Poggio, T., 2005. Object Recognition with Features Inspired by Visual Cortex, IEEE Computer Society Conference on Computer Vision and Pattern Recognition, pp. 994-1000.

Shao, W., Yang, W., Liu, G., Liu, J., 2012. Car detection from high-resolution aerial imagery using multiple features, Geoscience and Remote Sensing Symposium (IGARSS), 2012 IEEE International. IEEE, pp. 4379-4382.

Sun, Y., Wang, X., Tang, X., 2015. Deeply learned face representations are sparse, selective, and robust, Proceedings of the IEEE Conference on Computer Vision and Pattern Recognition, pp. 2892-2900.

Tang, J., Deng, C., Huang, G., Zhao, B., 2015. Compressed-Domain Ship Detection on Spaceborne Optical Image Using Deep Neural Network and Extreme Learning Machine. IEEE Transactions on Geoscience and Remote Sensing 53, 1174-1185.

Wei, D., Zhou, B., Torrabla, A., Freeman, W., 2015. Understanding intra-class knowledge inside CNN. arXiv preprint arXiv:1507.02379.

Yosinski, J., Clune, J., Bengio, Y., Lipson, H., 2014. How transferable are features in deep neural networks?, International Conference on Neural Information Processing Systems, pp. 3320-3328.

Yosinski, J., Clune, J., Nguyen, A., Fuchs, T., Lipson, H., 2015. Understanding neural networks through deep visualization. arXiv preprint arXiv:1506.06579.

Zhang, F., Du, B., Zhang, L., 2015. Saliency-guided unsupervised feature learning for scene classification. IEEE Transactions on Geoscience and Remote Sensing 53, 2175-2184.

Zhao, W., Du, S., 2016. Learning multiscale and deep representations for classifying remotely sensed imagery. ISPRS Journal of Photogrammetry and Remote Sensing 113, 155-165.

Zhao, W., Guo, Z., Yue, J., Zhang, X., Luo, L., 2015. On combining multiscale deep learning features for the classification of hyperspectral remote sensing imagery. International Journal of Remote Sensing 36, 3368-3379.

Zhou, B., Khosla, A., Lapedriza, A., Oliva, A., Torralba, A., 2016. Learning deep features for discriminative localization, Proceedings of the IEEE Conference on Computer Vision and Pattern Recognition, pp. 2921-2929.

Zhou, W., Newsam, S., Li, C., Shao, Z., 2017. Learning Low Dimensional Convolutional Neural Networks for High-Resolution Remote Sensing Image Retrieval. Remote Sensing 9, 489.